\documentclass[runningheads]{llncs}

\usepackage{eccv}

\usepackage{microtype}
\usepackage{subcaption}

\usepackage{enumitem}
\usepackage{multirow}
\usepackage{colortbl}
\usepackage{microtype}
\usepackage{comment}
\usepackage[hang,flushmargin]{footmisc}

\usepackage{amsfonts}

\setlist[itemize]{align=parleft,left=0pt,topsep=1mm,itemsep=0mm,parsep=1mm}

\definecolor{azure(colorwheel)}{rgb}{0.0, 0.5, 1.0}
\definecolor{R5}{rgb}{0.0, 0.7, 0.1}
\definecolor{yw}{rgb}{0.01176, 0.5490, 0.5490}
\definecolor{R123}{rgb}{0.36, 0.54, 0.66}
\definecolor{R1234}{rgb}{0.7, 0.75, 0.71}
\definecolor{applegreen}{rgb}{0.55, 0.71, 0.0}
\definecolor{R132}{rgb}{0.0, 0.0, 1.0}
\definecolor{postechred}{rgb}{0.784, 0.003, 0.313}
\definecolor{gu}{rgb}{0.5460, 0.1755, 0.2766}
\definecolor{el}{rgb}{0.9764, 0.447, 0.447}
\definecolor{hyos}{rgb}{0.662, 0.482, 0.960}
\definecolor{ballblue}{rgb}{0.13, 0.67, 0.8}
\definecolor{cornellred}{rgb}{0.7, 0.11, 0.11}
\definecolor{darkcyan}{rgb}{0.0, 0.55, 0.55}
\definecolor{CuGray}{gray}{0.9}
\definecolor{airforceblue}{rgb}{0.36, 0.54, 0.66}
\definecolor{rev}{rgb}{0.784, 0.003, 0.313}
\definecolor{pink}{cmyk}{0, 0.7808, 0.4429, 0.1412}
\definecolor{amethyst}{rgb}{0.6, 0.4, 0.8}
\definecolor{black}{rgb}{0.0, 0.0, 0.0}
\definecolor{tb3_yellow}{rgb}{0.996, 1.0, 0.6}
\definecolor{R123}{rgb}{0.980, 0.8, 0.604}
\definecolor{R512}{rgb}{0.972, 0.6, 0.6}
\definecolor{dimgray}{rgb}{0.41, 0.41, 0.41}
\definecolor{R3}{rgb}{0.8, 0.25, 0.33}
\definecolor{bleudefrance}{rgb}{0.19, 0.55, 0.91}
\definecolor{R6}{rgb}{0.265, 0.445, 0.765}
\definecolor{blue(ryb)}{rgb}{0.01, 0.28, 1.0}
\definecolor{R4}{rgb}{1.0, 0.49, 0.0}
\definecolor{Gray}{gray}{0.88}
\definecolor{green(ncs)}{rgb}{0.0, 0.62, 0.42}
\definecolor{brightpink}{rgb}{1.0, 0.0, 0.5}
\definecolor{alizarin}{rgb}{0.82, 0.1, 0.26}
\definecolor{clova}{rgb}{0.24, 0.63, 0.33}
\definecolor{orange-red}{rgb}{1.0, 0.27, 0.0}
\definecolor{nicegreen}{rgb}{0.0, 0.7, 0.1}
\definecolor{lightblue}{RGB}{235,245,255}
\definecolor{lightorange}{RGB}{255,183,0}

\definecolor{kellygreen}{rgb}{0.3, 0.73, 0.09}

\newcommand{\kjs}[1]{\textcolor{black}{#1}}
\newcommand{\cam}[1]{\textcolor{black}{#1}}

\newcommand{\rev}[1]{\textcolor{black}{#1}}
\newcommand{\lightorange}[1]{\textcolor{lightorange}{#1}}

\newcolumntype{g}{>{\columncolor{CuGray}}c}
\newcolumntype{z}{>{\columncolor{CuGray}}l}

\renewcommand{\paragraph}[1]{\vspace{1mm}\noindent\textbf{#1.}\,}

\usepackage{xspace}

\makeatletter
\def\@fnsymbol#1{\ensuremath{\ifcase#1\or *\or \dagger\or \ddagger\or
   \mathsection\or \mathparagraph\or \|\or **\or \dagger\dagger
   \or \ddagger\ddagger \else\@ctrerr\fi}}
\makeatother

\def\onedot{.\@\xspace}
 
\def\ie{\emph{i.e}\onedot}

\newcommand{\Sref}[1]{Sec.~\ref{#1}}

\newcommand{\Fref}[1]{Fig.~\ref{#1}}
\newcommand{\Tref}[1]{Table~\ref{#1}}

\newcommand{\be}{\begin{eqnarray}}
\newcommand{\ee}{\end{eqnarray}}
\newcommand{\bee}{\begin{eqnarray*}}
\newcommand{\eee}{\end{eqnarray*}}

\newcommand{\matrixb}{\left[ \begin{array}}
\newcommand{\matrixe}{\end{array} \right]}

\newcommand{\kim}[1]{\textcolor{black}{#1}}

\usepackage{amssymb}
\usepackage{pifont}
\newcommand{\xmark}{\ding{55}}%

\newcommand*{\crosssymbol}{%
    \text{%
      \raise 1ex\hbox{%
        \rlap{\vrule height.2pt depth.2pt width .75ex}%
        \hbox to .75ex{\hss\vrule height .5ex depth 1ex\hss}%
      }%
    }%
}
\newcommand{\red}[1]{{\color{red}#1}}

\makeatletter
\DeclareRobustCommand\onedot{\futurelet\@let@token\@onedot}
\def\@onedot{\ifx\@let@token.\else.\null\fi\xspace}
 
\def\ie{\emph{i.e}\onedot}

\usepackage{amsmath,amsfonts,bm}

\def\eqref#1{equation~\ref{#1}}

\def\1{\bm{1}}

\DeclareMathAlphabet{\mathsfit}{\encodingdefault}{\sfdefault}{m}{sl}
\SetMathAlphabet{\mathsfit}{bold}{\encodingdefault}{\sfdefault}{bx}{n}

\usepackage{wrapfig}
\usepackage{lipsum}
\usepackage{url}
\usepackage{float}

\usepackage{graphicx}
\usepackage{booktabs}

\usepackage[accsupp]{axessibility}  

\usepackage{hyperref}

\usepackage{orcidlink}

\begin{document}

\title{SA-ResGS: Self-Augmented Residual 3D Gaussian Splatting for Next Best View Selection} 

\titlerunning{SA-ResGS}

\author{Kim Jun-Seong\inst{1}\thanks{This work was conducted while authors were with Huawei Noah’s Ark Lab in London}\orcidlink{0000-0001-7570-6508} \quad
Tae-Hyun Oh\inst{2}\orcidlink{0000-0003-0468-1571} \\
Eduardo P\'erez-Pellitero\inst{3}\orcidlink{0000-0001-9096-4740} \quad
Youngkyoon Jang\inst{4\ast}\thanks{denotes the corresponding author}\orcidlink{0000-0002-0068-003X}
}

\authorrunning{K.~Jun-Seong et al.}

\institute{
{$^{1}$POSTECH} \quad
{$^{2}$KAIST}  \quad
{$^{3}$Huawei Noah's Ark Lab}  \quad
{$^{4}$Rivian} 
\email{junseong.kim@postech.ac.kr}\\
\url{https://saresgs.github.io/}
}

\maketitle

\begin{abstract}
We propose Self-Augmented Residual 3D Gaussian Splatting, a novel framework for stabilizing uncertainty quantification and enhancing uncertainty-aware supervision in Next-Best-View selection for active scene reconstruction. To efficiently estimate scene coverage, SA-ResGS generates geometry-consistent Self-Augmented point clouds (SA-Points) via triangulation between observed training views and rasterized extrapolated views. To address the lack of learning signals in underrepresented regions within sparse, wide-baseline settings, we introduce the first skip-connection-inspired residual learning strategy tailored for 3DGS. This mechanism amplifies gradient flow to weakly contributing, high-uncertainty Gaussians. Our contributions are threefold: (1) a physically grounded, diversified view selection strategy; (2) an uncertainty-aware residual supervision scheme that improves gradient flow and learning stability; and (3) implicitly debiased uncertainty quantification resulting from constrained view selection and residual supervision. Experiments on NeRF Synthetic, Mip-NeRF 360, and challenging extended benchmark from Deep Blending and Tanks and Temples demonstrate that SA-ResGS consistently outperforms state-of-the-art competing methods in both reconstruction quality and view selection robustness.
\keywords{Active Learning \and 3D Gaussian Splatting \and Next-best-view }
\end{abstract}
\section{Introduction}
\begin{figure*}[t!]
    \centering
    \includegraphics[width=1.0\linewidth]{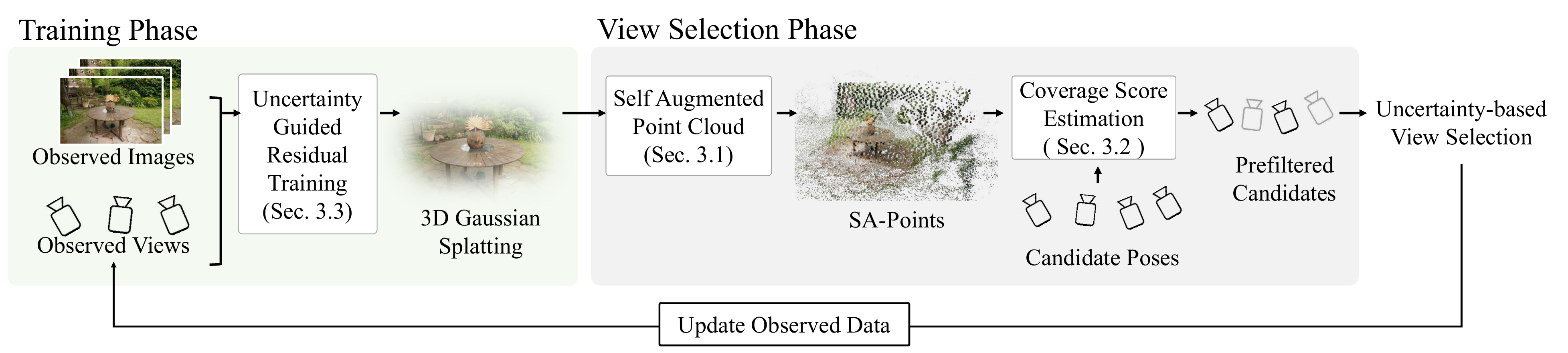}
    \caption{\textbf{Overview of SA-ResGS.} The framework alternates between view selection and training. At each NBV step, Self-Augmented Points (SA-Points) are generated via triangulation between a training view and its extrapolated render, enabling surface-aware coverage estimation (Sec.~\ref{subsec:3_1}). Candidate views are first physically filtered using hash-encoded feature dissimilarity, then ranked by uncertainty quantification scores for final selection, \rev{encouraging exploration of under-observed regions} (Sec.~\ref{subsec:3_2}). During training, residual supervision (Sec.~\ref{subsec:3_3}) combines full and uncertainty-intensified renders to reinforce gradients toward weakly contributing Gaussians, improving training stability and reconstruction quality under sparse views.}
    \label{fig:pipeline}
\end{figure*}

Recent advances in neural rendering, particularly Neural Radiance Fields (NeRFs) \cite{nerf} and 3D Gaussian Splatting (3DGS)~\cite{3dgs}, have significantly advanced photorealistic scene reconstruction~\cite{Yu2024MipSplatting, Niedermayr_2024_CVPR, kulhanek2024wildgaussians, hdr-nsff}, enabling high-fidelity, real-time applications across diverse environments~\cite{fpgs, fprf, drsplat}. Beyond static scene capture, these methods have also spurred broader interest in tackling complex challenges, such as active view selection~\cite{Xiao:CVPR24:NeRFDirector, chen2024gennbv} and uncertainty quantification for next-best-view (NBV) selection~\cite{fisherrf}. Although pre-captured, dense-view training methods can achieve impressive reconstruction quality, in-situ (active) reconstruction—where views are selected and added progressively—remains challenging due to artifacts caused by shape-radiance ambiguity, further exacerbated by limited training views and the dynamics of the view-addition strategy.
Despite the difficulty of reliable uncertainty estimation in this setting, post-hoc methods—such as Laplacian-approximation-based, model-agnostic approaches~\cite{fisherrf, bayesrays}—remain promising, as they provide uncertainty signals without changing the rendering pipeline.
However, we observe that the following 
challenges have been overlooked:
 \begin{itemize}
    \item \textbf{Disregarded physical constraints:} Computational uncertainty is often misaligned with the physical plausibility of reconstructed geometry.
    \item \textbf{Underutilized supervision:} Existing methods rarely convert uncertainty 
    into learning signals, leaving weakly contributing Gaussians undersupervised.
    \item \textbf{Performance dependency:} Reliability of uncertainty 
    remains 
    coupled with training dynamics, especially in early stages with incomplete scene coverage.
\end{itemize}

In response to these challenges, we propose SA-ResGS, a Self-Augmented Residual 3D Gaussian Splatting framework that stabilizes uncertainty quantification and enhances uncertainty-aware supervision for next-best-view selection in progressive scene reconstruction, as shown in Fig.~\ref{fig:pipeline}. SA-ResGS strategically decouples view selection from heavy reliance on uncertainty estimates that are sensitive to internal learning dynamics, thereby promoting more robust, geometry-aware surface coverage. Concretely, we first prefilter candidate views using geometric dissimilarity and then apply uncertainty-based scoring within this subset, effectively implementing a physically grounded, uncertainty-informed selection strategy. To support this process, we construct SA-Points by \kim{reconstructing 3D point maps from}
a training view and its rasterized extrapolated views at each view-selection step, after training on a fixed number of initial views. We encode these SA-Points with a hash-based scene representation to efficiently measure similarity between candidate and previously selected views. We then select the most dissimilar candidate views to improve coverage of unseen regions. 

While the physically grounded and uncertainty-informed selection strategy enhances overall scene coverage, it inadvertently reduces multi-view overlap, since more dissimilar views are less likely to observe shared regions, thereby weakening multi-view geometric constraints.
To counterbalance this effect without sacrificing the benefits of diverse view selection, we introduce a residual supervision mechanism
that provides additional learning signals targeted at under-optimized Gaussians. These Gaussians, which typically correspond to underrepresented regions in sparse views~\cite{jang2025comapgs}, are otherwise often overlooked due to their minimal contribution to the 3DGS rendering process.
Specifically, SA-ResGS additionally rasterizes color images using a targeted subset that combines a small fraction of the most uncertain Gaussians with a majority of the originally visible ones. Inspired by Dropout~\cite{DropGaussian_CVPR2025, srivastava2014dropout} and Hard Negative Mining~\cite{xuan2020hnm, JANG2019_hnm}, this strategy acts similarly to ResNet skip connections~\cite{He_2016_CVPR}; it amplifies supervision for under-optimized Gaussians that otherwise receive weak gradients, while remaining fully compatible with conventional 3DGS pipelines.


The main contributions of SA-ResGS are threefold:
 \begin{itemize}
    \item \textbf{Physically grounded view selection:} We propose a geometry-aware strategy using SA-Points to guide next-best-view selection, enforcing physical plausibility and promoting more balanced, coverage-oriented exploration.
    \item \textbf{Residual learning for 3DGS:} We introduce the first residual supervision framework specifically designed for 3DGS, addressing the vanishing gradient problem by reinforcing weakly supervised Gaussians and improving both optimization stability and reconstruction quality.
    \item \textbf{Unbiased uncertainty quantification:} By jointly improving the view distribution and supervising under-optimized Gaussians, SA-ResGS mitigates geometric sparsity and density bias, leading to fairer and more reliable uncertainty estimates throughout training.
\end{itemize}

\section{Related Work}
\label{sec:2}

\noindent\textbf{Next-best-view selection.}  
NBV selection originated from the robotics community as a strategy to efficiently guide in-situ scene capture, where the goal is to incrementally select viewpoints that maximally reduce reconstruction ambiguity~\cite{connolly1985determination, scott2003view, delmerico2018comparison}.
Classical NBV methods
\kjs{primarily relied on geometric heuristics,}
selecting views based on geometric coverage~\cite{Dunn2009NextBV, scone, macarons}, viewpoint entropy~\cite{view_entropy}, or visibility 
~\cite{bircher2016receding, Sun_2021_CVPR, naruto, activegamer}. 
While subsequent learning-based approaches attempted to model scene-specific view policies via reinforcement learning~\cite{WANG2024_rl-based, NBVrl_2021}, they often struggled with cross-scene generalization.
More recent efforts explore 
information theoretic formulations,  
such as FisherRF~\cite{fisherrf}, offer a principled formulation for uncertainty-based NBV in neural fields.
In our work, we build on this line~\cite{fisherrf, bayesrays, pupgs, pop_gs} by integrating physically grounded geometry priors to 
stabilize early-stage view planning, particularly when minimal visual input is available.

\noindent\textbf{Uncertainty estimation for 3DGS.}  
Quantification of uncertainty plays a pivotal role in active 3D reconstruction, particularly for guiding view selection.
\kjs{
In the context of 3DGS, earlier attempts to estimate uncertainty primarily relied on ensemble-based estimates~\cite{ensemblenerf} or variational inference~\cite{snerf, bayesiannerf, cfnerf, lyu2024manifold}.
However, these approaches require redundant multiple inferences or modifications to the 3DGS parameters, making them incompatible with the standard training and rendering pipelines of 3DGS.
To avoid these limitations, recent research has shifted toward post-hoc uncertainty estimation~\cite{fisherrf, bayesrays, pop_gs, pupgs, primu, wang2024avs}, which extracts uncertainty signals, after training is done, without altering the underlying model architecture.
Due to its architectural agnostic nature, post-hoc methods are increasingly explored across diverse applications~\cite{activegrasp, LLMFisher, NBsense}.
}

\kjs{
Representative post-hoc methods,
such as FisherRF~\cite{fisherrf} and BayesRays~\cite{bayesrays}, adapted classical Laplacian approximations on 3DGS setting, pioneering Bayesian approaches to 3DGS uncertainty estimation.
In subsequent work, ActiveViewSelector~\cite{wang2024avs} has introduced image-level rendering quality metrics, and PRIMU~\cite{primu} suggests 3D unprojection of 2D spatial errors to localize regions of high uncertainty.
Despite their promise, these existing post-hoc approaches often overlook the unique training dynamics inherent to 3DGS.
Specifically, these methods remain strongly coupled to the density of underlying Gaussians, leading to biased uncertainty estimates in early training stages when geometry is sparse or unevenly distributed and often misinterpreting under-observed regions as confident. 
To mitigate this limitation, we introduce residual learning, assisted by physically grounded view selection, enabling more loosely coupled uncertainty estimation during early-stage view selection while emphasizing supervision focused on high-uncertainty Gaussians.
}




\noindent\textbf{Residual supervision in 3DGS.}  
While accurate uncertainty estimation helps localize regions having deficient supervision, 
it alone 
is not sufficient to counterbalance 
under-optimized Gaussians in the 3DGS pipeline.
Existing 3DGS methods mainly rely on direct photometric losses~\cite{3dgs} or external depth priors~\cite{Li_2024_CVPR_DNGaussian, Xu_ECCV_2024_MVPGS}, which often fail to sufficiently supervise Gaussians with low opacity or minimal rendering contributions. 
Recent studies, such as pixelSplat~\cite{Charatan_2024_CVPR}, PAPR~\cite{Zhang_2023_NeurIPS}, and PAPR-in-Motion~\cite{Peng_2024_CVPR}, explicitly discuss the vanishing gradient issue and propose solutions including differentiable parameterization of Gaussians, proximity attention-based differentiable rendering, adaptive updates, and activation tuning. 

Despite the various strategies mitigating the vanishing gradient problem, prior approaches lack an explicit mechanism to correct weakly supervised Gaussians, leaving the problem largely unresolved due to insufficient gradient signals.
Although dropout-based approaches~\cite{DropGaussian_CVPR2025} help increase gradient diversity, they operate stochastically and do not target supervision toward the most uncertain or least-updated Gaussians. 
Our method addresses these limitations by introducing the residual supervision strategy for 3DGS.
Residual learning, as popularized by ResNet~\cite{He_2016_CVPR}, has proven effective in mitigating vanishing gradients and improving training stability through skip connections and additive refinement, yet it remains underexplored in the context of 3D Gaussian Splatting. 
We replicate skip-connections among 3D Gaussians along the ray, applying uncertainty-guided rendering to intentionally amplify gradients for under-supervised Gaussians—without altering the underlying rasterization process.

\begin{figure}[t!]
    \centering
    \includegraphics[width=\linewidth]{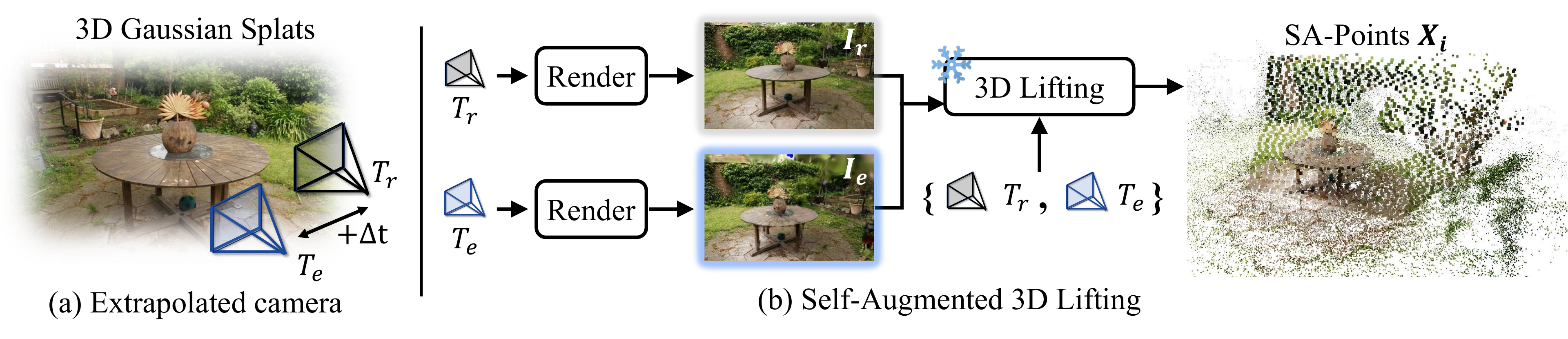}
    \caption{ \textbf{SA-Points Generation process.}
    (a) The virtual camera pose $T_e$ is extrapolated from the reference camera pose $T_r$ by adding translation $\Delta t$.
    (b) The reference image $I_r$ and extrapolated render $I_e$ are then jointly processed by the 3D lifting module to reconstruct SA-Points $X_i$. More details in Suppl.~\Sref{Supp:Implementation}.
    }
    \label{fig:SAPoints}
\end{figure}
\section{Method}
The proposed SA-ResGS framework is illustrated in Fig.~\ref{fig:pipeline}. Built on the 
next-best-view selection method, FisherRF~\cite{fisherrf} as a baseline, SA-ResGS extends it with SA-Points around two core ideas: (1) physically grounded view selection with reduced reliance on uncertainty estimates, and (2) residual supervision that explicitly strengthens weakly contributing Gaussians, which minimally affect rasterized pixels and therefore receive insufficient gradients.
Implementation details and hyperparameter settings are provided in Suppl.~\Sref{Supp:Implementation}.

\subsection{Generation of Self-Augmented Points
}
\label{subsec:3_1}

\kim{
The physically grounded view selection is enabled by a physically aligned surface representation, constructed using SA-Points derived from a single training view.
The overall 
SA-Points generation
pipeline
is visualized in Fig.~\ref{fig:SAPoints}.
Given a reference image $I_r$ with camera pose $\mathbf{T}_r = [\mathbf{R}_r \mid \mathbf{t}_r]$, we render an extrapolated image $I_e$ from a perturbed pose $\mathbf{T}_e = [\mathbf{R}_r \mid \mathbf{t}_r + \Delta \mathbf{t}]$ using 3DGS. 
Dense correspondences $\{ (\mathbf{p}_r^i, \mathbf{p}_e^i) \}$ between $I_r$ and $I_e$ are predicted using the pretrained MASt3R model~\cite{mast3r}, which is robust to moderate viewpoint changes and 
produces
contextually meaningful matches even in the presence of minor geometric distortions.
SA-Point $\mathbf{X}^i$ is triangulated from a 2D correspondences using the projection matrices $\mathbf{P}_r$, $\mathbf{P}_e$, derived by $\mathbf{T}_r$, $\mathbf{T}_e$, with intrinsic from COLMAP~\cite{colmap-0, colmap-1}.
}

However, as triangulation is performed repeatedly during training—while the model is still fitting to a sparse and incomplete geometry—rasterized extrapolated images may occasionally contain rendering noise due to inaccurately placed Gaussians. 
To ensure reliable geometry while fully leveraging the generalization capability of MASt3R, we apply reprojection error-based filtering. 
The reprojection error is defined as :
\begin{align}
\varepsilon^i = \tfrac{1}{2} \left( \left\| \mathbf{p}_r^i - \pi(\mathbf{P}_r \mathbf{X}^i) \right\|_2 + \left\| \mathbf{p}_e^i - \pi(\mathbf{P}_e \mathbf{X}^i) \right\|_2 \right),
\end{align}
\cam{where \( \|\cdot\|_2 \) is the \( \ell_2 \) norm in pixel space,} and SA-Points with $\varepsilon^i < \tau$ are retained.
This filtering step discards geometrically inconsistent points while preserving accurate SA-Points from dense, context-aware matches, even when the extrapolated image is noisier than the original training view. 
Compared to prior methods such as CoMapGS~\cite{jang2025comapgs} or MP-SfM~\cite{pataki2025mpsfm}, our triangulation pipeline produces scale-consistent, surface-aware geometry from a single image by leveraging extrapolated viewpoints rather than requiring multi-view input or monocular depth estimates.
\kjs{
Also, note that, while we use MASt3R as our default 3D lifting module, our framework is modular and compatible to other 3D foundational reconstruction methods.\footnote{Refer to 
Secs.~\ref{subsec:5_1} and \ref{subsec:4_3} for evaluation with an alternative, Depth-Anything-v3~\cite{da3}.}
}

\begin{figure*}[t!]
    \centering        \includegraphics[width=1.0\textwidth]{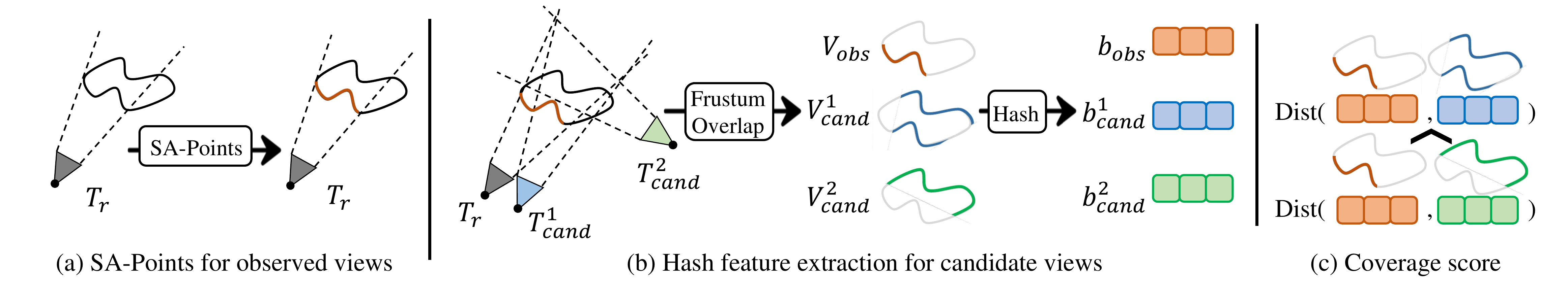}
    \caption{
    \textbf{Physically grounded candidate view selection via surface coverage.} 
    (a) SA-Points from training views define observed voxels \( \mathcal{V}_{\text{obs}} \). 
    (b) Each candidate view generates a binary hash-encoded feature 
    $\mathbf{b}$,
    via frustum-based visibility estimation. 
    (c) Normalized Hamming distance between hash-encoded features quantifies coverage dissimilarity, enabling efficient selection of geometrically complementary views without rendered images or uncertainty scores.
    }
    \label{fig:view_selec}
\end{figure*}
\subsection{Physically Grounded View Selection Algorithm}
\label{subsec:3_2}
We present our physically grounded view selection algorithm for next-best-view (NBV) selection, illustrated in Fig.~\ref{fig:view_selec}. As discussed in Sec.~\ref{sec:2}, NBV selection in 3D Gaussian Splatting (3DGS) is particularly challenging due to the tight coupling between quality of uncertainty estimation and the quality of reconstructed geometry—both are highly sensitive to the sparsity and distribution of Gaussian splats. Under sparse-view settings, where reconstruction starts from as few as four images and new views are incrementally added every 100 training iterations, uncertainty-based NBV strategies often become unreliable. This occurs because uncertainty signals are inherently biased or unstable when the geometry is incomplete or under-constrained. To address this, we introduce a surface-aware guidance mechanism based on SA-Points, which allows view selection to operate independently of the computed uncertainty quantification. By decoupling view selection from 3DGS internal training dynamics, our method provides more stable and physically meaningful candidate views during the early reconstruction, even before the model accumulates sufficient confidence to produce reliable uncertainty.

We begin by discretizing the 3D scene into a voxel grid \( \mathcal{V} = \{ v_k \}_{k=1}^{K} \), where each voxel represents a unit volume. The bounding volume of \( \mathcal{V} \) is defined by the sparse point cloud obtained via structure-from-motion (SfM). A voxel \( v_k \in \mathcal{V} \) is marked as observed if it intersects any SA-Point \( \mathbf{X}^i \) (Sec.~\ref{subsec:3_1}), forming the subset \( \mathcal{V}_{\text{obs}} \subset \mathcal{V} \). To account for potential localization errors and promote coverage continuity, we dilate each occupied voxel using a 3D kernel \( \mathcal{K}_r \) of radius \( r \):
\begin{align}
\tilde{\mathcal{V}}_{\text{obs}} = \bigcup\nolimits_{v_k \in \mathcal{V}_{\text{obs}}} \mathcal{K}_r(v_k),
\end{align}
where \( \tilde{\mathcal{V}}_{\text{obs}} \) denotes the dilated observed region for the current training views. 

For each candidate view \( j \), from the index set of all candidate views $\mathcal{C} = \{1, ..., M\}$, we compute a frustum \( \mathcal{F}_j \subset \mathcal{V} \), defined by the camera intrinsics (field of view) and near/far planes estimated from the SfM point distribution. A voxel is considered potentially visible from view \( j \) if its center lies within the frustum:
\begin{align}
\mathcal{V}_{\text{cand}}^{(j)} = \{ v_k \in \mathcal{V} \mid v_k \in \mathcal{F}_j \}.
\end{align}

To estimate geometric dissimilarity between current coverage and a candidate view (Fig.~\ref{fig:view_selec}), we compute the normalized Hamming distance:
\begin{align}
d_j = \frac{1}{K} \left\| \mathbf{b}_{\text{obs}} \oplus \mathbf{b}_{\text{cand}}^{(j)} \right\|_1,
\end{align}
where \( \oplus \) denotes the element-wise XOR operation between binary vectors, and \( \|\cdot\|_1 \) is the \( \ell_1 \) norm (i.e., the number of differing entries). Here, $\mathbf{b}$ is a binary occupancy vector obtained by mapping voxel coordinates through a fixed random hashing function, following the spatial hashing strategy of Instant-NGP~\cite{mueller2022instant, facthash}. The resulting value \( d_j \in [0, 1] \) measures the proportion of voxels with inconsistent occupancy status between the currently observed volume and the candidate view;
\cam{it quantifies the volume of non-overlapping occupancy between the observed voxel set and a candidate view's frustum voxels, \ie, how many newly covered voxels a candidate adds.} 
Candidate views are then ranked in descending order of their normalized Hamming distances \( d_j \), and the top \( N\% \) (e.g., \( N=20 \)) are retained as the physically filtered candidate set $\mathcal{C}'=\text{Top}_{\text{N}\%}(\mathcal{C}, d)$. 

We apply uncertainty quantification
\footnote{\cam{We adopt uncertainty estimation formulation from FisherRF~\cite{fisherrf}, the uncertainty and expected-information-gain (EIG) formulation is summarized in Suppl.~\Sref{Supp:Implementation}.
}} 
only within \( \mathcal{C}' \), and finalize view selection via finer-level scoring. This two-stage pipeline follows a coarse-to-fine strategy: 
SA-Points provide an explicit estimate of which regions are already observed, and we first select views that maximally expose the remaining unobserved regions to form a stable candidate subset and refines the choice through uncertainty-aware reasoning.
Restricting uncertainty evaluation to $\mathcal{C}'$ reduces computation by avoiding uncertainty scoring for every candidate view. 
At the same time, as $\mathcal{C}'$ is pre-filtered to favor views that expose unobserved regions, the final selection avoids redundant viewpoints and achieves more balanced scene coverage.
\subsection{Uncertainty-Guided Residual Learning in 3DGS}
\label{subsec:3_3}

\begin{figure*}[t!]
    \centering
    \includegraphics[width=\linewidth]{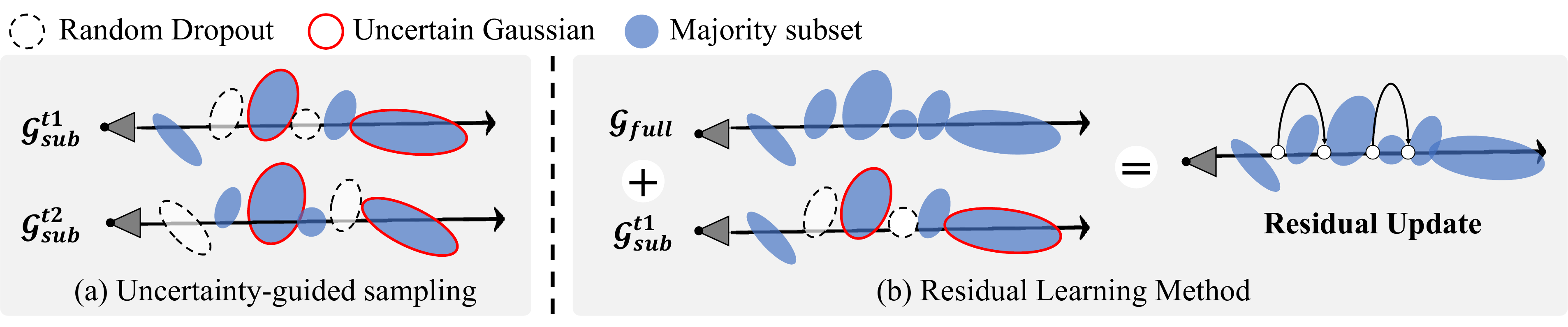}
    \caption{\textbf{Residual supervision in 3DGS.} 
    Overview of the proposed residual supervision strategy.
    (a) At each iteration ($t_1$, $t_2$), \( \mathcal{G}_{\text{sub}} \) combines random and top-uncertain Gaussians; 
    (b) residual supervision in 3DGS mimics ResNet-style skip connections.}
    \label{fig:res_learning}
\end{figure*}

We propose the first residual learning framework for 3DGS that emulates skip connections to address vanishing gradients in weakly contributing Gaussians, as shown in Fig.~\ref{fig:res_learning}. These Gaussians often receive insufficient supervision due to their limited impact on rasterized pixels, particularly in sparse or ambiguous regions. While ResNet~\cite{He_2016_CVPR} mitigates similar issues through skip connections, such mechanisms are infeasible in 3DGS given the dynamic, view-dependent nature of Gaussian properties. Instead, we introduce a rasterizer-agnostic strategy that improves gradient flow by generating auxiliary renders emphasizing high-uncertainty Gaussians. These renders are supervised with 
input
RGB images, forming the basis of the residual supervision scheme described below.

\noindent \textbf{Residual supervision.}  
To reinforce under-supervised Gaussians, we introduce a residual supervision scheme that uses two rendered images from the same training view: one rendered with the full Gaussian set $\mathcal{G}_{\text{full}}$, and another rendered with a guided subset $\mathcal{G}_{\text{sub}}$, as shown in Fig.~\ref{fig:res_learning}(a). We define this subset as:
\begin{align}
\mathcal{G}_{\text{sub}} = \mathcal{G}_{\text{rand}} \cup \mathcal{G}_{\text{uncertain}},
\end{align}
where $\mathcal{G}_{\text{rand}}$ is a random sample comprising $\alpha\%$ of $\mathcal{G}$ (e.g., $\alpha {=} 90$), and $\mathcal{G}_{\text{uncertain}}$ contains the top-$\beta$ most uncertain Gaussians (e.g., $\beta {=} 10$). To estimate uncertainty, we analyze two per-Gaussian attributes: opacity and scale. Gaussians with low opacity contribute minimally to alpha blending during rasterization, while those with large scales blur across pixels and tend to dominate ambiguous or low-texture regions. This ranking identifies Gaussians that are both visually suppressed and spatially diffuse, making them key targets for correction. 

We compute two rendered images: $I_{\text{full}}$ from the full Gaussian set $\mathcal{G}_{\text{full}}$, and $I_{\text{sub}}$ from the uncertainty-intensified subset $\mathcal{G}_{\text{sub}}$. Each is supervised independently against the 
input
image $I_{\text{gt}}$ using $\ell_1$ and SSIM losses:
\begin{align}
\mathcal{L} = \sum_{i \in \{\text{full}, \text{sub}\}} \lambda_i \left[ \mathcal{L}_{\text{rgb}}(I_i, I_{\text{gt}}) + \mathcal{L}_{\text{ssim}}(I_i, I_{\text{gt}}) \right],
\end{align}
where $\lambda_{\text{full}} + \lambda_{\text{sub}} = 1$, and we simply set both 
to $0.5$.
We denote the losses for $I_{\text{full}}$ and $I_{\text{sub}}$ as the full loss ($\mathcal{L}_{\text{full}}$) and subset loss ($\mathcal{L}_{\text{sub}}$), respectively.
This uncertainty-intensified rasterization strategy is conceptually inspired by Dropout~\cite{DropGaussian_CVPR2025, srivastava2014dropout} and Hard Negative Mining~\cite{xuan2020hnm, JANG2019_hnm}. Random sampling of $\mathcal{G}_{\text{rand}}$ provides stochastic diversity, allowing weakly contributing Gaussians to receive supervision when dominant ones are excluded. Meanwhile, deterministic inclusion of $\mathcal{G}_{\text{uncertain}}$ ensures consistent gradient flow to persistently under-optimized Gaussians. This dual mechanism reinforces learning in uncertain or ambiguous regions without modifying the rasterization process, while complementing full-image supervision to maintain global photometric fidelity.

\cam{
To see why the subset branch amplifies supervision for these Gaussians, we examine the gradient it induces.
For a pixel ray, the full branch renders
$C_{\mathrm{full}} = \sum_i T_i \alpha_i c_i$ with
$T_i = \prod_{j<i}(1-\alpha_j)$.
In the residual branch, we exclude the Gaussians not selected in $\mathcal{G}_{\text{sub}}$; for mathematical clarity, this is equivalent to masking them with zero opacity in that branch. 
Considering the effect of a single masked Gaussian $k$ in isolation, 
the modified transmittance becomes $\tilde{T}_i = T_i/(1-\alpha_k)$ for every later Gaussian $i > k$. which can be generalized by accumulating one such factor per masked Gaussian ahead of $i$.
As $\tilde{T}_i$ multiplies the
subset gradient with respect to \emph{any} parameter $\theta_i$, the
amplification carries over to the whole parameter set; we illustrate with the
color $c_i$, as the compositing map is linear
($\partial C / \partial c_i = T_i \alpha_i$). 
For $i>k$:
\begin{equation}
  \frac{\partial \mathcal{L}}{\partial c_i}
  = \lambda_{\mathrm{full}}
    \frac{\partial \mathcal{L}_{\mathrm{full}}}{\partial C_{\mathrm{full}}}
    T_i \alpha_i
  + \lambda_{\mathrm{sub}}
    \frac{\partial \mathcal{L}_{\mathrm{sub}}}{\partial C_{\mathrm{sub}}}
    \frac{T_i \alpha_i}{1-\alpha_k}.
  \label{eq:amplified-grad}
\end{equation}
}

\cam{
The amplification factor $1/(1-\alpha_k)$ grows as the Gaussian $k$ becomes more opaque. 
Intuitively, dropping a high-opacity occluder redirects gradient toward previously occluded, weakly contributing Gaussians, the source of our residual supervision.}
By supervising both full and uncertainty-intensified images, we promote stronger gradient flow toward uncertain or low-opacity Gaussians without compromising photometric quality. This strategy mirrors the effect of residual skip connections in ResNet~\cite{He_2016_CVPR} (Fig.~\ref{fig:res_learning}(b)), enabling more stable convergence and reducing overfitting in sparse or wide-baseline training settings. It is particularly effective during early next-best-view selection, when reconstruction is sensitive to both sparsely initialized regions and supervision bias from limited views.

\section{Experiments}
\paragraph{Dataset}
We evaluate our approach on two benchmark datasets: NeRF-Synthetic \cite{nerf} and Mip-NeRF 360~\cite{mip-nerf360}. Although these datasets span scenes from synthetic object-scale setups to real-world outdoor environments with full 360-degree coverage, their uniform, curated camera trajectories pose only a limited challenge for active view selection, as even simple heuristics (e.g., furthest-distance selection) perform reliably under balanced coverage~\cite{Xiao:CVPR24:NeRFDirector}. To address this limitation, we curate an extended benchmark with five diverse scenes from Deep Blending~\cite{deepblending} and Tanks and Temples~\cite{tankandtemples}, which introduce unbalanced view distributions and varied scene scales that better reflect practical conditions. All experiments use images at their original resolutions; for further dataset curation details, please refer to Suppl.~\Sref{Supp:Implementation}.

\paragraph{Competing Methods}
We compare our method quantitatively and qualitatively against several active 3DGS methods that operate solely on RGB images: FisherRF (baseline)~\cite{fisherrf}, ACP~\cite{acp}, and random view selection.
We also include 2D-based view selection methods from the Active View Selector framework~\cite{wang2024avs}, which incorporates two image quality assessment (IQA) models, MUSIQ~\cite{musiq} and CrossScore~\cite{crossscore}, to evaluate perceptual quality.
Both models are re-implemented following the authors' official instructions and publicly available code.
\cam{We additionally compare against earlier methods such as BayesRays~\cite{bayesrays} and ActiveNeRF~\cite{activenerf} adapted to 3DGS, results are reported in Suppl.~\Sref{Supp:extended_baseline}
}

\subsection{Active View Selection}
\label{subsec:5_1}
\paragraph{Experiment setup}
Following the protocol in~\cite{fisherrf}, we adopt the prescribed initial view configurations and view selection schedule.
Specifically, each experiment starts with four uniformly distributed views, then adds one view every 100 epochs until reaching 20 training views (for fewer training views, see Suppl.~\Sref{Supp:Ablation_robust}).
We apply the same active selection strategy across all datasets.
\cam{
For consistency, each model is initialized with the same random seed, trained for 20,000 iterations, and shares the same COLMAP SfM initialization\footnote{SA-Points are used only for view selection and are not included in the COLMAP SfM or 3DGS initialization.}.
}
All other settings remain unchanged across experiments, except for the view selection algorithm.
Because each process includes stochasticity, we repeat all experiments four times and report average scores (for details, see Suppl.~\Sref{Supp:statistics}.).

\begin{table*}[t!]
\centering
\caption{\textbf{Quantitative results for the Active View Selection}. 
We compare our model with (1) Rule-based models (Random, ACP~\cite{acp}), (2) 2D-based models~\cite{chen2024gennbv} (MUSIQ, CrossScore), and 3D-based model(FisherRF~\cite{fisherrf}). Results are averaged over 9 scenes from the Mip-NeRF 360, 7 scenes from NeRF-synthetic dataset and 5 additional scenes from the Deep Blending and Tanks and Temples. We conduct four trials for each scene and report the average. For further statistics please refer to the Suppl.~\Sref{Supp:statistics}.
}
\resizebox{\textwidth}{!}{%
\begin{tabular}{llccccccccc}
\toprule
\multirow{2}{*}{Category} 
& \multirow{2}{*}{Method}
& \multicolumn{3}{c}{NeRF Synthetic}
& \multicolumn{3}{c}{Mip-NeRF 360}
& \multicolumn{3}{c}{Extended Benchmark} \\
\cmidrule(lr){3-5} \cmidrule(lr){6-8} \cmidrule(lr){9-11}
& 
& PSNR$\uparrow$ & SSIM$\uparrow$ & LPIPS$\downarrow$
& PSNR$\uparrow$ & SSIM$\uparrow$ & LPIPS$\downarrow$
& PSNR$\uparrow$ & SSIM$\uparrow$ & LPIPS$\downarrow$ \\
\midrule

\multirow{2}{*}{Rule-based}
& Random
& 24.847 & 0.893 & 0.117
& 19.969 & 0.584 & 0.456
& 19.262 & 0.699 & 0.375 \\
& ACP
& 22.718 & 0.855 & 0.138
& 20.325 & 0.596 & \underline{0.449}
& 19.950 & 0.718 & \underline{0.361} \\

\midrule
\multirow{2}{*}{2D-based}
& MUSIQ
& 25.237 & 0.889 & 0.119
& 19.850 & 0.575 & 0.466
& 18.699 & 0.688 & 0.391 \\
& CrossScore
& 23.746 & 0.868 & 0.130
& 21.076 & \underline{0.612} & \textbf{0.448}
& 19.942 & 0.727 & \textbf{0.356} \\

\midrule
\multirow{3}{*}{3D-based}
& FisherRF
& 25.190 & 0.892 & 0.116
& 20.642 & 0.595 & 0.450
& 19.654 & 0.711 & 0.370 \\
& \cellcolor{lightblue}Ours (MASt3R)
& \cellcolor{lightblue}\textbf{26.580} & \cellcolor{lightblue}\textbf{0.907} & \cellcolor{lightblue}\textbf{0.110} & \cellcolor{lightblue}\textbf{21.410} & \cellcolor{lightblue}\textbf{0.613} & 
\cellcolor{lightblue}0.451 & 
\cellcolor{lightblue}\textbf{20.401} & \cellcolor{lightblue}\underline{0.732} & \cellcolor{lightblue}\underline{0.361} \\
& \cellcolor{lightblue}Ours (DA-v3)
& \cellcolor{lightblue}\underline{26.437}
& \cellcolor{lightblue}\underline{0.905}
& \cellcolor{lightblue}\underline{0.111}
& \cellcolor{lightblue}\underline{21.361}
& \cellcolor{lightblue}\textbf{0.613}
& \cellcolor{lightblue}\underline{0.449}
& \cellcolor{lightblue}\underline{20.348} & \cellcolor{lightblue}\textbf{0.733} & \cellcolor{lightblue}0.365 \\
\bottomrule
\end{tabular}
}
\label{table:quan_active_view}
\end{table*}

\begin{figure*}[t!]
    \centering
    \includegraphics[width=\textwidth]{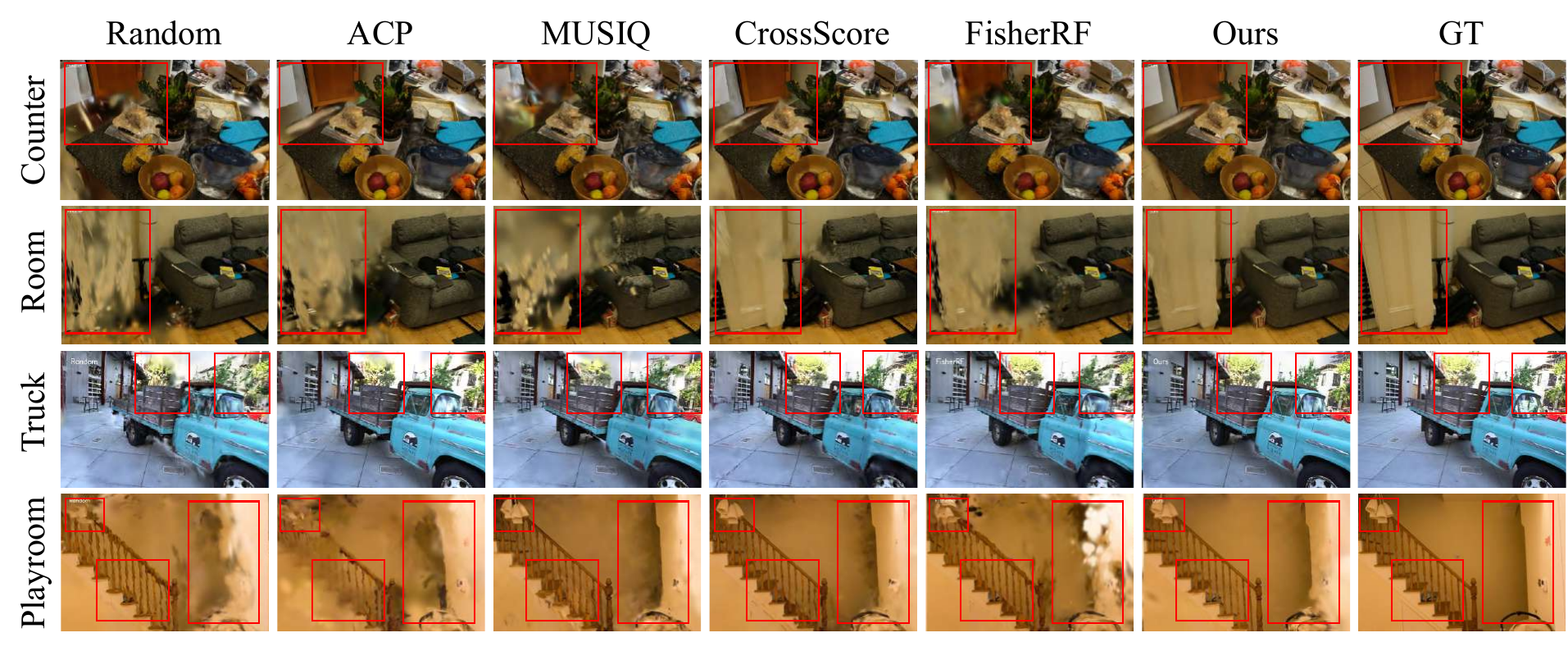}
    \caption{\textbf{Qualitative Comparison of Active View Selection.} Reconstruction from 20 selected views per scene. Our method shows improved completeness and fewer artifacts compared to counterparts models.
    Multi-view visualization and 360° rendering is provided in Suppl.~\Sref{Supp:Additional_qual}
    and Suppl. video.}
    \label{fig:qual}
\end{figure*}
\begin{figure}[h]
    \centering
    \includegraphics[width=1.0\linewidth]{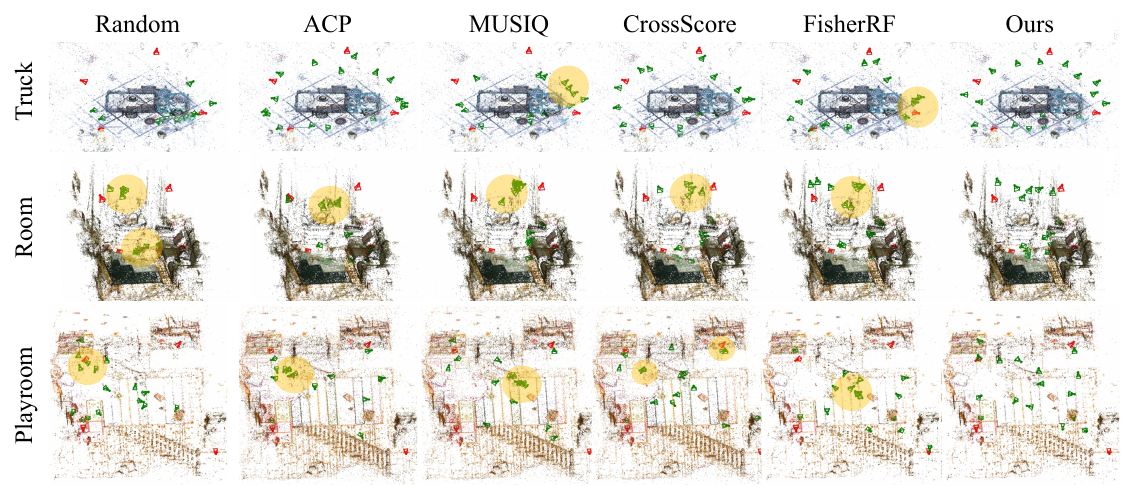}
    \caption{
    \textbf{Selected Camera View Distribution.} 
    Visualization of camera poses selected by each method on \textit{Truck, Playroom} scenes from Extended dataset, and \textit{Room} scene from Mip-NeRF 360 dataset.
    Red frustums indicate the initial views, while green frustums denote views added during active selection. 
    Our method produces a more uniformly distributed set of viewpoints, while counterpart models exhibit cluttering (highlighted by yellow circles),
    due to its reliance on uncertainty signals entangled with 3DGS training dynamics. 
    For more visualization please refer to Suppl.~\Sref{Supp:Camera_pose}
    }
    \label{fig:cam_pose_sample}
\end{figure}
\paragraph{Novel View Synthesis Results}
Quantitative and qualitative results on NeRF-Synthetic, Mip-NeRF 360, and the extended benchmark (Deep Blending and Tanks and Temples) are summarized in Table~\ref{table:quan_active_view} and Fig.~\ref{fig:qual}.
Competing methods show limited 3D reconstruction performance, especially in sparsely observed regions, primarily due to biased view selection and overfitting induced by vanishing gradients.
This leads to incomplete reconstructions, characterized by floating artifacts and missing geometry (\ie holes and missing objects).
\kjs{
In contrast, our method achieves the highest PSNR and SSIM on NeRF-Synthetic and Mip-NeRF 360 while maintaining comparable LPIPS performance.
The smaller LPIPS gains likely stem from smoother color reconstruction in uncertain regions, consistent with trends reported in~\cite{jang2025comapgs}. As also reflected in Table~\ref{table:quan_active_view}, variants using DA-v3 as the 3D lifting module show consistent gains over all competing methods, indicating our improvements are not tied to specific backbone.
}

Consistent with the trends above, results in Table~\ref{table:quan_active_view} on the extended datasets further validate the generalizability of our method.
In particular, Deep Blending results show robustness under diverse, real-world-like camera distributions.
Likewise, numerical improvements on Tanks and Temples indicate stronger scene coverage in large outdoor settings, as exemplified by the Truck scene in Fig.~\ref{fig:qual}.
\kjs{Even under challenging, realistic view configurations, our model remains superior in reconstruction quality and scene coverage while preserving comparable high-frequency details.
For multi-view evaluation, we provide visualizations and 360° rendering results in Suppl.~\Sref{Supp:Additional_qual}
and the Suppl.~video.
}

\paragraph{Camera Distribution Results}
\kjs{
To further support these quantitative and qualitative trends, we analyze the camera view distributions selected by different methods.
As illustrated in~\Fref{fig:cam_pose_sample}, competing methods often select clustered or redundant viewpoints, leading to uneven scene coverage.
This behavior is especially evident in FisherRF, whose selections concentrate around high-response regions due to its tight coupling with 3DGS training dynamics.
In contrast, our method selects more spatially dispersed and geometrically diverse viewpoints.
Together, these observations show that our physically grounded view selection and uncertainty-aware learning strategy improves reconstruction fidelity and coverage consistency.
For more camera-distribution results 
please refer to 
Suppl.~\Sref{Supp:Camera_pose}
}



\subsection{Comparison on Uncertainty Estimation}
\label{subsec:uncertainty_est}
\label{subsec:4_2}
\paragraph{Experiment setup}
We evaluate how effectively our method improves uncertainty estimation quality.
Specifically, we examine whether residual loss and self-augmented pre-filtering improve alignment between depth errors and predicted uncertainties under controlled conditions.
To measure, we adopt the Area Under the Sparsification Error (AUSE) metric, to evaluate uncertainty calibration adopted in~\cite{fisherrf, bayesrays, cfnerf}.
\kjs{
AUSE evaluates how accurately the high-magnitude regions of the uncertainty heatmap correspond to the largest actual depth errors.
}

Following CF-NeRF~\cite{cfnerf}, we use depth maps from NerfingMVS~\cite{nerfingmvs}, optimized at test time with stereo depth from COLMAP. Experiments are conducted on all nine Mip-NeRF 360 scenes under an identical view-selection schedule and evaluated on all test views.

\begin{figure}[t]
    \centering
    \includegraphics[width=\linewidth]{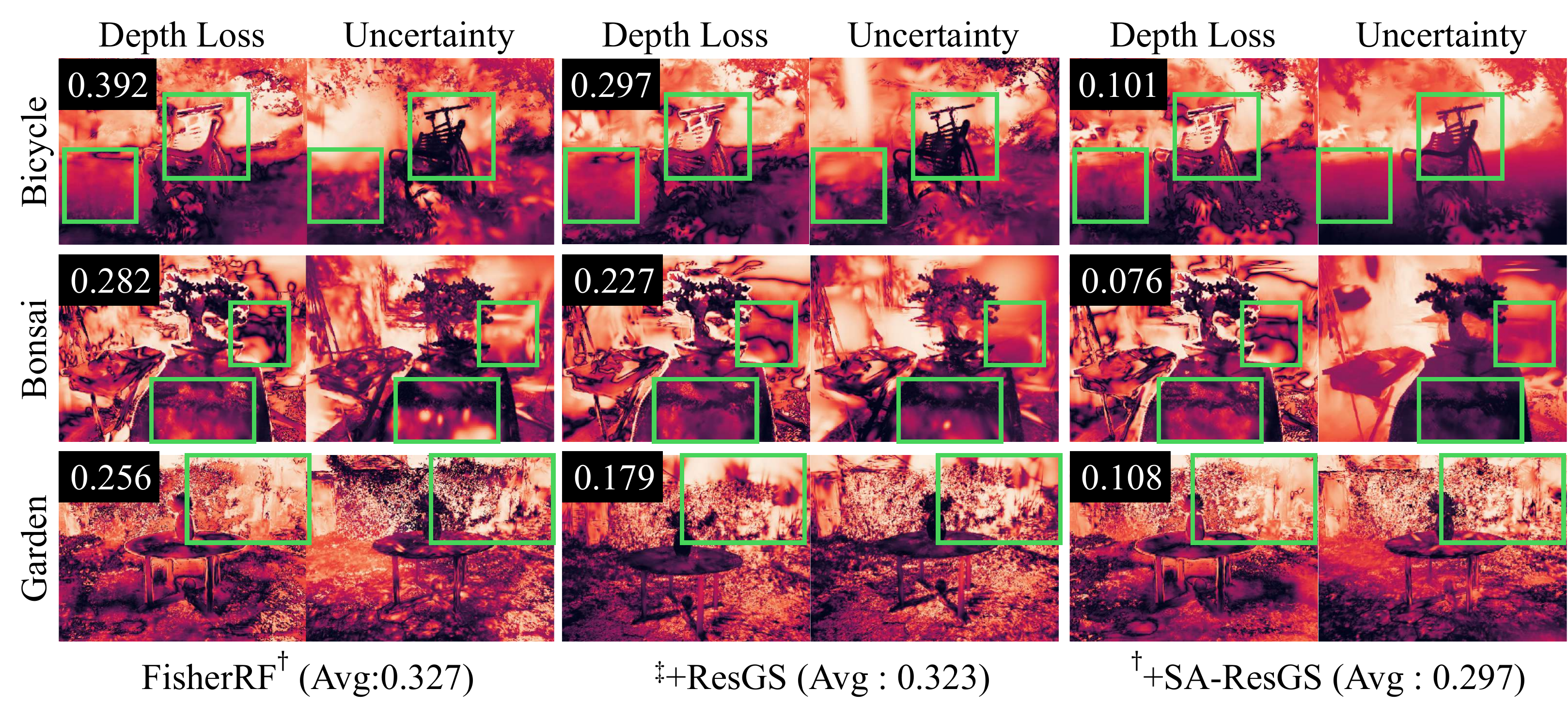}
    \caption{\textbf{Uncertainty Comparison.} 
    We visualize depth loss with their corresponding uncertainty with AUSE metric (lower is better). Highlighted green boxes show that $^{\dagger}$+SA-ResGS yields better alignment between large depth errors and high uncertainty. \kjs{Please note that $^{\ddagger}$ denotes fixed-order view selection, replicating the original selection sequence from FisherRF$^{\dagger}$, whereas $^{\dagger}$ indicates dynamically updated view selection.}
    }
    \label{fig:ause}
\end{figure}
\paragraph{Results}
\kjs{
Starting from the baseline (FisherRF$^{\dagger}$), adding residual learning on a fixed baseline view sequence ($^{\ddagger}$+ResGS) reduces AUSE from 0.327 to 0.323.
Adding prefiltering with SA-Points ($^{\ddagger}$+SA-ResGS) further reduces AUSE to 0.297.
}
These reductions indicate that both residual learning and our prefiltering improve uncertainty calibration.
\kjs{
Fig.~\ref{fig:ause} highlights representative miscalibration cases, including overconfident predictions in high-error regions and conservative uncertainty in low-error areas.
Compared with the baseline, our method reduces these ranking mismatches and aligns predicted uncertainty more closely with ground-truth depth error.
}

\kjs{
We attribute these gains to the complementary roles of residual learning and our physically grounded prefiltering strategy.
Residual learning stabilizes confidence in low-error regions while preserving adaptive refinement in uncertain areas via skip connections.
Meanwhile, prefiltering promotes more balanced spatial coverage during training, reducing localized miscalibration and improving global uncertainty–error consistency.
}
Consequently, our model achieves better structural and quantitative uncertainty calibration, improving accuracy in uncertainty-driven tasks such as active mapping.

\subsection{Ablation Studies}
\label{subsec:4_3}
In the following section, we present ablation studies that analyze (1) the effect of individual components and (2) the effect of the full loss.
Additional ablations are provided in Supplementary material. 
\cam{Specifically, we analyze our physically grounded view selection algorithm in Suppl.~\Sref{Supp:ablation_prefiltering}
and the uncertainty-guided residual learning in Suppl.~\Sref{Supp:ablation_residual}
.
In Suppl.~\Sref{Supp:Ablation_robust}
, we further study robustness to correspondence noise, the hash-encoding size, and selected view-count.
}

\begin{table}[!t]
    \centering
    \caption{\textbf{Ablation study on Mip-NeRF 360 and Extended dataset.} $^{\ddagger}$ denotes fixed-order view selection, 
    whereas $^{\dagger}$ indicates dynamically updated view selection.
    \kjs{We conduct four trials for each scene and report the average. Visual comparison is included in Suppl.~\Sref{Supp:Ablation_Qual}
    , and further statistics in Suppl.~\Sref{Supp:statistics}
    }} 
    \label{table:ablation}    
    \resizebox{\linewidth}{!}{
    \begin{tabular}{l c c c c c c c c c c}
    \toprule
    \multirow{2}{*}{\textbf{Methods}} & \multirow{2}{*}{\textbf{3D Lifting}} & \multicolumn{2}{c}{\textbf{Our proposed methods}} & \multicolumn{3}{c}{\textbf{Mip-NeRF 360}} & \multicolumn{3}{c}{\textbf{Extended dataset}}\\
    \cmidrule(lr){3-4}
    \cmidrule(lr){5-7}
    \cmidrule(lr){8-10}
      & & Sec.~\ref{subsec:3_2} & Sec.~\ref{subsec:3_3} & PSNR$\uparrow$ & SSIM$\uparrow$ & LPIPS$\downarrow$ & PSNR$\uparrow$ & SSIM$\uparrow$ & LPIPS$\downarrow$ \\
    \midrule
    FisherRF$^{\dagger}$ & - & - & - 
    & {20.642} & {0.595} & {0.450}
    & {19.490} & {0.708} & {0.373} \\
    \midrule
    $^{\ddagger}$+ResGS & - & - & $\checkmark$ 
    & {21.045} & {0.604} & {0.453}
    & {19.602} & {0.709} & {0.378} \\
    \midrule
    $^{\dagger}$+ResGS & - & - & $\checkmark$ 
    & {20.732} & {0.594} & {0.461}
    & {19.791} & {0.715} & {0.374} \\
    \midrule
    $^{\dagger}$+SA-HashGS & MASt3R & $\checkmark$ & - 
    & {21.093} & \underline{0.609} & \textbf{0.441}
    & {19.702} & {0.718} & {0.363} \\
    $^{\dagger}$+SA-ResGS & MASt3R & $\checkmark$  & $\checkmark$  
    & \textbf{21.410} & \textbf{0.613} & {0.451}
    & \textbf{20.401} & \underline{0.732} & \underline{0.361} \\
    \midrule
    $^{\dagger}$+SA-HashGS & DA-v3 & $\checkmark$ & - 
    & {21.076} & \underline{0.609} & \textbf{0.441}
    & {20.025} & {0.726} & \textbf{0.357} \\
    $^{\dagger}$+SA-ResGS & DA-v3 & $\checkmark$  & $\checkmark$  
    & \underline{21.361} & \textbf{0.613} & \underline{0.449}
    & \underline{20.348} & \textbf{0.733} & {0.365} \\
    \bottomrule
    \end{tabular}}
\end{table}

\paragraph{Effect of individual components}
\label{paragraph:components}
\kjs{
The ablation study in \Tref{table:ablation} highlights the individual and synergistic contributions of our proposed SA-ResGS framework.
While residual learning (ResGS) alone improves reconstruction quality under a fixed baseline view sequence ($^\ddagger$+ResGS), 
its gains are attenuated when combined with purely dynamic selection ($^\dagger$+ResGS).
}
This indicates that training improvements alone are insufficient, especially under large uncertainty quantification errors.
As shown in Sec.~\ref{subsec:uncertainty_est}, our training module effectively aligns predicted uncertainty with actual errors, yet purely uncertainty-driven view selection remains vulnerable to biases from internal learning dynamics.

\kjs{
Our physically grounded prefiltering module ($^\dagger$+SA-HashGS) stabilizes view selection by restricting the candidate set for Fisher uncertainty.
The full configuration ($^\dagger$+SA-ResGS) achieves the best performance, demonstrating a synergy in which SA-HashGS provides a reliable view sequence that enables residual supervision to further improve reconstruction in sparse or ambiguous regions. 
The same trend appears when replacing the 3D lifting module with DA-v3: prefiltering consistently stabilizes view selection, and adding ResGS yields further gains.
}

\begin{figure*}[t!]
    \centering
    \includegraphics[width=\linewidth]{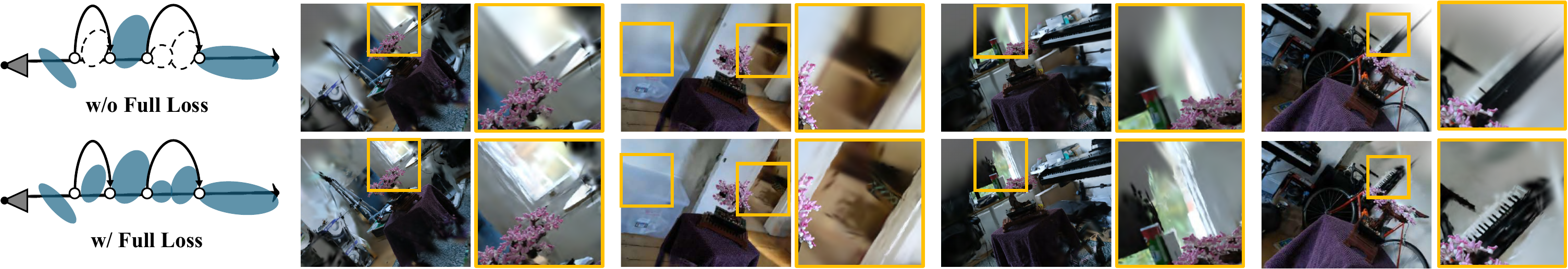}
    \caption{
    \textbf{Comparison of novel-view renderings with/without the Full Loss.} 
    We use a pre-selected view order from FisherRF$^\dagger$ to ignore view-selection effects.
    The w/o Full Loss ($\mathcal{L}_{sub}$ only) variant exhibits smoothing artifacts, whereas w/ Full Loss ($\mathcal{L}_{full} + \mathcal{L}_{sub}$) preserves scene structure and high-frequency details. }
    \label{fig:full_loss}
\end{figure*}

\paragraph{Effect of the full loss}
\label{paragraph:full_loss}
To assess the contribution of the Full Loss, we compare ResGS with a w/o Full Loss variant (Fig.~\ref{fig:full_loss}), where Gaussians are updated only through the guided subset term ($\mathcal{L}_{sub}$) and the default w/ Full Loss setting ($\mathcal{L}_{full}$+$\mathcal{L}_{sub}$).
The w/o Full Loss model tends to oversmooth low-confidence regions, causing a loss of high-frequency detail. 
Without $\mathcal{L}_{full}$ reinforcing these regions, the subset-only variant cannot preserve or reactivate Gaussians that require continued refinement. 
In contrast, w/ Full Loss
prevents under-updated Gaussians from collapsing and preserves both global structure and fine details.
For an ablation on uncertainty-guided sampling, see Suppl.~\Sref{Supp:Ablation_Qual}
, Fig.~S10.


\subsection{Computation Efficiency Analysis}
\label{subsec:4_4}
A key challenge in active view selection is computational cost, as FisherRF computes per-Gaussian Fisher information via backpropagation across all candidate views, creating a bottleneck in large-scale datasets. 
To evaluate this overhead, we conduct a runtime analysis on the \textit{Bonsai} scene using a mid-range GPU (38 TFLOPS fp32), as summarized in Table~\ref{table:temporal_analysis}. 
SA-ResGS replaces exhaustive Fisher evaluation with a four-step process: dense correspondence prediction (MASt3R), SA-Points triangulation (Triang.), physically grounded prefiltering (Prefilter), and Fisher computation on filtered views (Fisher). 
Despite these additional steps, view selection is 55\% faster, with only a modest increase in per-iteration cost. 
End-to-end runtime
is reduced by 40\%.
GPU memory usage rises slightly but remains within standard limits, 
supporting scalability and practicality.

\begin{table}[t]
    \centering
    \renewcommand{\arraystretch}{0.9}
    \caption{
    \textbf{Runtime comparison.} 
    Breakdown of 
    temporal training costs in the active view selection on \textit{Bonsai} scene. SA-ResGS introduces additional prefiltering steps but reduces total runtime by 40\%, with modest increases in 
    GPU memory usage.}
    \resizebox{0.9\linewidth}{!}{ 
    \begin{tabular}{c cccccccc}
    \toprule
    \multirow{2}{*}{\textbf{Method}} & \multicolumn{5}{c}{\textbf{Active View Selection} [s]} & \multirow{2}{30pt}{\centering{\textbf{Raster.} [s/iter]}} & \multirow{2}{65pt}{\centering{\textbf{End-to-end}}} & \multirow{2}{25pt}{\centering{\textbf{GPU} [K]}} \\
    \cmidrule(lr){2-6}
    & MASt3R & Triang. & Prefilter & Fisher  & \textbf{Total}   & & &  \\
    \midrule
    FisherRF             & —      & —       & —         & 28.00 & 28.00 & 0.005         & 32 m 59 s  & 8.0 \\
    Ours                 & 0.40 & 1.49  & 5.00      & 5.60  & 12.50 & 0.027         & 19 m 45 s & 10.5 \\
    \bottomrule
    \end{tabular}
    }
    \label{table:3d_loc}    
    \label{table:temporal_analysis}
\end{table}
\section{Conclusion}

This paper presents SA-ResGS, a framework that stabilizes uncertainty quantification and enhances uncertainty-aware supervision for next-best-view selection in active scene reconstruction.
We introduce Self-Augmented Points, reconstructed from a training view and a rasterized extrapolated view.
These points enable physically grounded view selection and help mitigate bias in uncertainty estimates. Furthermore, we propose the first residual learning strategy tailored to 3D Gaussian Splatting, with an emulated skip connection, enabling effective supervision for both uncertain image regions and weakly contributing Gaussian splats.
This leads to improved photometric reconstruction in novel view synthesis.
Extensive experiments for NBV selection and novel view synthesis demonstrate the effectiveness of SA-ResGS across a range of realistic scenes.

%
%
\vfill
{\scriptsize\paragraph{Acknowledgements}
This work was partially supported by Institute of Information \& Communications Technology Planning \& Evaluation (IITP) grant  (No. RS-2026-25518317, Development of AI memory mechanism that reflects human cognitive principles), the National Research Foundation of Korea (NRF) grant (No. RS-2024-00451947; No. RS-2024-00453301), and the InnoCORE program of the Ministry of Science and ICT(26-InnoCORE-01) funded by the Korea government (MSIT).
\par}

\bibliographystyle{splncs04}
\bibliography{main}

\title{Supplementary Material for ``SA-ResGS: Self-Augmented Residual 3D Gaussian Splatting for Next Best View Selection''} 

\titlerunning{SA-ResGS}

\author{Kim Jun-Seong\inst{1}\thanks{This work was conducted while authors were with Huawei Noah’s Ark Lab in London}\orcidlink{0000-0001-7570-6508} \quad
Tae-Hyun Oh\inst{2}\orcidlink{0000-0003-0468-1571} \\
Eduardo P\'erez-Pellitero\inst{3}\orcidlink{0000-0001-9096-4740} \quad
Youngkyoon Jang\inst{4\ast}\thanks{denotes the corresponding author}\orcidlink{0000-0002-0068-003X}
}

\authorrunning{K.~Jun-Seong et al.}

\institute{
{$^{1}$POSTECH} \quad
{$^{2}$KAIST}  \quad
{$^{3}$Huawei Noah's Ark Lab}  \quad
{$^{4}$Rivian} 
\email{junseong.kim@postech.ac.kr}\\
\url{https://saresgs.github.io/}
}

\maketitle

\appendix

\setcounter{figure}{0}
\setcounter{table}{0}
\renewcommand{\thefigure}{S\arabic{figure}}
\renewcommand{\thetable}{S\arabic{table}}
\renewcommand{\theHsection}{supp.\arabic{section}}

\pagestyle{headings}
\def\ECCVSubNumber{3703}  

\section*{Supplementary Material Overview}

This supplementary document provides additional implementation details and results that support and extend the main paper. It is organized as follows:
 \begin{itemize}
    \item \Sref{Supp:Implementation}. Implementation Details
    \item \Sref{Supp:Camera_pose}. Camera Distribution Analysis.
    \item \Sref{Supp:Additional_qual}. Extended Qualitative Comparisons.
    \item \Sref{Supp:Ablation_Qual}. Extended Qualitative Comparisons on Ablation Studies.
    \item \Sref{Supp:ablation_prefiltering}. Ablation study - Physically Grounded View Selection
    \item \Sref{Supp:ablation_residual}. Ablation study - Uncertainty-guided Residual Learning
    \item \Sref{Supp:Ablation_robust}. \rev{Robustness and Sensitivity Analysis}
    \item \Sref{Supp:extended_baseline}. Extended Baseline Comparison
    \item \Sref{Supp:statistics}. The Evaluation Statistics 
    \item \Sref{Supp:supple_video_overview}. Supplementary Video Overview.
\end{itemize}

\section{Implementation Details}\label{Sec:A}
\rev{
In this section, we provide additional details on both the implementation of our method and the experiment setup used throughout the paper.}
\cam{We first recap the FisherRF~\cite{fisherrf} uncertainty and expected-information-gain formulation we adopt in uncertainty estimation (Sec.~\ref{Supp:A_0}).
}
\rev{
Sec.~\ref{Supp:A_1}~to~\ref{Supp:A_4} describe the implementation details of the proposed physically grounded prefiltering, including the SA-Points generation (Sec.~3.1) and coverage estimation and encoding (Sec.~3.2).
}
\rev{
Sec.~\ref{Supp:A_5}~and~\ref{Supp:A_6} then present the experiment setup and dataset details, respectively.
}

\label{Supp:Implementation}
\subsection{Adopted Uncertainty Formulation (FisherRF Recap)}
\label{Supp:A_0}
\cam{For uncertainty estimation we adopt the mathematical formulation and code implementation suggested in  FisherRF~\cite{fisherrf}.
Our method does not modify uncertainty estimation part, but restricts the candidate set over which it is evaluated (Sec.~3.2) considering physical geometry, and adds residual supervision (Sec.~3.3).} 

\cam{Let $\theta$ denote the 3DGS parameters and let $\mathcal{D}$ be the set of all rays cast from the training views observed so far. 
FisherRF approximates the parameter posterior $p(\theta\mid\mathcal{D})$ by a Gaussian centered at the trained parameters~\ie Laplacian approximation.
The precision, inverse covariance, of this Gaussian is denoted $\mathbf{H}$ and measures how strongly the observations constrain $\theta$.
It is computed as the Gauss--Newton Hessian of the rendering loss, summed over all observed rays:}
\begin{align}
\mathbf{H} \;=\; \sum_{r \in \mathcal{D}}
\Big(\tfrac{\partial \hat{C}_r}{\partial \theta}\Big)^{\!\top}
\Lambda_r
\Big(\tfrac{\partial \hat{C}_r}{\partial \theta}\Big) \;+\; \epsilon \mathbf{I}.
\end{align}

\cam{Here $\hat{C}_r$ is the rendered color of ray $r$, and $\partial \hat{C}_r/\partial\theta$ is its sensitivity to the parameters. 
The weight $\Lambda_r$ is the confidence of each color observation; following FisherRF, all rays share the same isotropic value, so Eq.~(1) of~\cite{fisherrf} reduces to an unweighted squared-error loss. 
The term $\epsilon\mathbf{I}$ is a small prior added by the Laplace approximation that keeps $\mathbf{H}$ invertible. 
Inverting $\mathbf{H}$ gives the parameter covariance, and propagating it through the rendering Jacobian yields the predictive uncertainty of a view, $\mathcal{U} = (\partial \hat{C}/\partial\theta)\,\mathbf{H}^{-1}\,(\partial \hat{C}/\partial\theta)^{\top}$.
For tractability, FisherRF does not form $\mathbf{H}$ in full but keeps only its diagonal, so that the inverse below reduces to an element-wise reciprocal. 
}

\cam{Consider a view $v$ that has not yet been captured~\ie candidate view. 
If we capture it, it would contribute its own information, $\mathbf{H}_v$, computed exactly like $\mathbf{H}$ but using only the rays of view $v$. 
The expected information gain measures how much capturing $v$ would reduce the uncertainty we currently hold in $\mathbf{H}$:}
\begin{align}
\mathrm{EIG}(v) \;=\; \tfrac{1}{2}\,
\log\det\!\big(\mathbf{I} + \mathbf{H}_v\,\mathbf{H}^{-1}\big),
\end{align}
\cam{and the next-best view is $v^\star = \arg\max_{v}\mathrm{EIG}(v)$ (cf.~Eq.~(3) of~\cite{fisherrf}). 
In SA-ResGS, this acquisition is evaluated only within the physically grounded candidate subset as illustrated in manuscript. 
We refer to~\cite{fisherrf} for the complete derivation.}


\subsection{Overview of Physically-grounded View Selection}
\label{Supp:A_1}
Physically grounded view selection begins by constructing Self-Augmented Points (SA-Points) via 3D lifting between a training view and its extrapolated view. 
These SA-Points are used to estimate surface occupancy and to encode observed geometry into binary voxel features. 
Candidate views are scored based on their voxel-level dissimilarity to the encoded training views, supporting robust coverage estimation without relying on early uncertainty signals. 
In the following, we detail (1) extrapolated view generation, (2) Self-Augmented Point (SA-Point) generation using MASt3R~\cite{mast3r} and triangulation, (3) SA-Point generation using Depth-Anything-v3~\cite{da3} (DA-v3), and (4) coverage estimation and view frustum construction.

\subsection{Extrapolated View Generation}
\label{Supp:A_2}
\noindent To synthesize a novel rasterized view while maintaining sufficient scene overlap, we perturb the original camera center by $\pm$0.25 units along the x and y axes and translate it backward by 0.5 units along the z-axis. 
This backward-only perturbation ensures that the extrapolated view retains a high degree of visibility overlap with the original training view, keeping most scene content within the shared frustums. 
As a result, the extrapolated view covers a large portion of the original training image while still providing a novel perspective of the same surfaces, enabling reliable correspondence estimation for SA-Point generation.

\subsection{Self-Augmented Points (SA-Points) Generation}
\label{Supp:A_3}
\rev{
While our framework is modular and compatible with other 3D foundational reconstruction models, the manuscript instantiates the pipeline with MASt3R followed by triangulation. 
In this section, we first describe the default MASt3R-based SA-Points generation procedure used throughout our main experiments.
}

\noindent \textbf{Dense Correspondence Matching.}
We compute dense correspondences between the ground-truth image of the original training view and the rendering from the extrapolated view using the pretrained MASt3R model~\cite{mast3r} (\textit{MASt3R ViTLarge$\_$BaseDecoder$\_$512$\_$catmlpdpt$\_$metric}), which is robust to moderate viewpoint perturbations. 
As discussed in the manuscript, MASt3R produces context-aware dense correspondences by capturing structured scene semantics over 16×16 local patches.
This property makes it robust to moderate viewpoint perturbations, even when the extrapolated rendering contains geometric artifacts. 
As a result, we can extract reliable matches despite local distortions in the synthesized view.
This enables consistent 3D reconstruction from single-view observations augmented with extrapolated viewpoints.

\noindent \textbf{Triangulation and Reprojection Filtering.}  
To ensure geometric consistency, we triangulate 3D points from matched correspondences and filter them based on reprojection error.
Specifically, we discard points whose bidirectional reprojection error exceeds 1-pixel. 
For computational efficiency, we parallelize the triangulation process across multiple CPU threads and subsample correspondence pairs using a spatial stride of 5 pixels along both the x and y directions. 
These filtering strategies significantly reduce computation while preserving high-fidelity geometric structure.
\rev{For analysis of reprojection filtering threshold, see Sec.~\ref{Supp:Ablation_robust}.
}

\subsection{Depth-Anything-v3 Implementation}
\label{Supp:A_4}
\rev{
To investigate the modularity of our framework, we replace MASt3R with Depth-Anything-v3~\cite{da3} (DA-v3), a more recent feed-forward model.
This alternative instantiation preserves the overall physically grounded prefiltering pipeline while requiring several additional implementation choices. 
In the following, we describe the DA-v3-based 3D lifting procedure used in our experiments.
}

\begin{figure}[h]
    \centering
    \includegraphics[width=0.7\linewidth]{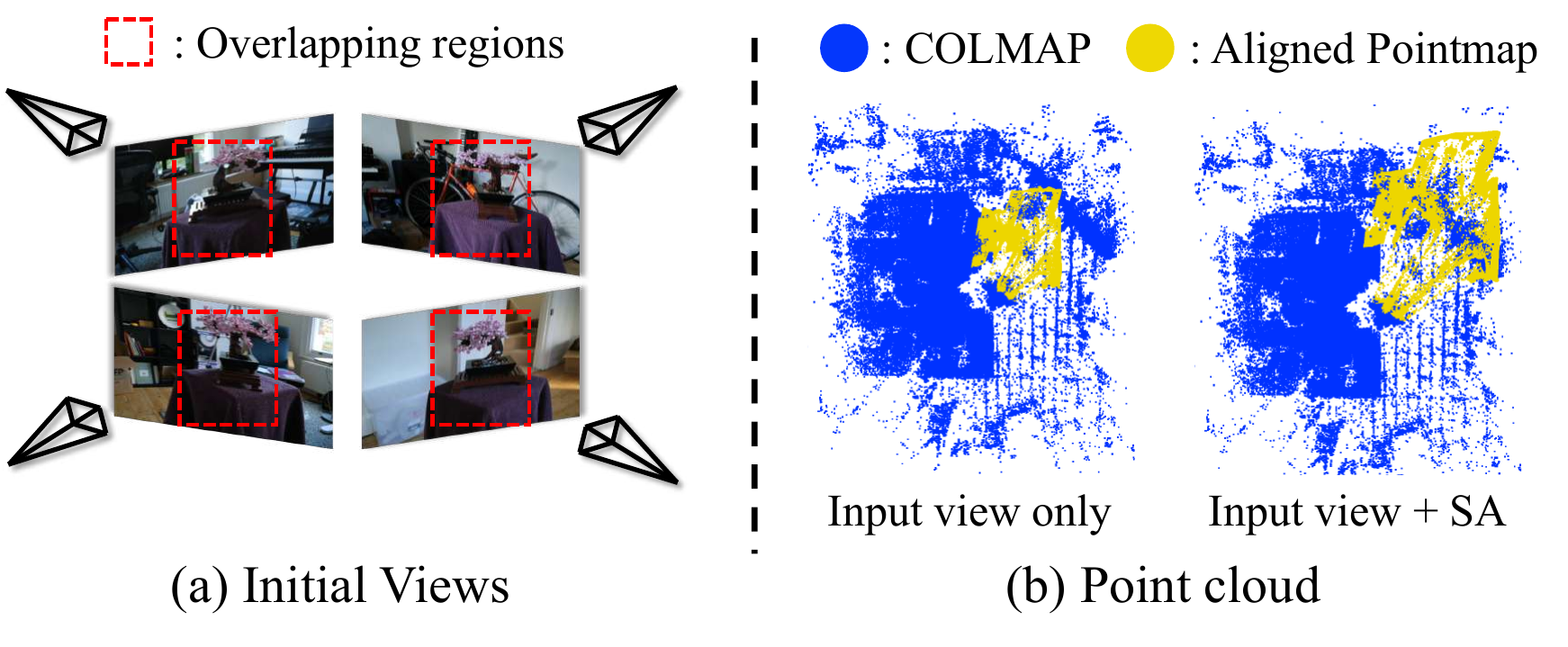}
    \caption{\textbf{Effect of self-augmentation on DA-v3-based lifting and alignment.} 
    (a) Initial input-view images used in our experiments. 
    In our experimental setting, the initial input views share only small foreground regions, while most distant background regions lack mutual observation.
    (b) Point clouds obtained from DA-v3 predictions, visualized after aligning the predicted camera extrinsics to the COLMAP reference frame via Umeyama alignment.
    \textbf{Without self-augmentation ((b)-left)}, 3D lifting relies only on the original sparse training views, in which only small regions are commonly observed.
    As a result, most image regions provide limited multi-view supervision. 
    Because shared geometric evidence is limited, the observed image evidence can often be explained by multiple intrinsic–extrinsic parameter combinations, making camera estimation ambiguous and weakly constrained.
    \textbf{With self-augmentation ((b)-right)}, in contrast, providing the same model with an extrapolated (self-augmented) view, which is designed to capture both the foreground and broader background geometry, imposes stricter multi-view constraints. 
    The resulting multi-view constraints between the training view and its extrapolated counterpart reduce ambiguity in camera parameter estimation, constrain the solution to more geometrically plausible configurations, and in turn lead to more stable Umeyama alignment.
    }
    \label{supp_fig:da3}
\end{figure}

\rev{
\paragraph{DA-v3-based 3D Lifting}
Unlike the default MASt3R-based pipeline, DA-v3 directly predicts 3D geometry together with per-point confidence.
Since the predicted depth points are defined in an independent coordinate system, we align them to the COLMAP~\cite{colmap-0, colmap-1} reference frame using the Umeyama algorithm~\cite{umeyama}, following the original DA-v3 pipeline.
We found that self-augmentation renders remain beneficial in this sparse-view setting.
We therefore also incorporate extrapolated views for DA-v3-based lifting (see Fig.~\ref{supp_fig:da3}).
Unlike MASt3R, DA-v3 can process multiple images jointly. Our preliminary experiments showed that using all eight extrapolated views, \ie all possible combinations of perturbation along the x- and y-axes, provides the best configuration, as shown in~\Tref{table:backbone_ablation}.
Accordingly, we jointly use all eight extrapolated views together with the training view.
We use a pretrained DA-v3 model (\textit{DA3NESTED-GIANT-LARGE}), and filter out depth points with confidence below 1.0 to retain reliable points.
}

\begin{table}[h]
    \centering
    \caption{\textbf{Effect of the number of extrapolated views in the DA-v3 backbone for SA-Point generation.} 
    Comparison between the default MASt3R-based implementation and the DA-v3-based lifting variant under the active-view reconstruction setting on Mip-NeRF 360.
    For DA-v3, we evaluate configurations with 2, 4, and 8 extrapolated views, since the model can process multiple images jointly.
    Note that the Umeyama method fails when only a single extrapolated view is used, as indicated by \xmark~ in the table.
    The DA-v3 model with eight extrapolated views yields the best performance.
}
    \label{table:backbone_ablation}
    \resizebox{0.6\linewidth}{!}{
    \begin{tabular}{ccccc}
        \toprule
        \textbf{Backbone} & \textbf{Extrapolated views} &\textbf{PSNR} $\uparrow$ & \textbf{SSIM} $\uparrow$ & \textbf{LPIPS} $\downarrow$ \\
        \midrule
        MASt3R (Ours)     & 1 & $21.325$ & $0.610$ & $0.450$ \\
        Depth-Anything-v3 & 1 & \xmark & \xmark & \xmark \\        
        Depth-Anything-v3 & 2 & $21.247$ & $0.612$ & $0.449$ \\        
        Depth-Anything-v3 & 4 & $21.254$ & $0.612$ & $0.450$ \\
        Depth-Anything-v3 & 8 & $21.429$ & $0.614$ & $0.448$ \\
        \bottomrule
    \end{tabular}
    \vspace{-7mm}
    }
\end{table}

\subsection{Coverage Estimation and View Frustum Construction}

\noindent \textbf{Voxel Grid Construction.} Given sparse SfM points, we define an axis-aligned bounding box (AABB) that covers the entire 3D point cloud. The AABB is computed from the minimum and maximum bounds of the reconstructed points.

\noindent \textbf{Initial Occupancy Estimation.} We discretize the scene into a voxel grid and mark a voxel as occupied if it contains a minimum number of SfM points (2 for outdoor scenes, and 5 for indoor cases). To better represent scene geometry, we apply $N$-fold upsampling to the occupied voxels.
    
\noindent \textbf{Observed Region Calculation.} SA-Points are mapped to their nearest voxels to define the observed surface region. To account for possible triangulation errors and improve spatial robustness, we apply a 3D dilation operation to the occupied voxels. For all cases, a dilation radius of 2 is applied.

\noindent \textbf{View Frustum Determination.} To evaluate candidate views, we define view frustums using camera intrinsics and the global maximum bounds computed from the SfM point cloud. These frustums ignore visibility constraints but serve as a conservative estimate of potential scene coverage. Coverage scores for physically grounded view selection are then computed by measuring voxel-level intersections between the candidate frustums and the observed surface, represented via a hash-encoded voxel occupancy representation.

\subsection{Experiment Setup}
\label{Supp:A_5}
All models presented in the manuscript, including the ablation variants of our proposed method, are trained and evaluated on a single 
\kjs{
mid-range GPU (38 TFLOPS fp32).
}
CPU-based components, \ie SA-Points generation, voxel grid processing, and triangulation, are executed across eight threads for efficiency.

\subsection{Datasets}
\label{Supp:A_6}
We evaluate our method and baselines on three types of datasets: (1) Mip-NeRF 360~\cite{mip-nerf360}, (2) NeRF-Synthetic~\cite{nerf} and (3) Extended next-best-view (NBV) benchmark datasets~\cite{deepblending, tankandtemples}. 
Mip-NeRF 360 consists of nine real-world scenes with dense 360-degree camera coverage, captured across both indoor and outdoor scenarios. 
The NeRF-Synthetic dataset consists of eight object-centric scenes with dense 360-degree camera view distributions, rendered in Blender. 
To ensure consistency across datasets, we generated sparse point clouds for all NeRF-Synthetic scenes using COLMAP. 
We note that the \textit{Ficus} scene fails to reconstruct reliably; therefore, we excluded it and evaluated on the remaining scenes.

While these datasets provide a controlled and well-curated benchmark, their uniform view distribution limits the difficulty of NBV evaluation. 
Such settings often make NBV strategies appear less critical, since even simple heuristics (e.g., furthest-distance selection) can perform reliably under balanced coverage~\cite{Xiao:CVPR24:NeRFDirector}.
To address this issue, we additionally construct an Extended NBV benchmark by selecting five challenging scenes from Deep Blending~\cite{deepblending} and Tanks and Temples~\cite{tankandtemples}, characterized by irregular camera trajectories and diverse scene scales.
This curated set introduces more realistic and unbalanced conditions, offering a complementary testbed for evaluating robustness in active view selection.

These datasets contain large-scale and geometrically complex scenes with irregular camera distributions, but their difficulty also means that many methods fail outright. 
To filter such degenerate cases, we trained the FisherRF baseline under the standard scheme and retained only scenes where it achieved at least 17 dB PSNR, ensuring that the comparisons remained fair and informative.
This procedure produced the following representative scenes: \textit{Horse}, \textit{Truck}, 
\textit{Ballroom}, 
\textit{Ponche}, and \textit{Playroom}. 
The selected set spans both indoor and outdoor environments and includes highly complex camera distributions (e.g., \textit{Ballroom}, \textit{Ponche}, and \textit{Playroom}) that deviate substantially from the curated coverage of Mip-NeRF 360. 
These characteristics create more realistic stress tests for NBV strategies by introducing occlusions, scale variations, and unbalanced observations. 
\rev{
In this section, we provide additional details on both the implementation of our method and the experiment setup used throughout the paper.}
\cam{We first recap the FisherRF~\cite{fisherrf} uncertainty and expected-information-gain formulation we adopt in uncertainty estimation (Sec.~\ref{Supp:A_0}).
}
\rev{
Sec.~\ref{Supp:A_1}~to~\ref{Supp:A_4} describe the implementation details of the proposed physically grounded prefiltering, including the SA-Points generation (Sec.~3.1) and coverage estimation and encoding (Sec.~3.2).
}
\rev{
Sec.~\ref{Supp:A_5}~and~\ref{Supp:A_6} then present the experiment setup and dataset details, respectively.
}
\section{Camera Distribution Analysis}\label{Sec:B}
\label{Supp:Camera_pose}
\begin{figure}[H]
    \centering
    \includegraphics[width=0.9\linewidth]{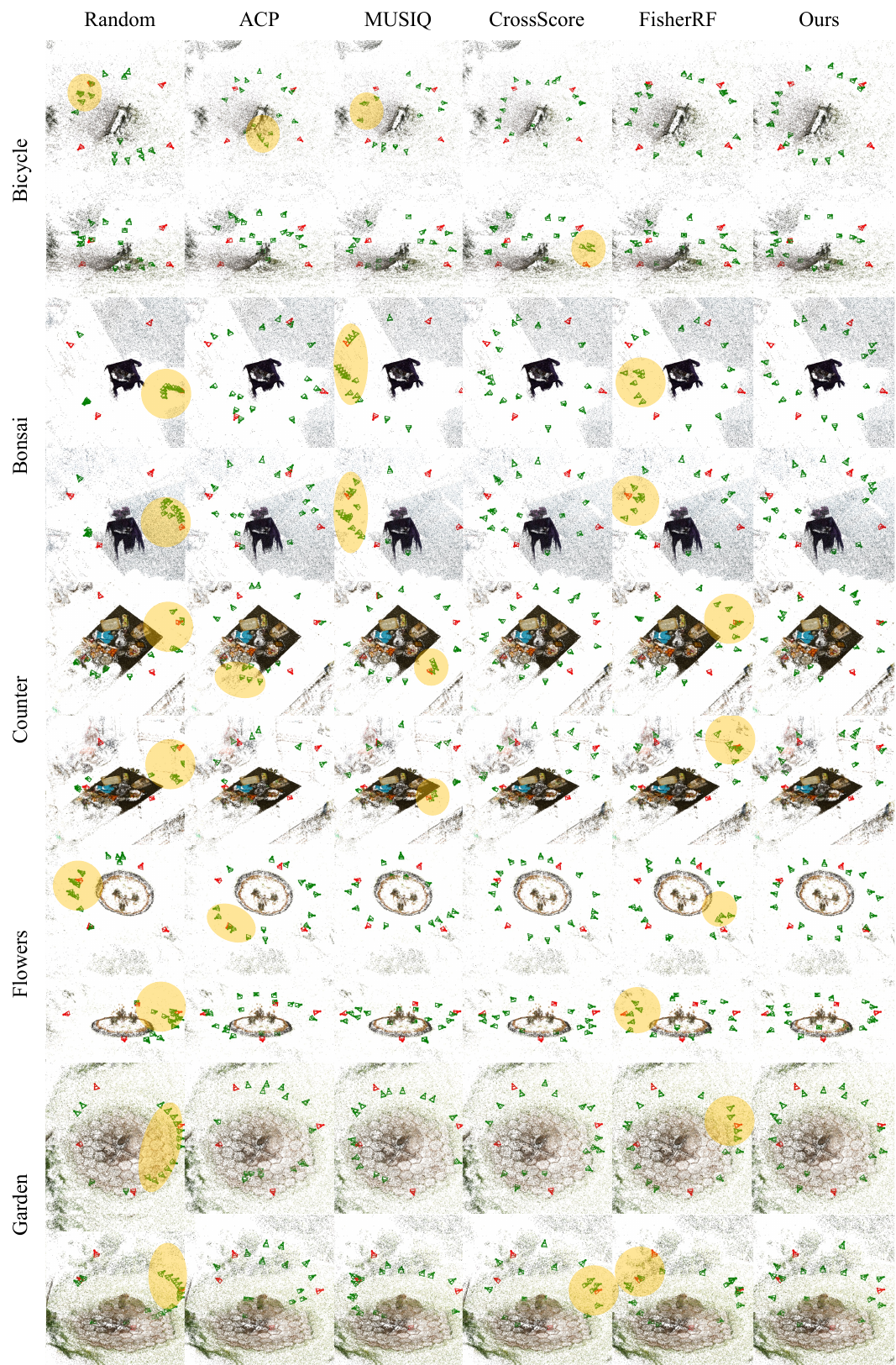}
    \caption{
    \textbf{Camera View Distribution on Mip-NeRF 360.} 
    Visualization of camera poses selected by each method on the Mip-NeRF 360 dataset. 
    Red frustums indicate the initial views, while green frustums denote views added during active selection. 
    Our method produces a more uniformly distributed set of viewpoints, while both baselines exhibit clustered view selections (highlighted by yellow circles), particularly in FisherRF due to its reliance on uncertainty signals entangled with 3DGS training dynamics.
    }
    \label{supp_fig:cam_pose1}
\end{figure}

\begin{figure}[H]
    \centering
    \includegraphics[width=1.0\linewidth]{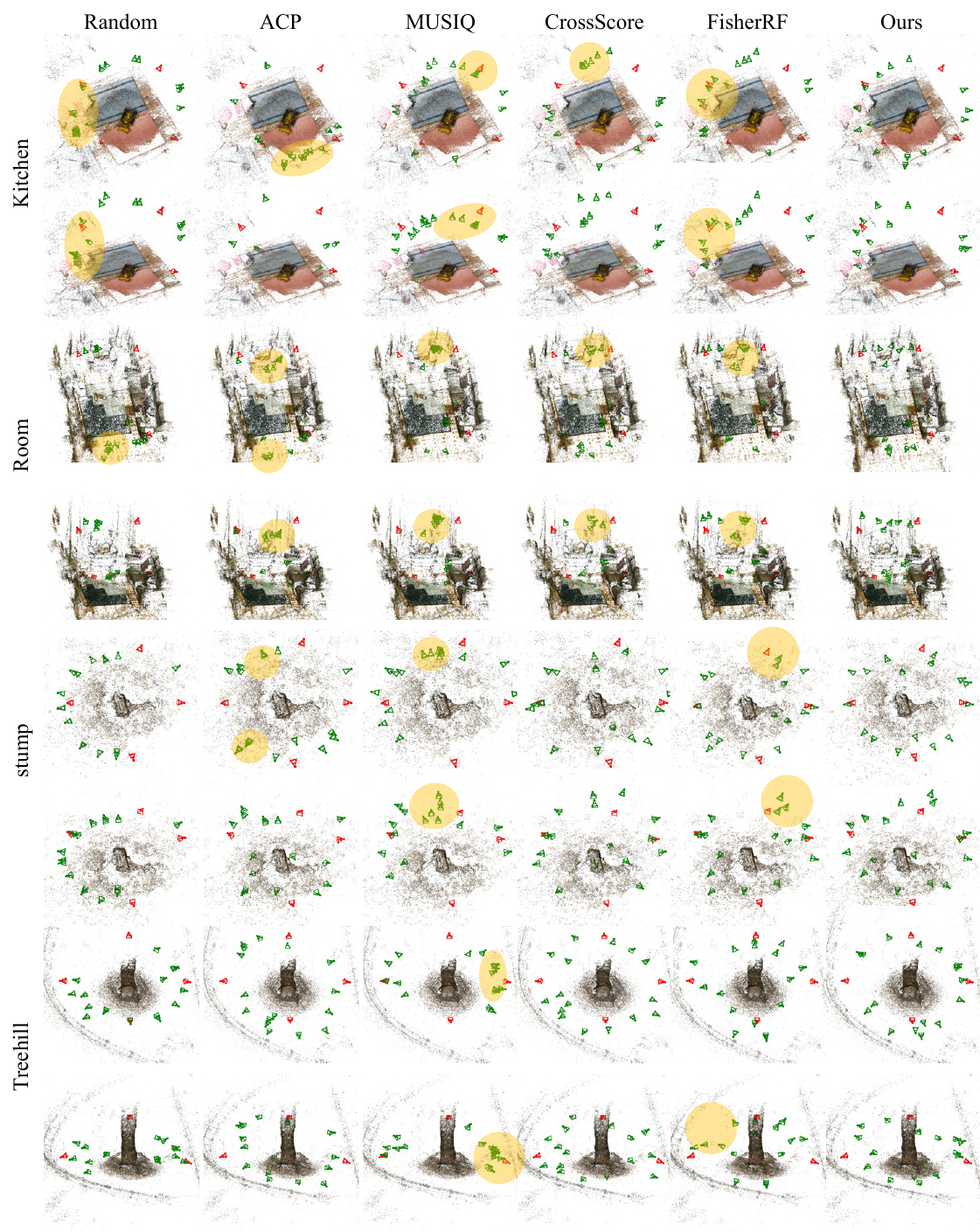}
    \caption{
    \textbf{Camera View Distribution on Mip-NeRF 360.} 
    Visualization of camera poses selected by each method on the Mip-NeRF 360 dataset. 
    Red frustums indicate the initial views, while green frustums denote views added during active selection. 
   Our method produces a more uniformly distributed set of viewpoints, while both baselines exhibit clustered view selections (highlighted by yellow circles), particularly in FisherRF due to its reliance on uncertainty signals entangled with 3DGS training dynamics.
    }
    \label{supp_fig:cam_pose2}
\end{figure}

\begin{figure}[H]
    \centering
    \includegraphics[width=0.9\linewidth]{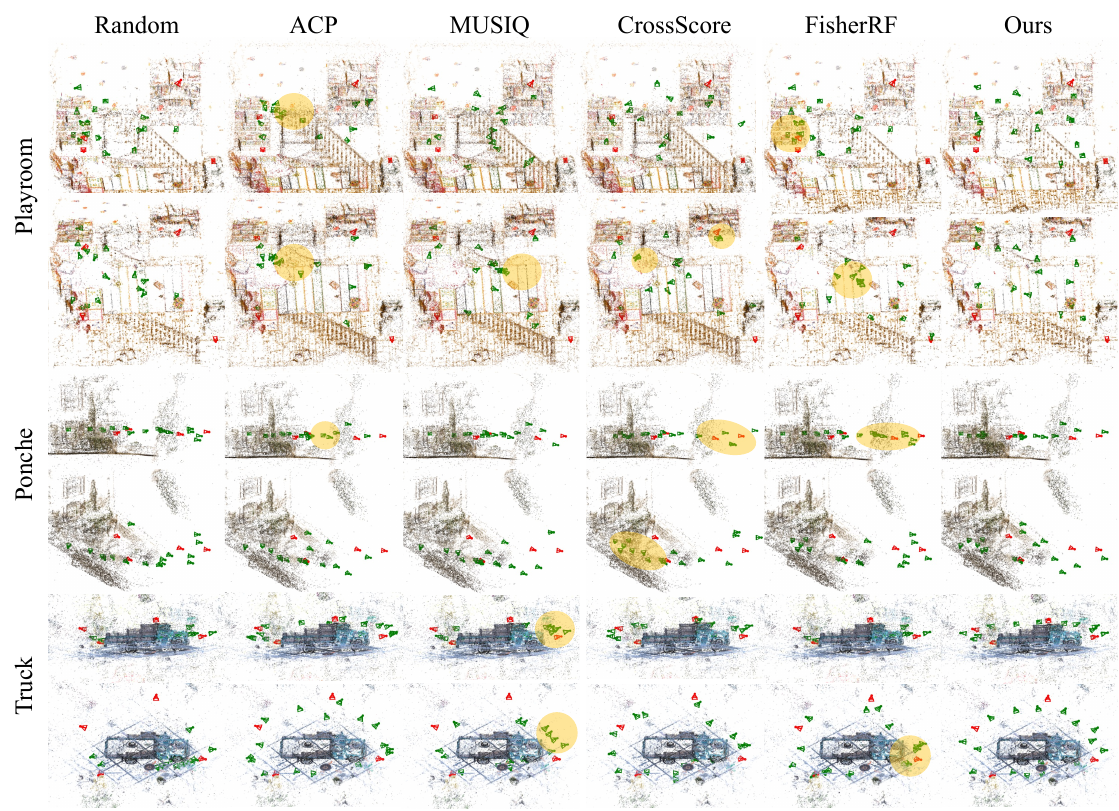}
    \caption{
    \textbf{Camera View Distribution on Deep Blending and Tanks \& Temples.} 
    View distributions on additional datasets, following the same color and annotation scheme as \Fref{supp_fig:cam_pose1}.
    Our method maintains broader spatial coverage, while baseline methods often exhibit clustering (yellow circles), consistent with the biases observed on Mip-NeRF 360.
    }
    \label{supp_fig:cam_pose3}
    \vspace{-4mm}
\end{figure}

To evaluate the effectiveness of our view selection strategy, we compare the camera distributions produced by different methods.
\Fref{supp_fig:cam_pose1}, \ref{supp_fig:cam_pose2}, and
\ref{supp_fig:cam_pose3} visualize these distributions from both bird's-eye and side perspectives. 
As discussed in the manuscript, FisherRF tends to produce clustered view selections due to its tight coupling with the internal 3DGS learning dynamics, as highlighted in the semi-transparent yellow regions. 
In contrast, our method yields a more spatially uniform and well-dispersed distribution of viewpoints.

\begin{figure}[H]
    \centering
    \vspace{-4mm}
    \includegraphics[width=0.85\linewidth]{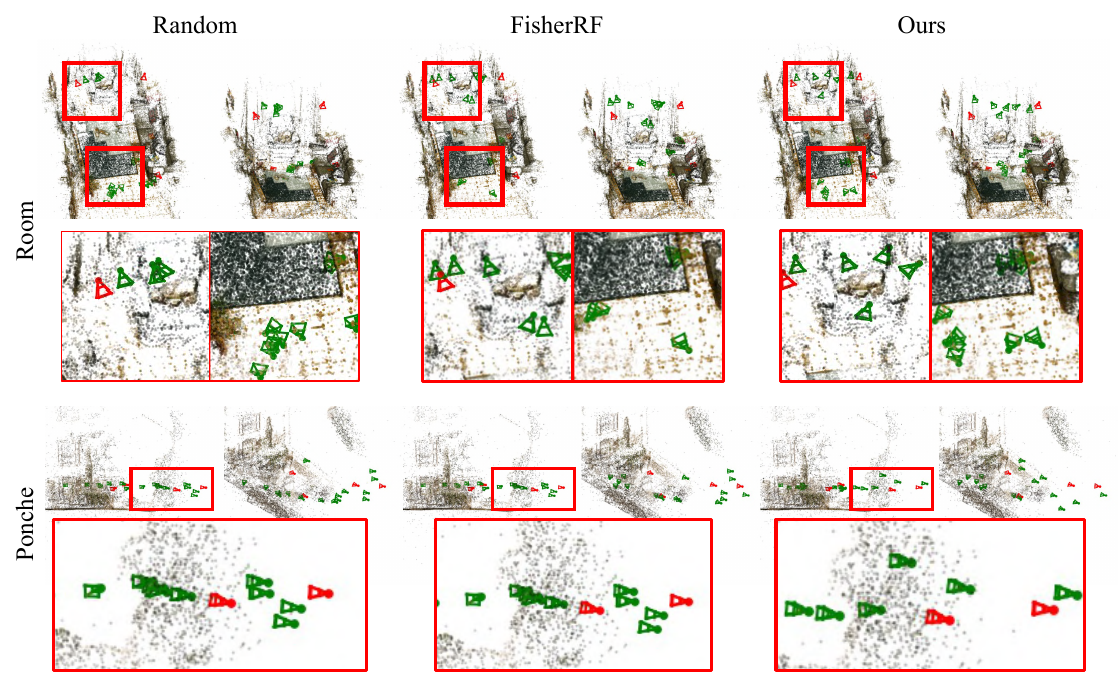}
    \caption{
    \textbf{Diversity of Selected Views (Zoomed-in Analysis).} 
    We highlight the spatial diversity of selected camera poses by presenting zoomed-in regions corresponding to the boxed areas above. 
    Compared to the baselines, our method selects views from a broader range of angles (\textit{Room}, top) and elevations (\textit{Ponche}, bottom), resulting in more comprehensive scene coverage.
    }
    \label{supp_fig:cam_pose4}
    \vspace{-4mm}
\end{figure}

This distinction is further illustrated in \Fref{supp_fig:cam_pose4}: in the \textit{Room} scene from the Mip-NeRF 360 dataset, FisherRF often selects redundant or near-parallel views. 
Our method instead promotes angular diversity, leading to broader scene exploration. 
A similar trend is observed in the Deep Blending dataset, where our approach selects viewpoints over a wider vertical range.
These comparisons support our claim that SA-ResGS facilitates physically grounded and geometrically diverse view selection, thereby improving scene coverage.
\newpage
\section{Extended Qualitative Comparisons}
\label{Supp:Additional_qual}

\begin{figure}[H]
    \centering
    \begin{subfigure}[t]{0.90\linewidth}
        \centering
        \includegraphics[width=1.0\linewidth]{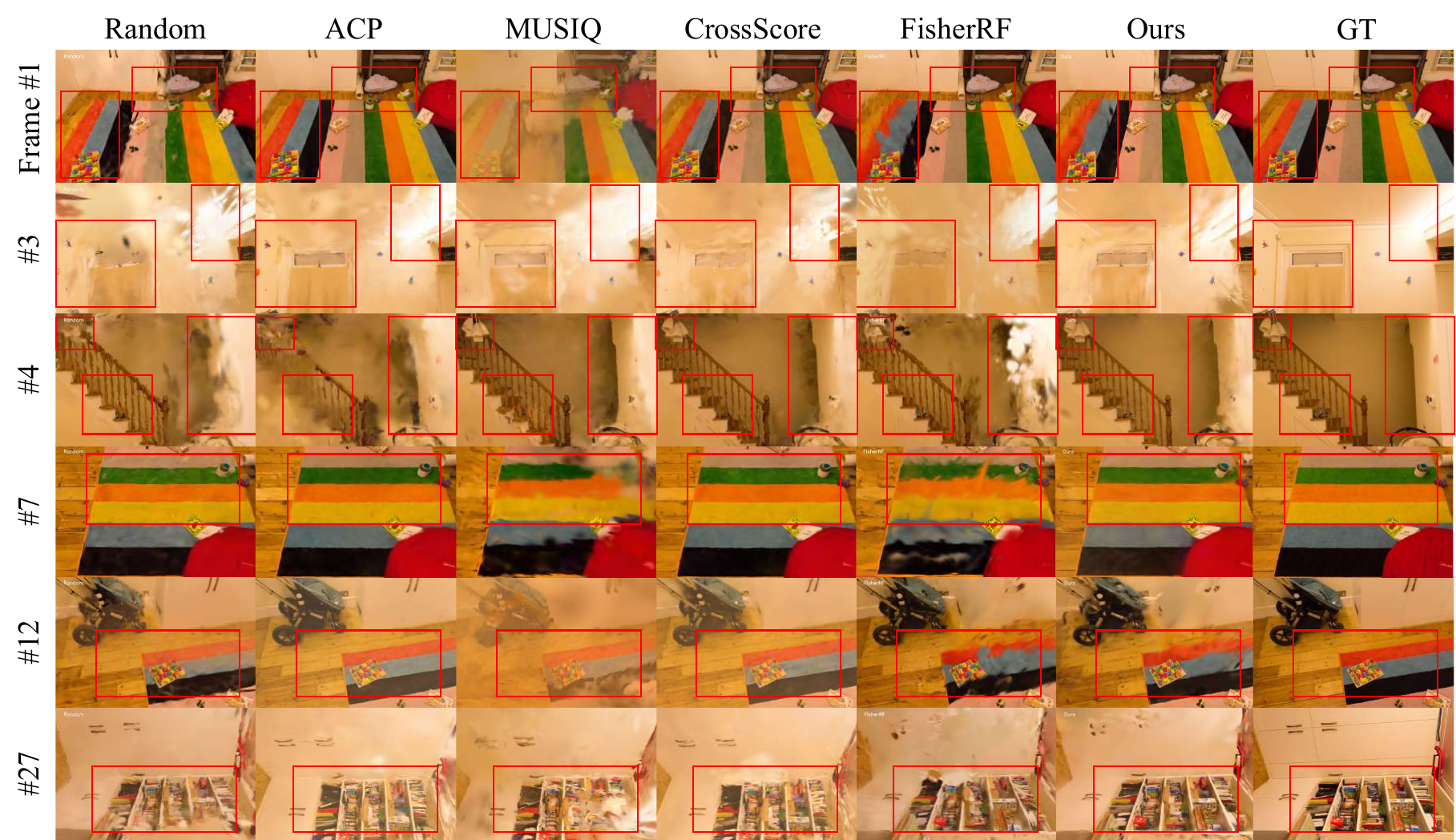}
        \caption{
            \textbf{\textit{Playroom} scene. Six test-time novel views.}
        }
        \label{supp_fig:multiview_5}
    \end{subfigure}

    \vspace{3mm}

    \begin{subfigure}[t]{0.90\linewidth}
        \centering
        \includegraphics[width=1.0\linewidth]{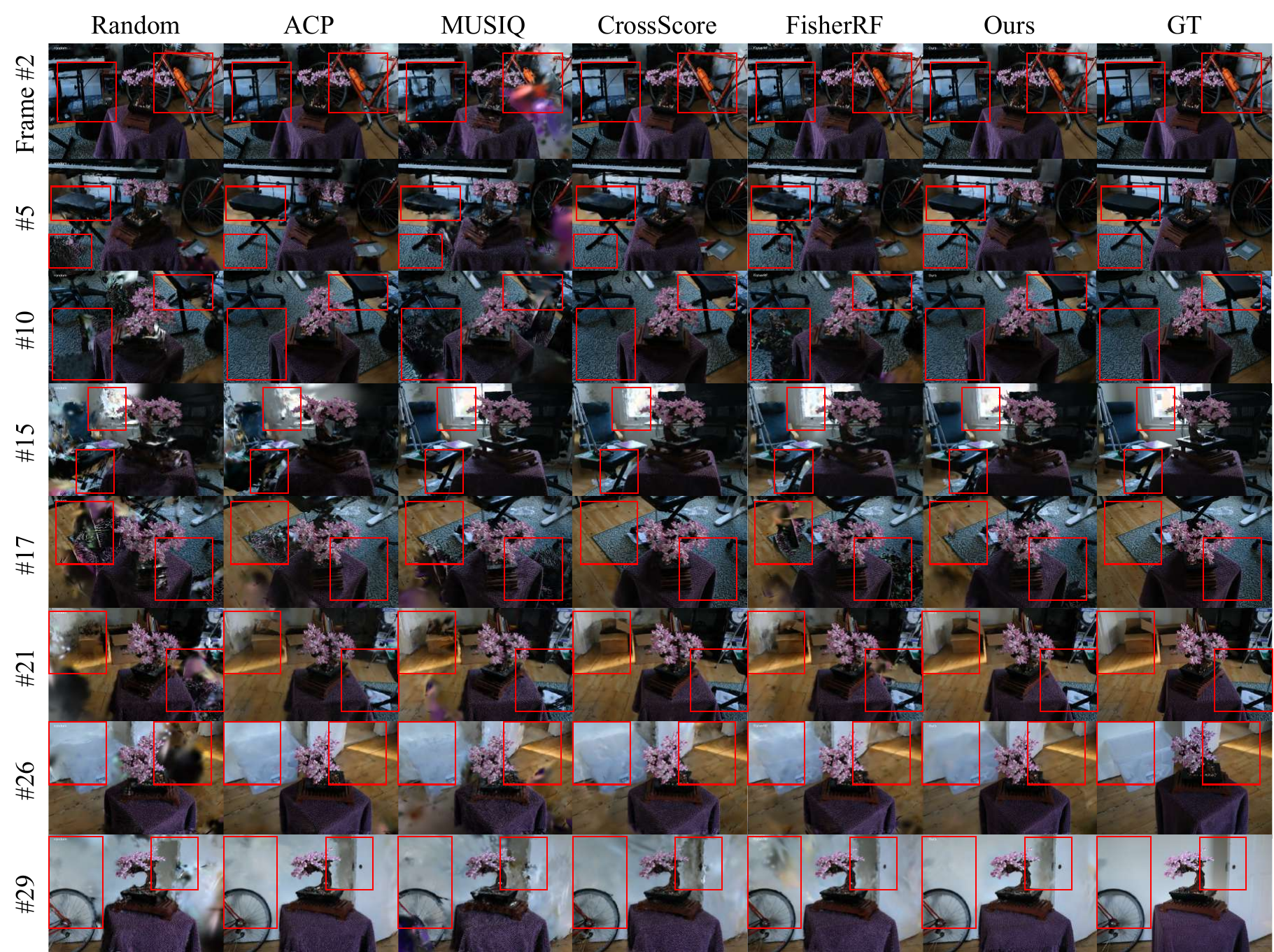}
        \caption{
            \textbf{\textit{Bonsai} scene. Eight test-time novel views.} 
        }
        \label{supp_fig:multiview_1}
    \end{subfigure}

    \caption{\textbf{Qualitative comparison across multiple test-time novel views.}
    Across all scenes, our method produces more complete and consistent reconstructions, particularly in \textbf{occluded or sparsely observed regions} (red boxes). 
    In contrast, baseline methods (Random, FisherRF) often exhibit \textbf{missing geometry, blurring, or structural artifacts} due to biased or clustered view selections.}
    \label{supp_fig:multiview_playroom_bonsai}
\end{figure}

\begin{figure}[H]
    \centering
    \begin{subfigure}[t]{0.95\linewidth}
        \centering
        \includegraphics[width=0.9\linewidth]{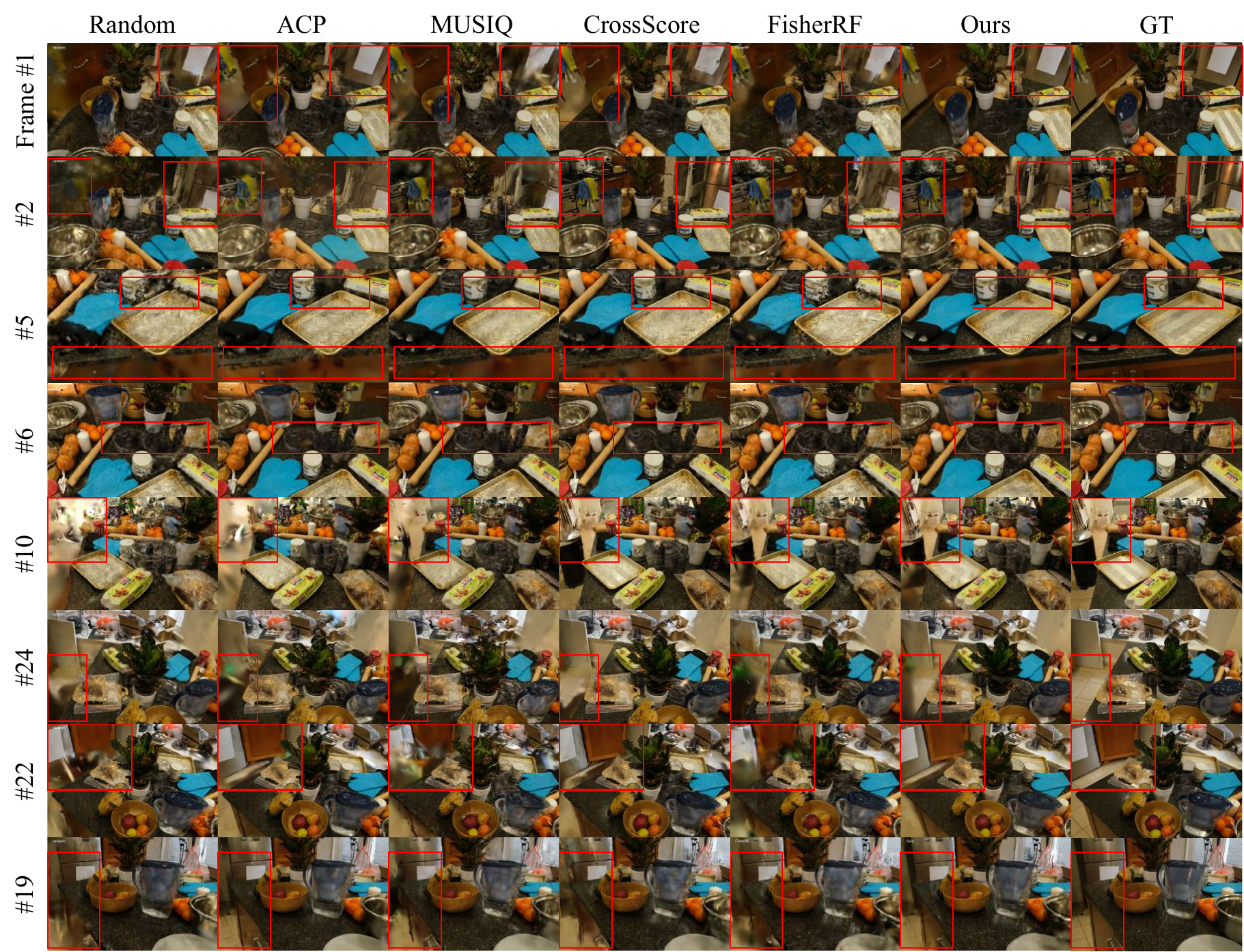}
        \caption{
            \textbf{\textit{Counter} scene. Eight test-time novel views.}
        }
        \label{supp_fig:multiview_2}
    \end{subfigure}

    \begin{subfigure}[t]{0.95\linewidth}
        \centering
        \includegraphics[width=0.9\linewidth]{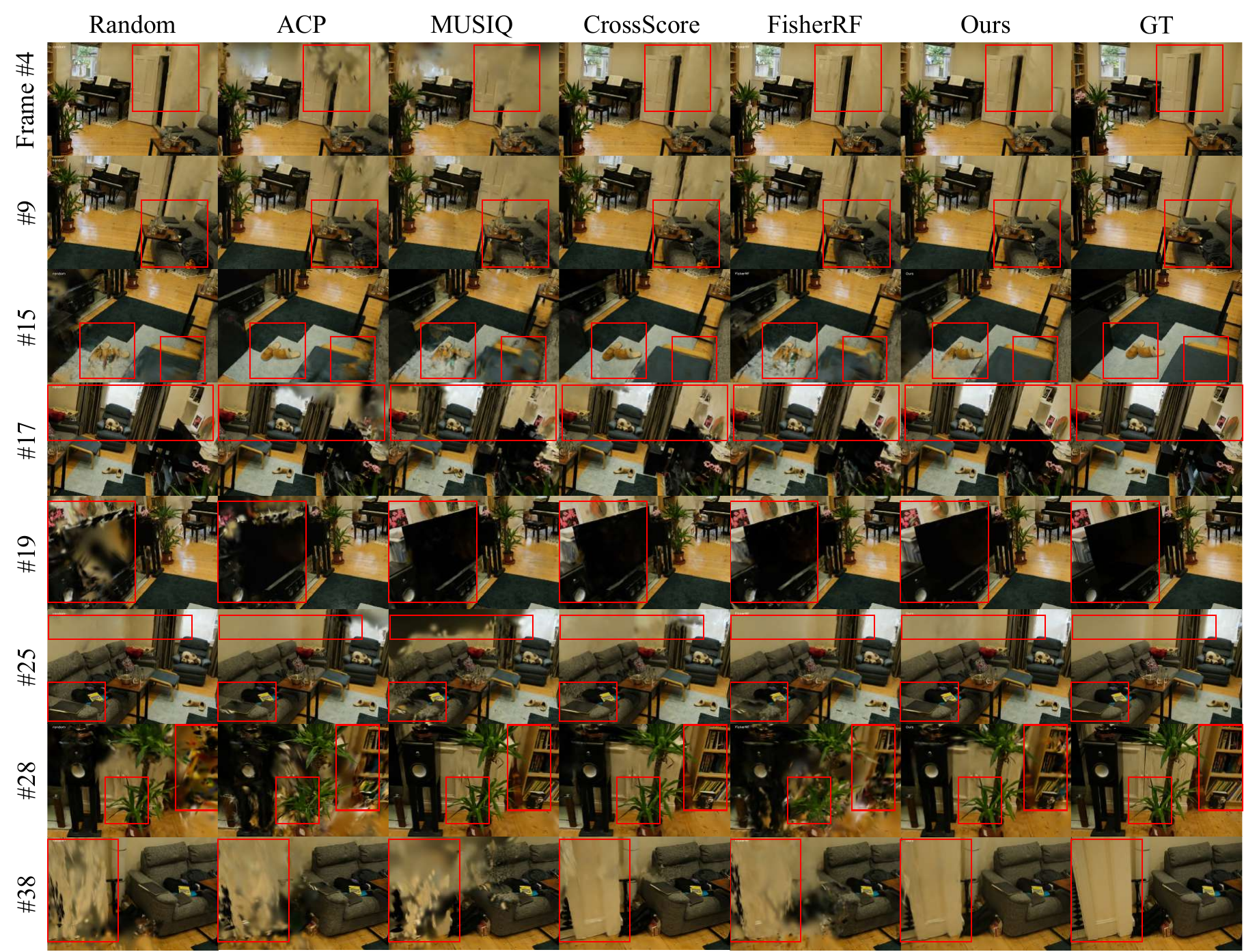}
        \caption{
            \textbf{\textit{Room} scene. Eight test-time novel views.}
        }
        \label{supp_fig:multiview_3}
    \end{subfigure}

    \caption{\textbf{Qualitative comparison across multiple test-time novel views.}
    Across all scenes, our method produces more complete and consistent reconstructions, particularly in occluded or sparsely observed regions (red boxes). In contrast, baseline methods (Random, FisherRF) often exhibit missing geometry, blurring, or structural artifacts due to biased or clustered view selections.
}
    \label{supp_fig:multiview_counter_room}
\end{figure}

\begin{figure}[H]
    \centering
    \includegraphics[width=1.0\linewidth]{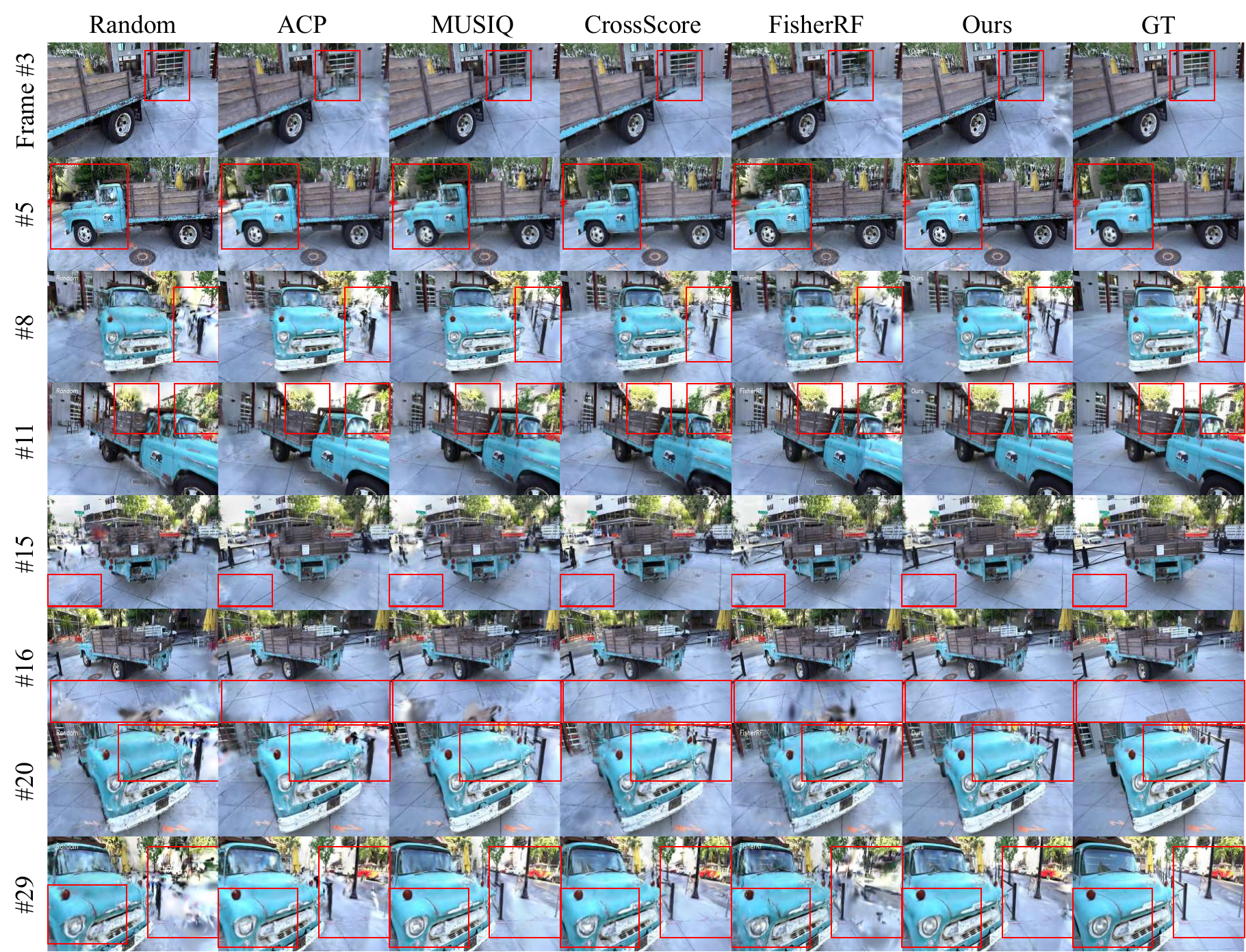}
    \caption{
    \textbf{Qualitative Comparison Across Multiple Test-Time Views (\textit{Truck Scene}).}
    Across all scenes, our method produces more complete and consistent reconstructions, particularly in occluded or sparsely observed regions (red boxes). In contrast, baseline methods (Random, FisherRF) often exhibit missing geometry, blurring, or structural artifacts due to biased or clustered view selections.
    }
    \label{supp_fig:multiview_4}
\end{figure}

The qualitative results presented in the manuscript are limited by space constraints, which may obscure the full advantages of our method. To address this, we provide extended visualizations from multiple test viewpoints. \Fref{supp_fig:multiview_playroom_bonsai} to \ref{supp_fig:multiview_4} display results from five scenes across the Mip-NeRF 360, Extended dataset, with six to eight novel test views per scene. Our method consistently achieves broader and more complete scene coverage than FisherRF. 

As discussed in the manuscript, our residual supervision strategy further improves geometric consistency and reconstruction robustness, particularly in sparse or limited-view scenarios. This is especially beneficial under the standard active or next-best-view selection protocol, where training begins with a small number of views (e.g., 4) and progressively adds new views, typically one at a time. The synergy between physically grounded view selection and residual learning enables high-fidelity reconstruction even from limited initial observations. As mentioned in \Sref{Supp:supple_video_overview}, we include a supplementary video with 360-degree novel-view renderings. We encourage reviewers to watch this video to better appreciate the improvements in coverage and structural accuracy provided by SA-ResGS.
\newpage

\section{Extended Qualitative Comparisons on Ablation Studies.}
\label{Supp:Ablation_Qual}
\begin{figure}[t!]
    \centering
    \includegraphics[width=0.98\linewidth]{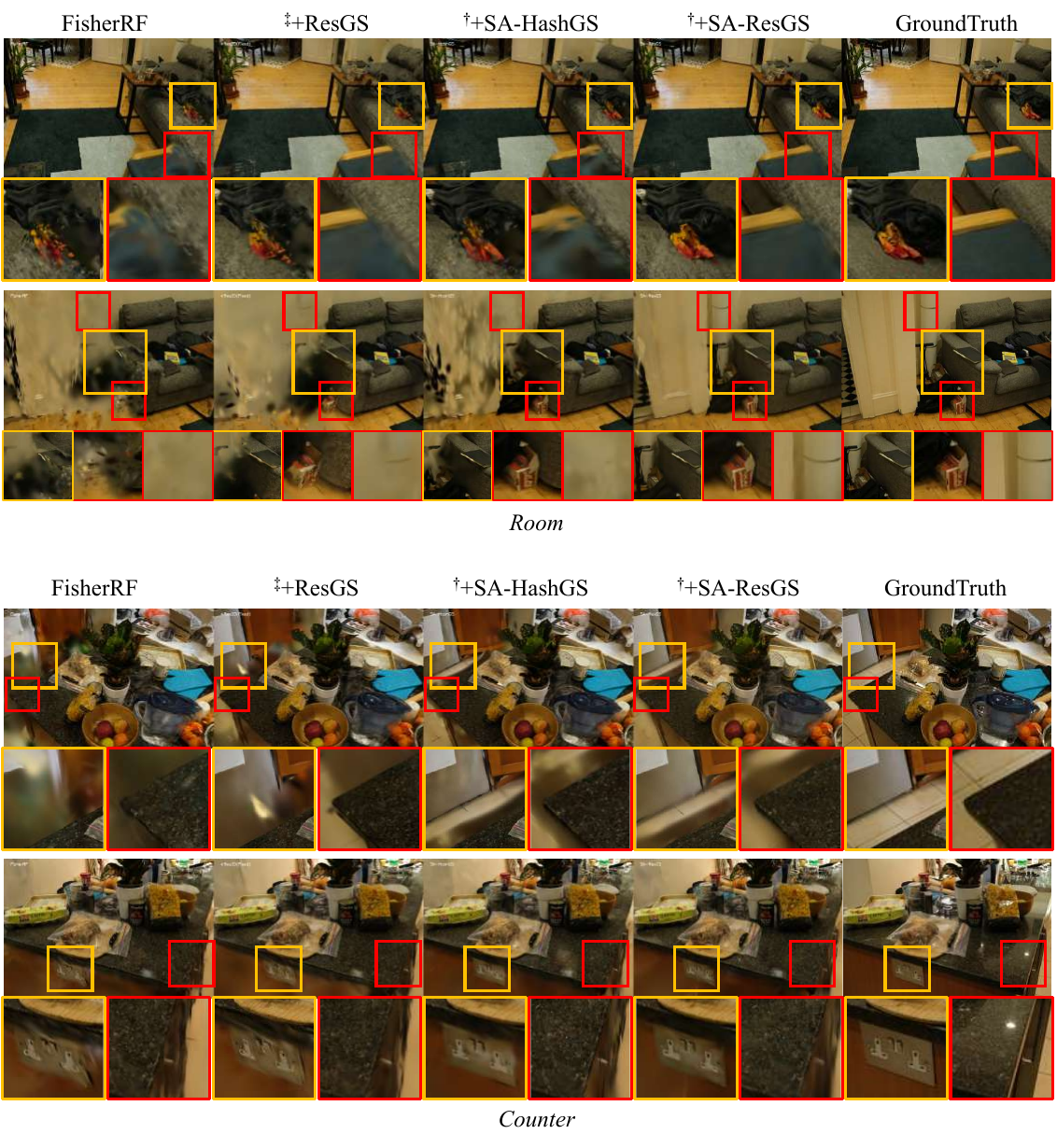}
    \caption{\textbf{Qualitative Self-Comparison for Ablation Study.} Results on the \textit{Room} and \textit{Counter} scenes comparing the baseline (FisherRF) with ablation variants, including ${^\ddagger}$+ResGS, ${^\dagger}$+SA-HashGS, and ${^\dagger}$+SA-ResGS. \textbf{\lightorange{Orange boxes}} highlight improvements from coverage-guided view selection, while \textbf{\red{red boxes}} emphasize the effects of residual supervision. Residual learning enhances geometric stability and reduces artifacts (e.g., jittering surfaces), whereas SA-HashGS recovers unobserved regions (e.g., occluded or shadowed areas). The full SA-ResGS combines both benefits, yielding the most complete and structurally faithful reconstructions.
    }
    \label{supp_fig:ablation}
\end{figure}
\rev{
This section provides additional results that complement the ablation studies in the manuscript. 
We first present qualitative comparisons of the ablation variants to illustrate the effect of each component more clearly.
We then examine the residual learning design under different sampling strategies.
}

\paragraph{Qualitative Results for Ablation Studies}
Figure~\ref{supp_fig:ablation} compares FisherRF (the baseline), $^\ddagger$+ResGS, $^\dagger$+SA-HashGS, and $^\dagger$+SA-ResGS on the \textit{Room} and \textit{Counter} scenes.
Here, $^\ddagger$+ResGS adds residual supervision with fixed-order view selection, while $^\dagger$+SA-HashGS introduces our geometry-aware view prefiltering. 
The full model, $^\dagger$+SA-ResGS, combines both view selection and residual supervision.
Orange boxes indicate improvements mainly associated with view selection, whereas red boxes highlight refinements brought by residual supervision.

In the \textit{Room} scene, the baseline FisherRF exhibits missing geometry (\ie, holes) and floating or blurry artifacts near partially occluded regions or thin object boundaries. 
Physically grounded prefiltering ($^{\dagger}$+SA-HashGS) alleviates missing geometry issues by improving surface coverage through more diverse viewpoint selection, filling holes in the baseline result as shown in the orange boxes.
Residual supervision ($^{\ddagger}$+ResGS), in contrast, provides stronger local refinement in observed regions, improving sofa edges and reducing floating artifacts on the wall and chair, as shown in the red boxes. 
Combining both, the full model ($^{\dagger}$+SA-ResGS) produces the most complete and visually stable reconstruction.

Similarly, in the \textit{Counter} scene, residual supervision ($^{\ddagger}$+ResGS) reduces floating artifacts and improves under-optimized regions, particularly around the shadowed left side of the countertop (first row).
Physically grounded prefiltering ($^{\dagger}$+SA-HashGS) mainly improves broader coverage and recovers missing floor regions that were previously unseen, as highlighted by the orange boxes. 
These results further support the complementary roles of view selection and residual supervision in sparse-view reconstruction.
Additional 360-degree renderings for these ablation results are included in the \textbf{supplementary video}.



\begin{figure}[h]
    \centering
    \includegraphics[width=1.0\linewidth]{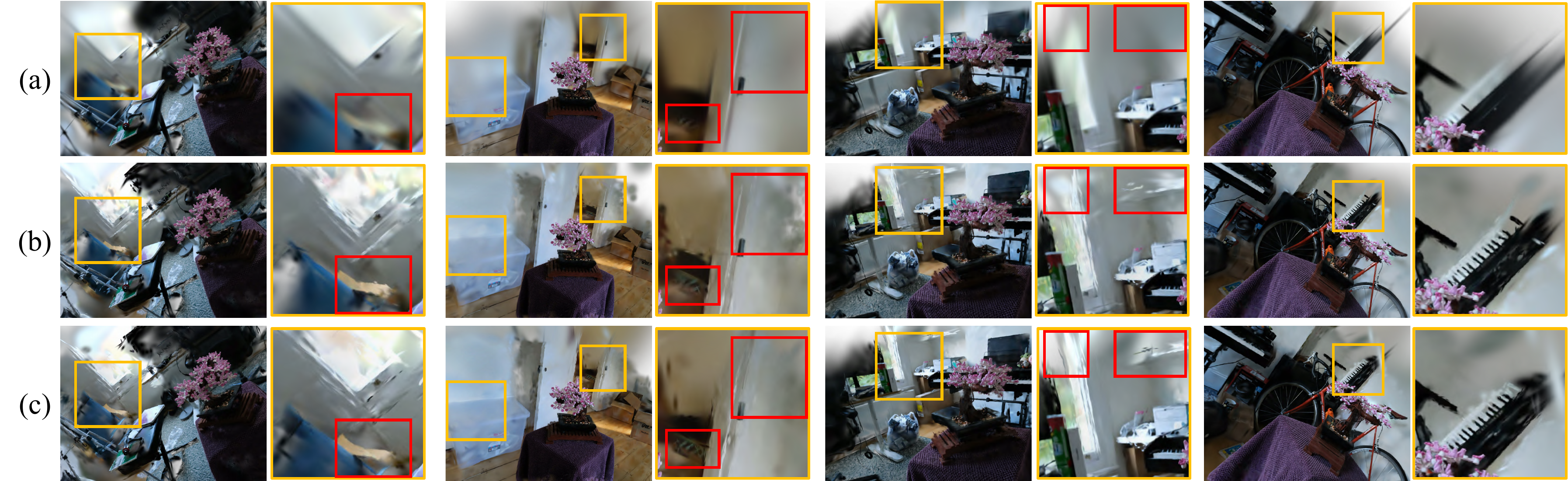}
    \caption{\textbf{Qualitative comparison across ablation variants of the residual-learning design.} Starting from the full configuration (c), w/ Full Loss and w/ Subset Loss using both uncertainty-guided and random sampling, we remove one component at a time: (a) w/o Full Loss while keeping the same Subset Loss, and (b) w/ Full Loss but w/ Subset Loss using random sampling only.
    Red boxes highlight regions where the differences are most pronounced.
    }
    \label{supp_fig:ResGS_ablation2}
\end{figure}

\paragraph{Ablation Study on Residual Learning}
The residual-learning design can be decomposed into two components: Full Loss and uncertainty-guided subset sampling.\footnote{Note that, Full Loss and Subset Loss is defined in Sec 3.3 of manuscript.}

\noindent\textbf{Effect of sampling strategy.}
In Sec.~4.3 and Fig.~8, we examine the ablation effect of Full Loss; here, we extend that analysis to investigate the effect of uncertainty-guided subset sampling, as presented in Fig.~\ref{supp_fig:ResGS_ablation2}. 
For parallel comparison, we also include the result already presented in Fig.~8.

For ease of presentation, we define each ablation variants as (a), (b), and (c), and refer to them by these labels.
We start from our full configuration model (c), w/ Full Loss and w/ Subset Loss using both uncertainty-guided sampling and random sampling.
We then remove one component at a time to form each variant: (a) is w/o Full Loss while keeping the same subset supervision as (c),
and (b) is w/ Full Loss but w/ Subset Loss using random sampling only. 
For fair comparison, the sampling ratio in (b) and (c) is matched by increasing the amount of random sampling in (b).
We also use fixed-order view selection following FisherRF$^\dagger$ so that the observed differences are not confounded by view-selection effects.

Comparing (a) and (c) confirms the role of the Full Loss.
As shown in Fig. 8, removing Full Loss leads to over-smoothed renderings and loss of fine details.
Comparing (b) and (c) further isolates the role of uncertainty-guided subset supervision.
Even with Full Loss present, random subset supervision alone still shows degradation in frequently occluded or under-refined regions, \ie, the staircase behind the door and the window-frame region.
By enforcing continuous gradient flow to under-updated or geometrically unstable Gaussians, particularly those that are floating or insufficiently activated, uncertainty-prior sampling effectively corrects misaligned geometry and improves reconstruction robustness in challenging regions.



\section{Ablation study - Physically Grounded View Selection}
\label{Supp:ablation_prefiltering}

\noindent\textbf{\cam{Pre-filtering criteria.}}
\cam{Table~\ref{table:prefilter} compares our coverage-based pre-filtering against rule-based (random subset), geometry-based (farthest view selection, FVS), and appearance-based (MUSIQ~\cite{musiq}, CrossScore~\cite{crossscore}) alternatives, keeping the uncertainty-based final selection (FisherRF) identical within each pre-filtered subset.
Our method consistently outperforms these alternatives by explicitly reasoning about 3D coverage rather than only promoting spatial or appearance diversity.
The gains are larger on the extended benchmark, where complex geometry and irregular camera distributions make coverage-aware filtering more important for identifying under-covered regions.}


\begin{table}[h]
    \centering
    \caption{\cam{\textbf{Comparison of viewpoint pre-filtering criteria.} Uncertainty-based final selection (FisherRF) is applied within each pre-filtered subset; only the pre-filtering criterion differs. Coverage-aware filtering is best on PSNR/LPIPS and competitive on SSIM, with larger gains on the extended benchmark.}}
    \label{table:prefilter}
    \resizebox{0.85\linewidth}{!}{
    \begin{tabular}{l l c c c c c c}
    \toprule
    \multirow{2}{*}{\textbf{Category}} & \multirow{2}{*}{\textbf{Pre-filtering}} & \multicolumn{3}{c}{\textbf{Mip-NeRF 360}} & \multicolumn{3}{c}{\textbf{Extended dataset}}\\
    \cmidrule(lr){3-5}
    \cmidrule(lr){6-8}
      & & PSNR$\uparrow$ & SSIM$\uparrow$ & LPIPS$\downarrow$ & PSNR$\uparrow$ & SSIM$\uparrow$ & LPIPS$\downarrow$ \\
    \midrule
    Rule-based 
    & Random  
    & 20.975 & 0.597 & 0.459
    & 19.899 & 0.729 & 0.384 \\
    \midrule
    Geo-based
    & FVS
    & \underline{21.337} & \underline{0.611} & \textbf{0.450}  
    & 19.649 & 0.729 & 0.383 \\
    \midrule
    \multirow{2}{*}{Appearance} 
    & MUSIQ
    & 20.652 & 0.594 & 0.463
    & 19.179 & 0.711 & 0.398 \\
    & CrossScore
    & 21.248 & 0.607 & 0.452
    & \underline{20.109} & \textbf{0.743} & \underline{0.372} \\
    \midrule
    Coverage
    & Ours
    & \textbf{21.410} & \textbf{0.613} & \underline{0.451}
    & \textbf{20.401} & \underline{0.732} & \textbf{0.361} \\
    \bottomrule
    \end{tabular}}
\end{table}

\noindent\textbf{\cam{Coverage-only view selection.}}
\cam{To isolate the effect of coverage alone, Table~\ref{table:geomonly} evaluates an SA-Points-only variant that selects the view with the highest SA-Point-based coverage/dissimilarity score, without using any uncertainty signal.
This variant improves PSNR/SSIM over FisherRF, confirming that SA-Points provide useful coverage cues.
However, it still underperforms our full method, showing that geometry-based coverage is helpful but insufficient; uncertainty still provides complementary gains.}


\begin{table}[h]
    \centering 
    \vspace{-3mm}
    \caption{\cam{\textbf{Coverage-only (SA-Points-only) view selection vs. the full method.} The SA-Points-only variant selects the view with the highest SA-Point-based coverage/dissimilarity score, without uncertainty. SA-Points provide useful coverage cues, but uncertainty yields complementary gains.}}
    \label{table:geomonly}
    \resizebox{0.7\linewidth}{!}{
    \begin{tabular}{l c c c c c c}
    \toprule
    \multirow{2}{*}{\textbf{Methods}} & \multicolumn{3}{c}{\textbf{Mip-NeRF 360}} & \multicolumn{3}{c}{\textbf{Extended dataset}}\\
    \cmidrule(lr){2-4}
    \cmidrule(lr){5-7}
      & PSNR$\uparrow$ & SSIM$\uparrow$ & LPIPS$\downarrow$ & PSNR$\uparrow$ & SSIM$\uparrow$ & LPIPS$\downarrow$ \\
    \midrule
    FisherRF 
    & 20.642 & 0.595 & \textbf{0.450}
    & 19.654 & 0.711 & 0.370 \\
    SA-Points-only
    & \underline{20.802} & \underline{0.601} & 0.455
    & \underline{19.727} & \underline{0.723} & \underline{0.369} \\
    Ours-Full &
    \textbf{21.410} & \textbf{0.613} & \underline{0.451} & 
    \textbf{20.401} & \textbf{0.732} & \textbf{0.361} \\
    \bottomrule
    \end{tabular}}
    \vspace{-8mm}
\end{table}
\section{Ablation study - Uncertainty-guided Residual Learning}
\label{Supp:ablation_residual}

\paragraph{\cam{Residual-Learning Design Choices}}
\cam{We analyze two key design choices of the residual supervision (Sec.~3.3), summarized in~\Tref{table:beta_attr}.
First, the sensitivity study on $\beta$, the ratio of selected uncertain Gaussians (left), shows that residual supervision works best at a moderate selection strength but degrades when too many Gaussians are selected, which weakens the dropout effect.
We use $\beta{=}10$ as a robust default: differences across moderate $\beta$ are small, and this value balances dropout strength and stability.
Second, the ablation on the attributes used to rank uncertain Gaussians (right) shows that using both opacity and scale is most effective, validating our design.
For a fair comparison, we double $\beta$ for the opacity-only and scale-only variants.}

\begin{table}[h]
    \vspace{-3mm}
    \centering
    \caption{\cam{\textbf{Residual-learning design choices on Mip-NeRF 360.} Left: sensitivity to the uncertain-Gaussian ratio $\beta$. Right: effect of the attributes used to rank uncertain Gaussians ($\beta$ is doubled for the single-attribute variants for a fair comparison). Residual supervision works best at a moderate $\beta$ and degrades when too many Gaussians are dropped; using both opacity and scale is most effective.}}
    \label{table:beta_attr}
    \begin{minipage}{0.3\linewidth}\centering
    \resizebox{\linewidth}{!}{
    \begin{tabular}{c c c c}
    \toprule
    $\beta$ (\%) & PSNR$\uparrow$ & SSIM$\uparrow$ & LPIPS$\downarrow$ \\
    \midrule
    0  & 21.191 & 0.609 & 0.451 \\
    5  & \textbf{21.387} & \textbf{0.613} & \textbf{0.448} \\
    10 & 21.347 & 0.611 & 0.450 \\
    20 & 21.329 & 0.610 & 0.451 \\
    30 & 21.215 & 0.610 & 0.450 \\
    \bottomrule
    \end{tabular}}
    \end{minipage}\hspace{1em}%
    \begin{minipage}{0.5\linewidth}\centering
    \resizebox{\linewidth}{!}{
    \begin{tabular}{l c c c}
    \toprule
    Attributes & PSNR$\uparrow$ & SSIM$\uparrow$ & LPIPS$\downarrow$ \\
    \midrule
    opacity-only    & 21.345 & 0.612 & 0.450 \\
    scale-only      & 21.309 & 0.609 & 0.451 \\
    opacity + scale & \textbf{21.410} & \textbf{0.613} & 0.451 \\
    \bottomrule
    \end{tabular}}
    \end{minipage}
    \vspace{-8mm}
\end{table}

\section{\rev{Robustness and Sensitivity Analysis}}
\label{Supp:Ablation_robust}
\rev{
This section presents additional ablation studies to evaluate the robustness and sensitivity of the proposed framework with respect to several design choices.
We examine the effects of correspondence noise, reprojection thresholds, and the hash encoding size in the coverage prefilter.
Lastly, we evaluate changes in reconstruction performance across varying numbers of selected views, to further validate the effectiveness of our framework during the active view selection process.
}


\begin{wraptable}{h}{0.45\linewidth}
    \vspace{-12mm}
    \renewcommand{\arraystretch}{0.8}
    \centering
    \caption{\textbf{Robustness to synthetic correspondence noise.} PSNR remains stable under moderate perturbations ($\le$ 5 pixels), showing SA-Points are resilient to realistic levels of noise.}
    \label{table:correspond_noise}
    \resizebox{\linewidth}{!}{         
        \begin{tabular}{cccccc}
        \toprule
        Noise      & 0.0                  & 0.5                  & 1.0                  & 5.0                  & 10.0                 \\
        \midrule
        PSNR       & 24.441               & 24.199               & 24.117               & 24.311               & 23.121               \\
    \bottomrule
    \vspace{-7mm}
    \end{tabular}
    }
\end{wraptable} 
\paragraph{Robustness to Correspondence Noise}
SA-Points are obtained from 2D dense correspondences used for triangulation, where matching errors directly affect 3D reconstruction quality and subsequent view selection.
To evaluate robustness, we synthetically inject gaussian noise into 2D correspondence, ranging from 0.0 to 10.0 pixels.
As shown in~\Tref{table:correspond_noise}, reconstruction quality remains stable in terms of PSNR under moderate noise levels up to 5 pixels, with a 1-pixel reprojection filter.
Performance degrades noticeably under larger perturbations, indicating that the system is affected by 2D correspondence accuracy but still tolerates moderate correspondence errors without significant loss.


\begin{wraptable}{h}{0.4\linewidth}
    \vspace{-12mm}
    \renewcommand{\arraystretch}{0.8}
    \centering
    \caption{\textbf{Effect of reprojection filtering thresholds.} A 1-pixel threshold offers the best trade-off between PSNR and coverage.}
    \label{table:reproj_error}
    \resizebox{\linewidth}{!}{
    \begin{tabular}{ccccc}
    \toprule
    Threshold  & PSNR    & SSIM   & LPIPS  & Coverage \\
    \midrule
    0.5              & 22.634  & 0.817  & 0.332  & 57.74\%  \\
    1.0              & 24.441  & 0.838  & 0.317  & 94.11\%  \\
    2.0              & 24.244  & 0.834  & 0.318  & 94.89\%  \\
    \bottomrule
    \end{tabular}}
    \vspace{-7mm}
\end{wraptable}
We also study the sensitivity to the reprojection threshold used for correspondence filtering.
To verify this, we vary the reprojection threshold and measure reconstruction accuracy and the coverage metric for the Next Best View selection task.
Coverage is measured by counting the number of observed voxels when 10 views are selected.
As shown in~\Tref{table:reproj_error}, with a tight threshold of 0.5 pixels, many correspondences are filtered out, resulting in reduced coverage and degraded reconstruction quality.
At a looser threshold of 2.0 pixels, more SA-Points are preserved for coverage estimation; however, these points are less accurate, which slightly reduces performance.
Among the tested values, 1.0 pixel provides the best PSNR–coverage trade-off. 
Based on this observation, we use a 1-pixel reprojection threshold in all experiments.


\begin{wraptable}{r}{0.35\linewidth}
    \vspace{-12mm}
    \centering
    \caption{\textbf{Effect of hash encoding size on coverage prefiltering.} Performance comparison across different hash-table sizes for coverage prefiltering.}
    \label{table:hash_ablation}
    \resizebox{\linewidth}{!}{
    \begin{tabular}{cccc}
    \toprule
    Hash size & PSNR    & SSIM	 & LPIPS \\
    \midrule
    $2^{11}$              & 21.555  & 0.617    & 0.446 \\
    $2^{13}$              & 21.593  & 0.618    & 0.446 \\
    $2^{15}$              & 21.507  & 0.616    & 0.447 \\
    $2^{17}$              & 21.261  & 0.612    & 0.450 \\
    $2^{19}$              & 21.282  & 0.610    & 0.451 \\
    \midrule
    No Collision	& 21.325  & 0.610    & 0.450 \\
    \bottomrule
    \vspace{-7mm}
    \end{tabular}}
\end{wraptable}
\paragraph{Ablation on Hash Encoding Size}
To assess the sensitivity of the hash-encoded voxel grid used in coverage prefiltering, we vary the hash-table size from $2^{11}$ to $2^{19}$, as well as a no-collision variant implemented with direct indexing.
Despite more than a 200$\times$ difference in hash capacity, the performance remains stable across all settings, with only minor variations in PSNR, SSIM, and LPIPS, as shown in~\Tref{table:hash_ablation}.
No consistent degradation trend is observed as the hash size decreases.

We attribute this insensitivity to two aspects of the pipeline:
(1) Only occupied voxels are hashed, while empty regions are skipped entirely; because occupancy varies widely across scenes, the effective load factor of the hash table remains low even for relatively small hash sizes.
(2) Residual inconsistencies introduced by collisions are further mitigated by the subsequent Fisher-based fine selection stage, which provides an additional layer of error correction.
Together,  
these results suggest that the coverage prefiltering is robust to hash size over a wide range in our benchmark setting.


\begin{wraptable}{r}{0.5\linewidth}
    \centering
    \renewcommand{\arraystretch}{0.8}
    \vspace{-12mm}
    \caption{\textbf{Effect of the number of selected views.} Our method yields higher PSNR in sparse-view regimes (7–10 views) and continues to improve as more views are added, highlighting both early stability and long-term scalability.}
    \label{table:number_of_view}
    \resizebox{\linewidth}{!}{
    \begin{tabular}{ccccccc}
    \toprule
    \multirow{2}{*}{Methods} & \multicolumn{6}{c}{Selected Views} \\
    \cmidrule(lr){2-7}
                   & 4      & 7      & 10     & 13     & 16     & 19     \\
    \midrule
    random         & 15.929 & 16.485 & 17.573 & 17.694 & 18.233 & 18.400 \\
    FisherRF       & 15.985 & 17.269 & 18.773 & 20.188 & 21.722 & 22.650 \\
    Ours           & 15.970 & 17.606 & 18.895 & 20.159 & 23.229 & 24.064 \\
    \bottomrule
    \vspace{-7mm}
    \end{tabular}}
\end{wraptable}
\paragraph{Effects of the Number of Selected Views}
Our physically grounded view selection algorithm (Sec. 3.2 in the manuscript) is designed to stabilize early training when the geometry is sparse or unevenly distributed, while also benefiting from later uncertainty-driven refinement. 
To further examine this effect, we report intermediate results obtained with fewer selected views.
Specifically, on the \textit{Bonsai} scene, we evaluate performance with 4, 7, 10, 13, 16, and 19 selected views in~\Tref{table:number_of_view}.
Under sparse conditions (7 and 10 views), our method outperforms FisherRF, demonstrating the benefit of physically grounded filtering in stabilizing early-stage. 
As more views are added, the initial advantage narrows, but new gains appear from 16 views onward, where residual learning further refines Gaussians and improves geometry, enabling later performance gains.
This trend is consistent with the design intuition of our model.
\section{Extended Baseline Comparison}
\label{Supp:extended_baseline}

\begin{table}[h]
    \centering
    \caption{\cam{\textbf{Additional baseline comparison.} BayesRays uses its original Nerfacto backbone; ActiveNeRF is adapted to 3DGS in our pipeline. Our method outperforms both on Mip-NeRF 360 and the extended benchmark, consistent with the trend reported in the FisherRF paper~\cite{fisherrf}.}}
    \label{table:morebaselines}
    \resizebox{0.8\linewidth}{!}{
    \begin{tabular}{l c c c c c c}
    \toprule
    \multirow{2}{*}{\textbf{Method}} & \multicolumn{3}{c}{\textbf{Mip-NeRF 360}} & \multicolumn{3}{c}{\textbf{Extended Benchmark}}\\
    \cmidrule(lr){2-4} \cmidrule(lr){5-7}
     & PSNR$\uparrow$ & SSIM$\uparrow$ & LPIPS$\downarrow$ & PSNR$\uparrow$ & SSIM$\uparrow$ & LPIPS$\downarrow$ \\
    \midrule
    Nerfacto + BayesRays~\cite{bayesrays}    & 15.430 & 0.466 & 0.673 & 17.158 & 0.565 & 0.448 \\
    3DGS + ActiveNeRF~\cite{activenerf}      & 18.749 & 0.542 & 0.474 & 19.266 & 0.702 & 0.375 \\
    3DGS + Ours (MASt3R)    & \textbf{21.410} & \textbf{0.613} & \textbf{0.451} & \textbf{20.401} & \textbf{0.732} & \textbf{0.361} \\
    \bottomrule
    \end{tabular}}
\end{table}

\noindent\textbf{\cam{Additional baselines.}}
\cam{Following the broader baseline set used in FisherRF~\cite{fisherrf}, we additionally compare against BayesRays~\cite{bayesrays} and an ActiveNeRF-style acquisition~\cite{activenerf} adapted to 3DGS, as reported in Table~\ref{table:morebaselines}.
BayesRays is reported with its original Nerfacto backbone, while ActiveNeRF is adapted to 3DGS in our pipeline.
Our method consistently outperforms both baselines on Mip-NeRF 360 and the extended benchmark, consistent with the trend reported in the FisherRF paper.}

\begin{table*}[h]
    \centering
    \caption{\textbf{Scene-wise quantitative results on the NeRF Synthetic dataset.} Each subtable reports PSNR, SSIM, and LPIPS, respectively. All values are computed over four repeated trials with different random seeds.}
    \label{table:scenewise_metrics_synth}
    \begin{subtable}[h]{\linewidth}
        \centering
        \resizebox{\linewidth}{!}{
        \begin{tabular}{cccccccc}
        \toprule
        \multirow{2}{*}{\textbf{Methods}} & \multicolumn{7}{c}{PSNR$\uparrow$} \\
        \cmidrule(lr){2-8}
        & chair & drums & hotdog & lego & materials & mic & ship \\	
        \midrule
        random & 24.626  \scriptsize{$\pm$3.489} & 20.918  \scriptsize{$\pm$1.347} & 30.397  \scriptsize{$\pm$1.161} & 28.812  \scriptsize{$\pm$0.285} & 20.013  \scriptsize{$\pm$0.646} & 23.813  \scriptsize{$\pm$1.236} & 25.348  \scriptsize{$\pm$0.830} \\
        ACP & 25.989  \scriptsize{$\pm$0.219} & 19.176  \scriptsize{$\pm$0.604} & 24.409  \scriptsize{$\pm$0.755} & 23.474  \scriptsize{$\pm$0.104} & 19.125  \scriptsize{$\pm$0.295} & 22.687  \scriptsize{$\pm$0.476} & 24.168  \scriptsize{$\pm$0.726} \\
        MUSIQ & 27.899  \scriptsize{$\pm$0.164} & 20.948  \scriptsize{$\pm$1.108} & 30.308  \scriptsize{$\pm$0.140} & 28.796  \scriptsize{$\pm$0.159} & 20.268  \scriptsize{$\pm$0.433} & 23.219  \scriptsize{$\pm$0.437} & 25.219  \scriptsize{$\pm$0.335} \\
        CrossScore & 24.521  \scriptsize{$\pm$0.531} & 18.817  \scriptsize{$\pm$0.094} & 28.799  \scriptsize{$\pm$0.661} & 27.629  \scriptsize{$\pm$0.985} & 18.326  \scriptsize{$\pm$0.187} & 23.011  \scriptsize{$\pm$0.378} & 25.120  \scriptsize{$\pm$0.325} \\
        FisherRF & 27.066  \scriptsize{$\pm$0.981} & 21.844  \scriptsize{$\pm$0.608} & 30.998  \scriptsize{$\pm$0.371} & 26.108  \scriptsize{$\pm$2.142} & 20.516  \scriptsize{$\pm$0.488} & 24.153  \scriptsize{$\pm$0.668} & 25.645  \scriptsize{$\pm$0.408} \\
        \midrule
        Ours (MASt3R) & 28.303  \scriptsize{$\pm$0.280} & 22.948  \scriptsize{$\pm$0.304} & 31.195  \scriptsize{$\pm$0.166} & 29.703  \scriptsize{$\pm$0.450} & 21.206  \scriptsize{$\pm$0.187} & 26.267  \scriptsize{$\pm$0.271} & 26.437  \scriptsize{$\pm$0.042} \\
        Ours (DA-v3) & 28.375  \scriptsize{$\pm$0.142} & 22.906  \scriptsize{$\pm$0.185} & 31.266  \scriptsize{$\pm$0.592} & 29.941  \scriptsize{$\pm$0.222} & 20.818  \scriptsize{$\pm$0.452} & 25.455  \scriptsize{$\pm$0.238} & 26.294  \scriptsize{$\pm$0.341} \\
        \bottomrule
        \end{tabular}
        }
        \caption{Average PSNR on NeRF Synthetic dataset}
        \label{tab:psnr_synth}
    \end{subtable}

    \vspace{3mm}

    \begin{subtable}[t]{\linewidth}
        \centering
        \resizebox{\linewidth}{!}{
        \begin{tabular}{cccccccccc}
        \toprule
        \multirow{2}{*}{\textbf{Methods}} & \multicolumn{7}{c}{SSIM$\uparrow$} \\
        \cmidrule(lr){2-8}
        & chair & drums & hotdog & lego & materials & mic & ship \\	
        \midrule
        random & 0.932  \scriptsize{$\pm$0.019} & 0.878  \scriptsize{$\pm$0.018} & 0.962  \scriptsize{$\pm$0.005} & 0.937  \scriptsize{$\pm$0.001} & 0.814  \scriptsize{$\pm$0.014} & 0.896  \scriptsize{$\pm$0.015} & 0.829  \scriptsize{$\pm$0.014} \\
        ACP & 0.920  \scriptsize{$\pm$0.001} & 0.849  \scriptsize{$\pm$0.012} & 0.920  \scriptsize{$\pm$0.007} & 0.850  \scriptsize{$\pm$0.002} & 0.804  \scriptsize{$\pm$0.010} & 0.860  \scriptsize{$\pm$0.005} & 0.785  \scriptsize{$\pm$0.011} \\        
        MUSIQ & 0.940  \scriptsize{$\pm$0.003} & 0.880  \scriptsize{$\pm$0.010} & 0.961  \scriptsize{$\pm$0.001} & 0.938  \scriptsize{$\pm$0.002} & 0.817  \scriptsize{$\pm$0.006} & 0.867  \scriptsize{$\pm$0.003} & 0.820  \scriptsize{$\pm$0.013} \\
        CrossScore & 0.930  \scriptsize{$\pm$0.003} & 0.826  \scriptsize{$\pm$0.008} & 0.951  \scriptsize{$\pm$0.003} & 0.925  \scriptsize{$\pm$0.004} & 0.776  \scriptsize{$\pm$0.007} & 0.848  \scriptsize{$\pm$0.011} & 0.823  \scriptsize{$\pm$0.007} \\
        FisherRF & 0.945  \scriptsize{$\pm$0.003} & 0.885  \scriptsize{$\pm$0.012} & 0.965  \scriptsize{$\pm$0.002} & 0.908  \scriptsize{$\pm$0.021} & 0.810  \scriptsize{$\pm$0.005} & 0.898  \scriptsize{$\pm$0.007} & 0.834  \scriptsize{$\pm$0.007} \\
        \midrule
        Ours (MASt3R) & 0.949  \scriptsize{$\pm$0.002} & 0.903  \scriptsize{$\pm$0.004} & 0.965  \scriptsize{$\pm$0.001} & 0.941  \scriptsize{$\pm$0.004} & 0.826  \scriptsize{$\pm$0.010} & 0.918  \scriptsize{$\pm$0.002} & 0.846  \scriptsize{$\pm$0.002} \\
        Ours (DA-v3) & 0.950  \scriptsize{$\pm$0.001}	& 0.902  \scriptsize{$\pm$0.002} & 0.965  \scriptsize{$\pm$0.002}	& 0.944  \scriptsize{$\pm$0.002} & 0.825  \scriptsize{$\pm$0.004}	& 0.909  \scriptsize{$\pm$0.005}	& 0.843  \scriptsize{$\pm$0.005} \\ 
        \bottomrule
        \end{tabular}
        }
        \caption{Average SSIM on NeRF Synthetic dataset}
        \label{tab:ssim_synth}
    \end{subtable}

    \vspace{3mm}

    \begin{subtable}[t]{\linewidth}
        \centering
        \resizebox{\linewidth}{!}{
        \begin{tabular}{cccccccccc}
        \toprule
        \multirow{2}{*}{\textbf{Methods}} & \multicolumn{7}{c}{LPIPS$\downarrow$} \\
        \cmidrule(lr){2-8}
        & chair & drums & hotdog & lego & materials & mic & ship \\	
        \midrule
        random & 0.070  \scriptsize{$\pm$0.016} & 0.111  \scriptsize{$\pm$0.009} & 0.068  \scriptsize{$\pm$0.004} & 0.076  \scriptsize{$\pm$0.001} & 0.181  \scriptsize{$\pm$0.008} & 0.108  \scriptsize{$\pm$0.008} & 0.202  \scriptsize{$\pm$0.007} \\
        ACP & 0.075  \scriptsize{$\pm$0.002} & 0.125  \scriptsize{$\pm$0.006} & 0.109  \scriptsize{$\pm$0.003} & 0.121  \scriptsize{$\pm$0.002} & 0.198  \scriptsize{$\pm$0.007} & 0.126  \scriptsize{$\pm$0.002} & 0.213  \scriptsize{$\pm$0.006} \\
        MUSIQ & 0.062  \scriptsize{$\pm$0.002} & 0.109  \scriptsize{$\pm$0.007} & 0.071  \scriptsize{$\pm$0.002} & 0.077  \scriptsize{$\pm$0.002} & 0.190  \scriptsize{$\pm$0.005} & 0.119  \scriptsize{$\pm$0.003} & 0.204  \scriptsize{$\pm$0.003} \\
        CrossScore & 0.069  \scriptsize{$\pm$0.002} & 0.125  \scriptsize{$\pm$0.003} & 0.088  \scriptsize{$\pm$0.004} & 0.086  \scriptsize{$\pm$0.003} & 0.213  \scriptsize{$\pm$0.005} & 0.123  \scriptsize{$\pm$0.003} & 0.203  \scriptsize{$\pm$0.004} \\
        FisherRF& 0.059  \scriptsize{$\pm$0.002} & 0.108  \scriptsize{$\pm$0.006} & 0.066  \scriptsize{$\pm$0.003} & 0.093  \scriptsize{$\pm$0.011} & 0.177  \scriptsize{$\pm$0.008} & 0.111  \scriptsize{$\pm$0.003} & 0.199  \scriptsize{$\pm$0.003} \\
        \midrule
        Ours (MASt3R) & 0.058  \scriptsize{$\pm$0.001} & 0.099  \scriptsize{$\pm$0.002} & 0.066  \scriptsize{$\pm$0.001} & 0.078  \scriptsize{$\pm$0.002} & 0.175  \scriptsize{$\pm$0.002} & 0.095  \scriptsize{$\pm$0.002} & 0.199  \scriptsize{$\pm$0.001} \\ 
        Ours (DA-v3) & 0.058  \scriptsize{$\pm$0.001} & 0.100  \scriptsize{$\pm$0.001} & 0.067  \scriptsize{$\pm$0.001} & 0.076  \scriptsize{$\pm$0.001} & 0.178  \scriptsize{$\pm$0.004} & 0.102  \scriptsize{$\pm$0.002} & 0.200  \scriptsize{$\pm$0.003} \\
        \bottomrule
        \end{tabular}
        }
        \caption{Average LPIPS on NeRF Synthetic dataset}
        \label{tab:lpips_synth}
    \end{subtable}
\end{table*}

\begin{table*}[h]
    \centering
    \caption{\textbf{Scene-wise quantitative results on Mip-NeRF 360.} Each subtable reports PSNR, SSIM, and LPIPS, respectively. All values are computed over four repeated trials with different random seeds.}
    \label{table:scenewise_metrics}
    \begin{subtable}[h]{\linewidth}
        \centering
        \resizebox{\linewidth}{!}{
        \begin{tabular}{cccccccccc}
        \toprule
        \multirow{2}{*}{\textbf{Methods}} & \multicolumn{9}{c}{PSNR$\uparrow$} \\
        \cmidrule(lr){2-10}
        & room & counter & kitchen & bonsai & bicycle & flowers & garden & stump & treehill \\
        \midrule
        random & 22.256  \scriptsize{$\pm$0.575} & 20.544  \scriptsize{$\pm$0.450} & 21.908  \scriptsize{$\pm$0.671} & 21.086  \scriptsize{$\pm$1.547} & 18.642  \scriptsize{$\pm$0.295} & 15.916  \scriptsize{$\pm$0.583} & 21.131  \scriptsize{$\pm$0.242} & 20.027  \scriptsize{$\pm$0.841} & 18.211  \scriptsize{$\pm$0.401} \\
        ACP & 21.868  \scriptsize{$\pm$0.604} & 21.390  \scriptsize{$\pm$0.382} & 21.279  \scriptsize{$\pm$0.464} & 22.158  \scriptsize{$\pm$0.101} & 19.290  \scriptsize{$\pm$0.213} & 16.543  \scriptsize{$\pm$0.192} & 21.542  \scriptsize{$\pm$0.105} & 20.575  \scriptsize{$\pm$0.339} & 18.278  \scriptsize{$\pm$0.455} \\
        MUSIQ & 22.057  \scriptsize{$\pm$0.097} & 20.452  \scriptsize{$\pm$0.190} & 22.804  \scriptsize{$\pm$0.194} & 20.088  \scriptsize{$\pm$0.375} & 18.332  \scriptsize{$\pm$0.146} & 16.946  \scriptsize{$\pm$0.069} & 20.736  \scriptsize{$\pm$0.245} & 19.217  \scriptsize{$\pm$0.297} & 18.016  \scriptsize{$\pm$0.269} \\
        CrossScore & 22.924  \scriptsize{$\pm$0.254} & 22.480  \scriptsize{$\pm$0.124} & 22.994  \scriptsize{$\pm$0.117} & 24.021  \scriptsize{$\pm$0.330} & 18.485  \scriptsize{$\pm$0.933} & 17.222  \scriptsize{$\pm$0.221} & 21.876  \scriptsize{$\pm$0.099} & 21.003  \scriptsize{$\pm$0.544} & 18.681  \scriptsize{$\pm$0.596} \\
        FisherRF & 22.500  \scriptsize{$\pm$0.642} & 21.613  \scriptsize{$\pm$0.110} & 23.123  \scriptsize{$\pm$0.361} & 23.125  \scriptsize{$\pm$0.733} & 18.715  \scriptsize{$\pm$0.153} & 16.616  \scriptsize{$\pm$0.158} & 21.459  \scriptsize{$\pm$0.101} & 20.230  \scriptsize{$\pm$0.316} & 18.396  \scriptsize{$\pm$0.209} \\
        \midrule
        Ours (MASt3R) & 24.513  \scriptsize{$\pm$0.110} & 22.742  \scriptsize{$\pm$0.029} & 24.182  \scriptsize{$\pm$0.111} & 24.564  \scriptsize{$\pm$0.025} & 18.182  \scriptsize{$\pm$0.116} & 16.930  \scriptsize{$\pm$0.154} & 22.182  \scriptsize{$\pm$0.035} & 20.605  \scriptsize{$\pm$0.224} & 18.789  \scriptsize{$\pm$0.374} \\
        Ours (DA-v3) & 24.321  \scriptsize{$\pm$0.498} & 22.608  \scriptsize{$\pm$0.172} & 24.074  \scriptsize{$\pm$0.068} & 23.886  \scriptsize{$\pm$0.245} & 18.520  \scriptsize{$\pm$0.217} & 16.848  \scriptsize{$\pm$0.189} & 22.195  \scriptsize{$\pm$0.074} & 20.576   \scriptsize{$\pm$0.303} & 19.219  \scriptsize{$\pm$0.274} \\
        \bottomrule
        \end{tabular}
        }
        \caption{Average PSNR on Mip-NeRF 360}
        \label{tab:psnr_mip}
    \end{subtable}

    \vspace{3mm}

    \begin{subtable}[h]{\linewidth}
        \centering
        \resizebox{\linewidth}{!}{
        \begin{tabular}{cccccccccc}
        \toprule
        \multirow{2}{*}{\textbf{Methods}} & \multicolumn{9}{c}{SSIM$\uparrow$} \\
        \cmidrule(lr){2-10}
        & room & counter & kitchen & bonsai & bicycle & flowers & garden & stump & treehill \\
        \midrule
        random & 0.782  \scriptsize{$\pm$0.017} & 0.719  \scriptsize{$\pm$0.011} & 0.774  \scriptsize{$\pm$0.022} & 0.758  \scriptsize{$\pm$0.037} & 0.412  \scriptsize{$\pm$0.008} & 0.318  \scriptsize{$\pm$0.012} & 0.578  \scriptsize{$\pm$0.008} & 0.457  \scriptsize{$\pm$0.028} & 0.457  \scriptsize{$\pm$0.009} \\
        ACP & 0.779  \scriptsize{$\pm$0.015} & 0.746  \scriptsize{$\pm$0.007} & 0.757  \scriptsize{$\pm$0.010} & 0.791  \scriptsize{$\pm$0.011} & 0.429  \scriptsize{$\pm$0.006} & 0.334  \scriptsize{$\pm$0.005} & 0.596  \scriptsize{$\pm$0.003} & 0.476  \scriptsize{$\pm$0.012} & 0.458  \scriptsize{$\pm$0.008} \\
        MUSIQ & 0.780  \scriptsize{$\pm$0.004} & 0.723  \scriptsize{$\pm$0.006} & 0.782  \scriptsize{$\pm$0.009} & 0.726  \scriptsize{$\pm$0.016} & 0.400  \scriptsize{$\pm$0.003} & 0.333  \scriptsize{$\pm$0.002} & 0.549  \scriptsize{$\pm$0.009} & 0.423  \scriptsize{$\pm$0.011} & 0.459  \scriptsize{$\pm$0.008} \\
        CrossScore & 0.811  \scriptsize{$\pm$0.004} & 0.772  \scriptsize{$\pm$0.003} & 0.803  \scriptsize{$\pm$0.004} & 0.836  \scriptsize{$\pm$0.002} & 0.397  \scriptsize{$\pm$0.034} & 0.340  \scriptsize{$\pm$0.006} & 0.593  \scriptsize{$\pm$0.006} & 0.494  \scriptsize{$\pm$0.021} & 0.465  \scriptsize{$\pm$0.010} \\
        FisherRF & 0.773  \scriptsize{$\pm$0.017} & 0.751  \scriptsize{$\pm$0.003} & 0.790  \scriptsize{$\pm$0.007} & 0.810  \scriptsize{$\pm$0.016} & 0.411  \scriptsize{$\pm$0.004} & 0.331  \scriptsize{$\pm$0.005} & 0.573  \scriptsize{$\pm$0.004} & 0.461  \scriptsize{$\pm$0.015} & 0.457  \scriptsize{$\pm$0.006} \\
        \midrule
        Ours (MASt3R) & 0.825  \scriptsize{$\pm$0.003} & 0.783  \scriptsize{$\pm$0.001} & 0.822  \scriptsize{$\pm$0.002} & 0.841  \scriptsize{$\pm$0.003} & 0.396  \scriptsize{$\pm$0.003} & 0.334  \scriptsize{$\pm$0.003} & 0.584  \scriptsize{$\pm$0.002} & 0.473  \scriptsize{$\pm$0.008} & 0.457  \scriptsize{$\pm$0.004} \\
        Ours (DA-v3) & 0.819  \scriptsize{$\pm$0.009} & 0.780  \scriptsize{$\pm$0.004} & 0.821  \scriptsize{$\pm$0.004} & 0.831  \scriptsize{$\pm$0.007} & 0.406  \scriptsize{$\pm$0.009} & 0.333  \scriptsize{$\pm$0.001} & 0.585  \scriptsize{$\pm$0.001} & 0.470  \scriptsize{$\pm$0.010} & 0.471  \scriptsize{$\pm$0.006} \\
        \bottomrule
        \end{tabular}
        }
        \caption{Average SSIM on Mip-NeRF 360}
        \label{tab:ssim_mip}
    \end{subtable}

    \vspace{3mm}

    \begin{subtable}[h]{\linewidth}
        \centering
        \resizebox{\linewidth}{!}{
        \begin{tabular}{cccccccccc}
        \toprule
        \multirow{2}{*}{\textbf{Methods}} & \multicolumn{9}{c}{LPIPS$\downarrow$} \\
        \cmidrule(lr){2-10}
        & room & counter & kitchen & bonsai & bicycle & flowers & garden & stump & treehill \\
        \midrule
        random & 0.357  \scriptsize{$\pm$0.009} & 0.382  \scriptsize{$\pm$0.010} & 0.288  \scriptsize{$\pm$0.018} & 0.364  \scriptsize{$\pm$0.024} & 0.566  \scriptsize{$\pm$0.001} & 0.612  \scriptsize{$\pm$0.011} & 0.417  \scriptsize{$\pm$0.002} & 0.556  \scriptsize{$\pm$0.016} & 0.561  \scriptsize{$\pm$0.006} \\
        ACP & 0.358  \scriptsize{$\pm$0.012} & 0.362  \scriptsize{$\pm$0.008} & 0.303  \scriptsize{$\pm$0.008} & 0.343  \scriptsize{$\pm$0.008} & 0.557  \scriptsize{$\pm$0.003} & 0.598  \scriptsize{$\pm$0.003} & 0.410  \scriptsize{$\pm$0.003} & 0.548  \scriptsize{$\pm$0.006} & 0.563  \scriptsize{$\pm$0.008} \\
        MUSIQ & 0.381  \scriptsize{$\pm$0.007} & 0.384  \scriptsize{$\pm$0.004} & 0.288  \scriptsize{$\pm$0.007} & 0.394  \scriptsize{$\pm$0.014} & 0.579  \scriptsize{$\pm$0.002} & 0.601  \scriptsize{$\pm$0.002} & 0.429  \scriptsize{$\pm$0.004} & 0.573  \scriptsize{$\pm$0.005} & 0.567  \scriptsize{$\pm$0.005} \\
        CrossScore & 0.354  \scriptsize{$\pm$0.003} & 0.355  \scriptsize{$\pm$0.003} & 0.269  \scriptsize{$\pm$0.002} & 0.318  \scriptsize{$\pm$0.003} & 0.605  \scriptsize{$\pm$0.045} & 0.609  \scriptsize{$\pm$0.002} & 0.418  \scriptsize{$\pm$0.006} & 0.541  \scriptsize{$\pm$0.010} & 0.564  \scriptsize{$\pm$0.007} \\
        FisherRF & 0.370  \scriptsize{$\pm$0.008} & 0.356  \scriptsize{$\pm$0.002} & 0.278  \scriptsize{$\pm$0.005} & 0.332  \scriptsize{$\pm$0.013} & 0.571  \scriptsize{$\pm$0.002} & 0.603  \scriptsize{$\pm$0.003} & 0.420  \scriptsize{$\pm$0.002} & 0.556  \scriptsize{$\pm$0.009} & 0.562  \scriptsize{$\pm$0.003} \\
        \midrule
        Ours (MASt3R) & 0.338  \scriptsize{$\pm$0.002} & 0.342  \scriptsize{$\pm$0.001} & 0.277  \scriptsize{$\pm$0.043} & 0.313  \scriptsize{$\pm$0.002} & 0.594  \scriptsize{$\pm$0.002} & 0.618  \scriptsize{$\pm$0.002} & 0.431  \scriptsize{$\pm$0.002} & 0.570  \scriptsize{$\pm$0.004} & 0.574  \scriptsize{$\pm$0.003} \\
        Ours (DA-v3) & 0.340  \scriptsize{$\pm$0.005} & 0.345  \scriptsize{$\pm$0.004} & 0.255  \scriptsize{$\pm$0.004} & 0.321  \scriptsize{$\pm$0.003} & 0.588  \scriptsize{$\pm$0.005} & 0.620  \scriptsize{$\pm$0.002} & 0.432  \scriptsize{$\pm$0.001} & 0.571  \scriptsize{$\pm$0.006} & 0.567  \scriptsize{$\pm$0.003} \\
        \bottomrule
        \end{tabular}
        }
        \caption{Average LPIPS on Mip-NeRF 360}
        \label{tab:lpips_mip}
    \end{subtable}
\end{table*}

\begin{table*}[h]
    \centering
    \caption{\textbf{Scene-wise quantitative results on the Extended datasets.} Each subtable reports PSNR, SSIM, and LPIPS, respectively. All values are computed over four repeated trials with different random seeds.}
    \label{table:scenewise_metrics_extended}
    \begin{subtable}[h]{0.8\linewidth}
        \centering
        \resizebox{\linewidth}{!}{
        \begin{tabular}{ccccccc}
        \toprule
        \multirow{2}{*}{\textbf{Methods}} & \multicolumn{5}{c}{PSNR$\uparrow$}\\
        \cmidrule(lr){2-6}
        & ballroom & horse & playroom & ponche & truck \\
        \midrule
        random & 16.970 \scriptsize{$\pm$0.320} & 19.119 \scriptsize{$\pm$0.513} & 18.506 \scriptsize{$\pm$0.311} & 19.981 \scriptsize{$\pm$0.899} & 21.737 \scriptsize{$\pm$0.451} \\
        ACP & 17.499 \scriptsize{$\pm$0.114} & 20.147 \scriptsize{$\pm$0.252} & 20.983 \scriptsize{$\pm$0.569} & 21.634 \scriptsize{$\pm$0.153} & 19.487 \scriptsize{$\pm$0.401} \\
        MUSIQ & 16.443 \scriptsize{$\pm$0.235} & 18.693 \scriptsize{$\pm$0.206} & 19.131 \scriptsize{$\pm$0.162} & 18.438 \scriptsize{$\pm$0.394} & 20.793 \scriptsize{$\pm$0.169} \\
        CrossScore & 18.099 \scriptsize{$\pm$0.085} & 20.354 \scriptsize{$\pm$0.119} & 19.842 \scriptsize{$\pm$0.150} & 19.899 \scriptsize{$\pm$0.809} & 21.517 \scriptsize{$\pm$0.525} \\
        FisherRF & 17.250 \scriptsize{$\pm$0.326} & 19.834 \scriptsize{$\pm$0.396} & 19.334 \scriptsize{$\pm$0.179} & 19.833 \scriptsize{$\pm$0.828} & 22.017 \scriptsize{$\pm$0.096} \\
        \midrule
        Ours (MASt3R) & 18.281 \scriptsize{$\pm$0.179} & 20.015 \scriptsize{$\pm$0.232} & 20.689 \scriptsize{$\pm$1.105} & 20.906 \scriptsize{$\pm$1.067} & 22.115 \scriptsize{$\pm$0.256} \\
        Ours (DA-v3) & 18.216 \scriptsize{$\pm$0.133} & 20.471 \scriptsize{$\pm$0.227} & 20.271 \scriptsize{$\pm$0.309} & 20.727 \scriptsize{$\pm$1.111} & 22.058 \scriptsize{$\pm$0.178} \\
        \bottomrule
        \end{tabular}
        }
        \caption{Average PSNR on Extended datasets}
        \label{tab:psnr_ext}
    \end{subtable}

    \vspace{3mm}

    \begin{subtable}[H]{0.8\linewidth}
        \centering
        \resizebox{\linewidth}{!}{
        \begin{tabular}{cccccc}
        \toprule
        \multirow{2}{*}{\textbf{Methods}} & \multicolumn{5}{c}{SSIM$\uparrow$} \\
        \cmidrule(lr){2-6}
        & ballroom & horse & playroom & ponche & truck \\
        \midrule
        random & 0.551 \scriptsize{$\pm$0.012} & 0.756 \scriptsize{$\pm$0.011} & 0.679 \scriptsize{$\pm$0.012} & 0.753 \scriptsize{$\pm$0.018} & 0.757 \scriptsize{$\pm$0.008} \\
        ACP & 0.564 \scriptsize{$\pm$0.003} & 0.787 \scriptsize{$\pm$0.007} & 0.772 \scriptsize{$\pm$0.007} & 0.752 \scriptsize{$\pm$0.004} & 0.713 \scriptsize{$\pm$0.016} \\
        MUSIQ & 0.524 \scriptsize{$\pm$0.008} & 0.749 \scriptsize{$\pm$0.008} & 0.700 \scriptsize{$\pm$0.005} & 0.723 \scriptsize{$\pm$0.004} & 0.741 \scriptsize{$\pm$0.002} \\
        CrossScore & 0.605  \scriptsize{$\pm$0.004} & 0.792 \scriptsize{$\pm$0.003} & 0.724  \scriptsize{$\pm$0.004} & 0.753  \scriptsize{$\pm$0.012} & 0.759  \scriptsize{$\pm$0.005} \\
        FisherRF & 0.555  \scriptsize{$\pm$0.013} & 0.777  \scriptsize{$\pm$0.007} & 0.709  \scriptsize{$\pm$0.006} & 0.751  \scriptsize{$\pm$0.016} & 0.763  \scriptsize{$\pm$0.002} \\
        \midrule
        Ours (MASt3R) & 0.614  \scriptsize{$\pm$0.007} & 0.779  \scriptsize{$\pm$0.007} & 0.741  \scriptsize{$\pm$0.016} & 0.765  \scriptsize{$\pm$0.013} & 0.761  \scriptsize{$\pm$0.002} \\
        Ours (DA-v3) & 0.610  \scriptsize{$\pm$0.002} & 0.791  \scriptsize{$\pm$0.005} & 0.742  \scriptsize{$\pm$0.013} & 0.759  \scriptsize{$\pm$0.008} & 0.761  \scriptsize{$\pm$0.002} \\
        \bottomrule
        \end{tabular}
        }
        \caption{Average SSIM on Extended datasets}
        \label{tab:ssim_ext}
    \end{subtable}

    \begin{subtable}[H]{0.8\linewidth}
        \centering
        \resizebox{\linewidth}{!}{
        \begin{tabular}{cccccc}
        \toprule
        \multirow{2}{*}{\textbf{Methods}} & \multicolumn{5}{c}{LPIPS$\downarrow$} \\
        \cmidrule(lr){2-6}
        & ballroom & horse & playroom & ponche & truck \\
        \midrule
        random & 0.385 \scriptsize{$\pm$0.007} & 0.286 \scriptsize{$\pm$0.009} & 0.348 \scriptsize{$\pm$0.011} & 0.466 \scriptsize{$\pm$0.019} & 0.388 \scriptsize{$\pm$0.006} \\
        ACP & 0.376 \scriptsize{$\pm$0.002} & 0.259 \scriptsize{$\pm$0.005} & 0.449 \scriptsize{$\pm$0.005} & 0.397 \scriptsize{$\pm$0.003} & 0.322 \scriptsize{$\pm$0.014} \\
        MUSIQ & 0.411 \scriptsize{$\pm$0.006} & 0.298 \scriptsize{$\pm$0.008} & 0.338 \scriptsize{$\pm$0.003} & 0.506 \scriptsize{$\pm$0.004} & 0.401 \scriptsize{$\pm$0.001} \\
        CrossScore & 0.352 \scriptsize{$\pm$0.003} & 0.257 \scriptsize{$\pm$0.002} & 0.318 \scriptsize{$\pm$0.003} & 0.466 \scriptsize{$\pm$0.014} & 0.386 \scriptsize{$\pm$0.003} \\
        FisherRF & 0.389 \scriptsize{$\pm$0.009} & 0.273 \scriptsize{$\pm$0.007} & 0.324 \scriptsize{$\pm$0.004} & 0.479 \scriptsize{$\pm$0.016} & 0.387 \scriptsize{$\pm$0.001} \\
        \midrule
        Ours (MASt3R) & 0.350  \scriptsize{$\pm$0.005} & 0.277  \scriptsize{$\pm$0.015} & 0.338  \scriptsize{$\pm$0.041} & 0.437  \scriptsize{$\pm$0.031} & 0.403  \scriptsize{$\pm$0.002} \\
        Ours (DA-v3) & 0.353  \scriptsize{$\pm$0.004} & 0.272  \scriptsize{$\pm$0.004} & 0.352  \scriptsize{$\pm$0.071} & 0.444  \scriptsize{$\pm$0.027} & 0.404  \scriptsize{$\pm$0.002} \\
        \bottomrule
        \end{tabular}
        }
        \caption{Average LPIPS on Extended datasets}
        \label{tab:lpips_ext}
    \end{subtable}
\end{table*}
\newpage

\begin{table*}[h]
    \centering
    \caption{\textbf{Scene-wise quantitative results of the ablation study on Mip-NeRF 360.} Each subtable reports PSNR, SSIM, and LPIPS, respectively.}
    \label{table:scenewise_ablation_mipnerf}
    \begin{subtable}[h]{1.0\linewidth}
        \centering
        \resizebox{\linewidth}{!}{
        \begin{tabular}{ccccccccccc}
        \toprule
        \multirow{2}{*}{\textbf{Methods}} & \multirow{2}{*}{\textbf{3D Lifting}} & \multicolumn{9}{c}{PSNR$\uparrow$} \\
        \cmidrule(lr){3-11}
        & & room & counter & kitchen & bonsai & bicycle & flowers & garden & stump & treehill  \\
        \midrule
        FisherRF$^{\dagger}$ & - & 22.500 \scriptsize{$\pm$0.642} & 21.613 \scriptsize{$\pm$0.110} & 23.123 \scriptsize{$\pm$0.361} & 23.125 \scriptsize{$\pm$0.733} & 18.715 \scriptsize{$\pm$0.153} & 16.616 \scriptsize{$\pm$0.158} & 21.459 \scriptsize{$\pm$0.101} & 20.230 \scriptsize{$\pm$0.316} & 18.396 \scriptsize{$\pm$0.209} \\
        \midrule
        $^{\ddagger}$+ResGS & - & 23.637 \scriptsize{$\pm$0.053} & 21.694 \scriptsize{$\pm$0.097} & 24.052 \scriptsize{$\pm$0.034} & 23.652 \scriptsize{$\pm$0.106} & 19.278 \scriptsize{$\pm$0.041} & 16.463 \scriptsize{$\pm$0.098} & 21.438 \scriptsize{$\pm$0.033} & 20.630 \scriptsize{$\pm$0.013} & 18.558 \scriptsize{$\pm$0.028} \\
        $^{\dagger}$+ResGS & - & 23.295 \scriptsize{$\pm$0.170} & 21.650 \scriptsize{$\pm$0.269} & 23.657 \scriptsize{$\pm$0.089} & 22.501 \scriptsize{$\pm$0.329} & 18.621 \scriptsize{$\pm$0.380} & 16.404 \scriptsize{$\pm$0.231} & 21.506 \scriptsize{$\pm$0.102} & 20.250 \scriptsize{$\pm$0.361} & 18.700 \scriptsize{$\pm$0.289} \\
        \midrule
        $^{\dagger}$+SA-HashGS & MASt3R & 23.998 \scriptsize{$\pm$0.313} & 22.512 \scriptsize{$\pm$0.091} & 23.576 \scriptsize{$\pm$0.158} & 24.157 \scriptsize{$\pm$0.097} & 18.173 \scriptsize{$\pm$0.100} & 16.732 \scriptsize{$\pm$0.039} & 21.944 \scriptsize{$\pm$0.231} & 20.155 \scriptsize{$\pm$0.337} & 18.587 \scriptsize{$\pm$0.372} \\
        $^{\dagger}$+SA-ResGS & MASt3R & 24.513 \scriptsize{$\pm$0.110} & 22.742 \scriptsize{$\pm$0.029} & 24.182 \scriptsize{$\pm$0.111} & 24.564 \scriptsize{$\pm$0.025} & 18.182 \scriptsize{$\pm$0.116} & 16.930 \scriptsize{$\pm$0.154} & 22.182 \scriptsize{$\pm$0.035} & 20.605 \scriptsize{$\pm$0.224} & 18.789 \scriptsize{$\pm$0.374} \\
        \midrule
        $^{\dagger}$+SA-HashGS & DA-v3 & 24.456 \scriptsize{$\pm$0.142} & 22.768 \scriptsize{$\pm$0.104} & 23.726 \scriptsize{$\pm$0.093} & 23.609 \scriptsize{$\pm$0.675} & 17.954 \scriptsize{$\pm$0.199} & 16.378 \scriptsize{$\pm$0.149} & 22.098 \scriptsize{$\pm$0.058} & 20.182 \scriptsize{$\pm$0.418} & 18.513 \scriptsize{$\pm$0.277} \\
        $^{\dagger}$+SA-ResGS & DA-v3 & 24.321 \scriptsize{$\pm$0.498} & 22.608 \scriptsize{$\pm$0.172} & 24.074 \scriptsize{$\pm$0.068} & 23.886 \scriptsize{$\pm$0.245} & 18.520 \scriptsize{$\pm$0.217} & 16.848 \scriptsize{$\pm$0.189} & 22.195 \scriptsize{$\pm$0.074} & 20.576 \scriptsize{$\pm$0.303} & 19.219 \scriptsize{$\pm$0.274} \\
        \bottomrule
        \end{tabular}
        }
        \caption{Average PSNR results for Ablation study on Mip-NeRF 360 dataset}
        \label{tab:psnr_ext}
    \end{subtable}
    \begin{subtable}[h]{1.0\linewidth}
        \centering
        \resizebox{\linewidth}{!}{
        \begin{tabular}{ccccccccccc}
        \toprule
        \multirow{2}{*}{\textbf{Methods}} & \multirow{2}{*}{\textbf{3D Lifting}} & \multicolumn{9}{c}{SSIM$\uparrow$} \\
        \cmidrule(lr){3-11}
        & & room & counter & kitchen & bonsai & bicycle & flowers & garden & stump & treehill  \\
        \midrule
        FisherRF$^{\dagger}$ & - & 0.773 \scriptsize{$\pm$0.017} & 0.751 \scriptsize{$\pm$0.003} & 0.790 \scriptsize{$\pm$0.007} & 0.810 \scriptsize{$\pm$0.016} & 0.411 \scriptsize{$\pm$0.004} & 0.331 \scriptsize{$\pm$0.005} & 0.573 \scriptsize{$\pm$0.004} & 0.461 \scriptsize{$\pm$0.015} & 0.457 \scriptsize{$\pm$0.006} \\
        \midrule
        $^{\ddagger}$+ResGS & - & 0.800 \scriptsize{$\pm$0.002} & 0.754 \scriptsize{$\pm$0.002} & 0.814 \scriptsize{$\pm$0.000} & 0.819 \scriptsize{$\pm$0.001} & 0.426 \scriptsize{$\pm$0.001} & 0.331 \scriptsize{$\pm$0.001} & 0.555 \scriptsize{$\pm$0.001} & 0.472 \scriptsize{$\pm$0.000} & 0.461 \scriptsize{$\pm$0.000} \\
        $^{\dagger}$+ResGS & - & 0.788 \scriptsize{$\pm$0.004} & 0.750 \scriptsize{$\pm$0.006} & 0.795 \scriptsize{$\pm$0.006} & 0.799 \scriptsize{$\pm$0.010} & 0.408 \scriptsize{$\pm$0.009} & 0.325 \scriptsize{$\pm$0.004} & 0.560 \scriptsize{$\pm$0.004} & 0.460 \scriptsize{$\pm$0.010} & 0.462 \scriptsize{$\pm$0.007} \\
        \midrule
        $^{\dagger}$+SA-HashGS & MASt3R & 0.820 \scriptsize{$\pm$0.005} & 0.779 \scriptsize{$\pm$0.004} & 0.810 \scriptsize{$\pm$0.004} & 0.836 \scriptsize{$\pm$0.002} & 0.396 \scriptsize{$\pm$0.006} & 0.334 \scriptsize{$\pm$0.003} & 0.594 \scriptsize{$\pm$0.012} & 0.459 \scriptsize{$\pm$0.011} & 0.455 \scriptsize{$\pm$0.009} \\
        $^{\dagger}$+SA-ResGS & MASt3R & 0.825 \scriptsize{$\pm$0.003} & 0.783 \scriptsize{$\pm$0.001} & 0.822 \scriptsize{$\pm$0.002} & 0.841 \scriptsize{$\pm$0.003} & 0.396 \scriptsize{$\pm$0.003} & 0.334 \scriptsize{$\pm$0.003} & 0.584 \scriptsize{$\pm$0.002} & 0.473 \scriptsize{$\pm$0.008} & 0.457 \scriptsize{$\pm$0.004} \\
        \midrule
        $^{\dagger}$+SA-HashGS & DA-v3 & 0.824 \scriptsize{$\pm$0.005} & 0.783 \scriptsize{$\pm$0.003} & 0.812 \scriptsize{$\pm$0.004} & 0.822 \scriptsize{$\pm$0.013} & 0.394 \scriptsize{$\pm$0.006} & 0.333 \scriptsize{$\pm$0.002} & 0.602 \scriptsize{$\pm$0.002} & 0.459 \scriptsize{$\pm$0.016} & 0.457 \scriptsize{$\pm$0.007} \\
        $^{\dagger}$+SA-ResGS & DA-v3 & 0.819 \scriptsize{$\pm$0.009} & 0.780 \scriptsize{$\pm$0.004} & 0.821 \scriptsize{$\pm$0.004} & 0.831 \scriptsize{$\pm$0.007} & 0.406 \scriptsize{$\pm$0.009} & 0.333 \scriptsize{$\pm$0.001} & 0.585 \scriptsize{$\pm$0.001} & 0.470 \scriptsize{$\pm$0.010} & 0.471 \scriptsize{$\pm$0.006} \\
        \bottomrule
        \end{tabular}
        }
        \caption{Average SSIM results for Ablation study on Mip-NeRF 360 dataset}
        \label{tab:psnr_ext}
    \end{subtable}
    \vspace{3mm}
    \begin{subtable}[h]{1.0\linewidth}
        \centering
        \resizebox{\linewidth}{!}{
        \begin{tabular}{ccccccccccc}
        \toprule
        \multirow{2}{*}{\textbf{Methods}} & \multirow{2}{*}{\textbf{3D Lifting}} & \multicolumn{9}{c}{LPIPS$\downarrow$} \\
        \cmidrule(lr){3-11}
        & & room & counter & kitchen & bonsai & bicycle & flowers & garden & stump & treehill  \\
        \midrule
        FisherRF$^{\dagger}$ & - & 0.370 \scriptsize{$\pm$0.008} & 0.356 \scriptsize{$\pm$0.002} & 0.278 \scriptsize{$\pm$0.005} & 0.332 \scriptsize{$\pm$0.013} & 0.571 \scriptsize{$\pm$0.002} & 0.603 \scriptsize{$\pm$0.003} & 0.420 \scriptsize{$\pm$0.002} & 0.556 \scriptsize{$\pm$0.009} & 0.562 \scriptsize{$\pm$0.003} \\
        \midrule
        $^{\ddagger}$+ResGS & - & 0.354 \scriptsize{$\pm$0.001} & 0.357 \scriptsize{$\pm$0.001} & 0.261 \scriptsize{$\pm$0.000} & 0.326 \scriptsize{$\pm$0.000} & 0.576 \scriptsize{$\pm$0.001} & 0.621 \scriptsize{$\pm$0.001} & 0.443 \scriptsize{$\pm$0.001} & 0.569 \scriptsize{$\pm$0.000} & 0.569 \scriptsize{$\pm$0.000} \\
        $^{\dagger}$+ResGS & - & 0.367 \scriptsize{$\pm$0.004} & 0.362 \scriptsize{$\pm$0.004} & 0.279 \scriptsize{$\pm$0.007} & 0.339 \scriptsize{$\pm$0.004} & 0.587 \scriptsize{$\pm$0.004} & 0.627 \scriptsize{$\pm$0.004} & 0.441 \scriptsize{$\pm$0.002} & 0.576 \scriptsize{$\pm$0.004} & 0.570 \scriptsize{$\pm$0.004} \\
        \midrule
        $^{\dagger}$+SA-HashGS & MASt3R & 0.338 \scriptsize{$\pm$0.003} & 0.341 \scriptsize{$\pm$0.003} & 0.263 \scriptsize{$\pm$0.003} & 0.313 \scriptsize{$\pm$0.001} & 0.580 \scriptsize{$\pm$0.003} & 0.603 \scriptsize{$\pm$0.004} & 0.412 \scriptsize{$\pm$0.003} & 0.558 \scriptsize{$\pm$0.007} & 0.564 \scriptsize{$\pm$0.007} \\
        $^{\dagger}$+SA-ResGS & MASt3R & 0.338 \scriptsize{$\pm$0.002} & 0.342 \scriptsize{$\pm$0.001} & 0.277 \scriptsize{$\pm$0.043} & 0.313 \scriptsize{$\pm$0.002} & 0.594 \scriptsize{$\pm$0.002} & 0.618 \scriptsize{$\pm$0.002} & 0.431 \scriptsize{$\pm$0.002} & 0.570 \scriptsize{$\pm$0.004} & 0.574 \scriptsize{$\pm$0.003} \\
        \midrule
        $^{\dagger}$+SA-HashGS & DA-v3 & 0.335 \scriptsize{$\pm$0.004} & 0.337 \scriptsize{$\pm$0.003} & 0.260 \scriptsize{$\pm$0.004} & 0.322 \scriptsize{$\pm$0.009} & 0.580 \scriptsize{$\pm$0.002} & 0.603 \scriptsize{$\pm$0.001} & 0.409 \scriptsize{$\pm$0.002} & 0.559 \scriptsize{$\pm$0.009} & 0.563 \scriptsize{$\pm$0.004} \\
        $^{\dagger}$+SA-ResGS & DA-v3 & 0.340 \scriptsize{$\pm$0.005} & 0.345 \scriptsize{$\pm$0.004} & 0.255 \scriptsize{$\pm$0.004} & 0.321 \scriptsize{$\pm$0.003} & 0.588 \scriptsize{$\pm$0.005} & 0.620 \scriptsize{$\pm$0.002} & 0.432 \scriptsize{$\pm$0.001} & 0.571 \scriptsize{$\pm$0.006} & 0.567 \scriptsize{$\pm$0.003} \\
        \bottomrule
        \end{tabular}
        }
        \caption{Average LPIPS results for Ablation study on Mip-NeRF 360 dataset}
        \label{tab:psnr_ext}
    \end{subtable}
    \vspace{3mm}
\end{table*}

\begin{table*}[h]
    \centering
    \caption{\textbf{Scene-wise quantitative results of the ablation study on the Extended datasets.} Each subtable reports PSNR, SSIM, and LPIPS, respectively.}
    \label{table:scenewise_ablation_extended}
    \begin{subtable}[h]{0.9\linewidth}
        \centering
        \resizebox{\linewidth}{!}{
        \begin{tabular}{ccccccccccc}
        \toprule
        \multirow{2}{*}{\textbf{Methods}} & \multirow{2}{*}{\textbf{3D Lifting}} & \multicolumn{5}{c}{PSNR$\uparrow$} \\
        \cmidrule(lr){3-7}
        & & ballroom & horse & playroom & ponche & truck \\
        \midrule
        FisherRF$^{\dagger}$ & - & 17.075 \scriptsize{$\pm$0.197} & 19.580 \scriptsize{$\pm$0.346} & 19.383 \scriptsize{$\pm$0.276} & 19.447 \scriptsize{$\pm$0.508} & 21.964 \scriptsize{$\pm$0.022} \\
        \midrule
        $^{\ddagger}$+ResGS & - & 17.181 \scriptsize{$\pm$0.044} & 20.199 \scriptsize{$\pm$0.104} & 19.318 \scriptsize{$\pm$0.038} & 19.150 \scriptsize{$\pm$0.155} & 22.159 \scriptsize{$\pm$0.222} \\
        $^{\dagger}$+ResGS & - & 17.781 \scriptsize{$\pm$0.313} & 20.140 \scriptsize{$\pm$0.266} & 19.608 \scriptsize{$\pm$0.155} & 19.127 \scriptsize{$\pm$0.242} & 22.300 \scriptsize{$\pm$0.221} \\
        \midrule
        $^{\dagger}$+SA-HashGS & MASt3R & 17.609 \scriptsize{$\pm$0.189} & 19.106 \scriptsize{$\pm$0.306} & 19.781 \scriptsize{$\pm$0.127} & 19.921 \scriptsize{$\pm$0.354} & 22.095 \scriptsize{$\pm$0.173} \\
        $^{\dagger}$+SA-ResGS & MASt3R & 18.281 \scriptsize{$\pm$0.179} & 20.015 \scriptsize{$\pm$0.232} & 20.689 \scriptsize{$\pm$1.105} & 20.906 \scriptsize{$\pm$1.067} & 22.115 \scriptsize{$\pm$0.256} \\
        \midrule
        $^{\dagger}$+SA-HashGS & DA-v3 & 17.830 \scriptsize{$\pm$0.149} & 20.043 \scriptsize{$\pm$0.220} & 19.885 \scriptsize{$\pm$0.196} & 20.239 \scriptsize{$\pm$0.697} & 22.126 \scriptsize{$\pm$0.224} \\
        $^{\dagger}$+SA-ResGS & DA-v3 & 18.216 \scriptsize{$\pm$0.133} & 20.471 \scriptsize{$\pm$0.227} & 20.271 \scriptsize{$\pm$0.309} & 20.727 \scriptsize{$\pm$1.111} & 22.058 \scriptsize{$\pm$0.178} \\
        \bottomrule
        \end{tabular}
        }
        \caption{Average PSNR results for Ablation study on Mip-NeRF 360 dataset}
        \label{tab:psnr_ext}
    \end{subtable}
    \vspace{3mm}
    \begin{subtable}[h]{0.9\linewidth}
        \centering
        \resizebox{\linewidth}{!}{
        \begin{tabular}{ccccccccccc}
        \toprule
        \multirow{2}{*}{\textbf{Methods}} & \multirow{2}{*}{\textbf{3D Lifting}} & \multicolumn{5}{c}{SSIM$\uparrow$} \\
        \cmidrule(lr){3-7}
        & & ballroom & horse & playroom & ponche & truck \\
        \midrule
        FisherRF$^{\dagger}$ & - & 0.550 \scriptsize{$\pm$0.011} & 0.758 \scriptsize{$\pm$0.035} & 0.724 \scriptsize{$\pm$0.037} & 0.744 \scriptsize{$\pm$0.012} & 0.763 \scriptsize{$\pm$0.001} \\
        \midrule
        $^{\ddagger}$+ResGS & - & 0.555 \scriptsize{$\pm$0.002} & 0.785 \scriptsize{$\pm$0.001} & 0.704 \scriptsize{$\pm$0.002} & 0.743 \scriptsize{$\pm$0.002} & 0.758 \scriptsize{$\pm$0.002} \\
        $^{\dagger}$+ResGS & - & 0.577 \scriptsize{$\pm$0.009} & 0.780 \scriptsize{$\pm$0.007} & 0.717 \scriptsize{$\pm$0.006} & 0.742 \scriptsize{$\pm$0.002} & 0.760 \scriptsize{$\pm$0.002} \\
        \midrule
        $^{\dagger}$+SA-HashGS & MASt3R & 0.584 \scriptsize{$\pm$0.011} & 0.757 \scriptsize{$\pm$0.007} & 0.726 \scriptsize{$\pm$0.002} & 0.756 \scriptsize{$\pm$0.009} & 0.766 \scriptsize{$\pm$0.003} \\
        $^{\dagger}$+SA-ResGS & MASt3R & 0.614 \scriptsize{$\pm$0.007} & 0.779 \scriptsize{$\pm$0.007} & 0.741 \scriptsize{$\pm$0.016} & 0.765 \scriptsize{$\pm$0.013} & 0.761 \scriptsize{$\pm$0.002} \\
        \midrule
        $^{\dagger}$+SA-HashGS & DA-v3 & 0.594 \scriptsize{$\pm$0.006} & 0.785 \scriptsize{$\pm$0.005} & 0.728 \scriptsize{$\pm$0.006} & 0.760 \scriptsize{$\pm$0.007} & 0.766 \scriptsize{$\pm$0.004} \\
        $^{\dagger}$+SA-ResGS & DA-v3 & 0.610 \scriptsize{$\pm$0.002} & 0.791 \scriptsize{$\pm$0.005} & 0.742 \scriptsize{$\pm$0.013} & 0.759 \scriptsize{$\pm$0.008} & 0.761 \scriptsize{$\pm$0.002} \\
        \bottomrule
        \end{tabular}
        }
        \caption{Average SSIM results for Ablation study on Extended dataset}
        \label{tab:psnr_ext}
    \end{subtable}
    \begin{subtable}[h]{0.9\linewidth}
        \centering
        \resizebox{\linewidth}{!}{
        \begin{tabular}{ccccccccccc}
        \toprule
        \multirow{2}{*}{\textbf{Methods}} & \multirow{2}{*}{\textbf{3D Lifting}} & \multicolumn{5}{c}{LPIPS$\downarrow$} \\
        \cmidrule(lr){3-7}
        & & ballroom & horse & playroom & ponche & truck \\
        \midrule
        FisherRF$^{\dagger}$ & - & 0.391 \scriptsize{$\pm$0.007} & 0.287 \scriptsize{$\pm$0.026} & 0.312 \scriptsize{$\pm$0.028} & 0.485 \scriptsize{$\pm$0.010} & 0.388 \scriptsize{$\pm$0.001} \\
        \midrule
        $^{\ddagger}$+ResGS & - & 0.389 \scriptsize{$\pm$0.001} & 0.276 \scriptsize{$\pm$0.001} & 0.335 \scriptsize{$\pm$0.001} & 0.485 \scriptsize{$\pm$0.003} & 0.407 \scriptsize{$\pm$0.001} \\
        $^{\dagger}$+ResGS & - & 0.376 \scriptsize{$\pm$0.002} & 0.280 \scriptsize{$\pm$0.005} & 0.328 \scriptsize{$\pm$0.005} & 0.480 \scriptsize{$\pm$0.005} & 0.406 \scriptsize{$\pm$0.001} \\
        \midrule
        $^{\dagger}$+SA-HashGS & MASt3R & 0.363 \scriptsize{$\pm$0.008} & 0.290 \scriptsize{$\pm$0.008} & 0.313 \scriptsize{$\pm$0.002} & 0.463 \scriptsize{$\pm$0.007} & 0.384 \scriptsize{$\pm$0.002} \\
        $^{\dagger}$+SA-ResGS & MASt3R & 0.350 \scriptsize{$\pm$0.005} & 0.277 \scriptsize{$\pm$0.015} & 0.338 \scriptsize{$\pm$0.041} & 0.437 \scriptsize{$\pm$0.031} & 0.403 \scriptsize{$\pm$0.002} \\
        \midrule
        $^{\dagger}$+SA-HashGS & DA-v3 & 0.360 \scriptsize{$\pm$0.003} & 0.266 \scriptsize{$\pm$0.004} & 0.313 \scriptsize{$\pm$0.005} & 0.462 \scriptsize{$\pm$0.009} & 0.382 \scriptsize{$\pm$0.004} \\
        $^{\dagger}$+SA-ResGS & DA-v3 & 0.353 \scriptsize{$\pm$0.004}	& 0.272 \scriptsize{$\pm$0.004} & 0.352 \scriptsize{$\pm$0.071} & 0.444 \scriptsize{$\pm$0.027} & 0.404 \scriptsize{$\pm$0.002} \\
        \bottomrule
        \end{tabular}
        }
        \caption{Average LPIPS results for Ablation study on Extended dataset}
        \label{tab:psnr_ext}
    \end{subtable}
\end{table*}
\section{The Evaluation Statistics}
\label{Supp:statistics}


\rev{
\paragraph{Active View Selection}
Per-scene results for active view selection are reported in Table~\ref{table:scenewise_metrics_synth},~\ref{table:scenewise_metrics}, and~\ref{table:scenewise_metrics_extended}, where each value is reported as the mean and standard deviation over four runs with different random seeds.
Across the three datasets, our method generally achieves stronger PSNR and SSIM than competing methods, while remaining competitive on LPIPS. 
The standard deviations further suggest that our method improves stability in active reconstruction.
These results support the effectiveness of the proposed strategy across diverse input conditions.
}

\rev{
For several scenes in the Extended dataset, such as \textit{playroom}, \textit{ponche}, and \textit{truck}, our method exhibits relatively large standard deviations.
Although the standard deviations in these scenes appear less consistent, closer inspection across repeated runs shows that the model maintains strong and stable 
performance.
At the same time, the observed variance indicates further room for improvement, as a small number of runs achieve noticeably larger gains that increase both the mean and the overall spread.
We attribute this behavior to the stochastic training dynamics of our full framework, where physically grounded view selection and residual learning with uncertainty-guided Gaussian selection can reinforce each other and occasionally yield significant performance boosts.
This synergy benefits not only training stability but also the stabilization of uncertainty estimation, which in turn improves view selection quality.
Therefore, we interpret the larger variance not as a sign of instability, but as evidence that the proposed method remains robust while retaining additional upside for further performance gains.
}

\rev{
We also observe that ACP performs well on several scenes in the Extended dataset.
This suggests that its rule-based global exploration can be effective for certain scene layouts. 
However, the per-scene analysis also shows that such a hand-crafted strategy can suffer noticeable degradation depending on scene characteristics. 
In contrast, our method combines Laplacian-based uncertainty estimation with physically grounded prefiltering. 
As a result, it achieves more stable and consistently strong performance across a wider range of scene configurations.
}

\rev{
\paragraph{Ablation Study}
Per-scene ablation results are also summarized in Table~\ref{table:scenewise_ablation_mipnerf} and~\ref{table:scenewise_ablation_extended}, where all values are reported as the mean and standard deviation over four repeated runs.
Adding either ResGS or Hash prefiltering to the fixed-view setting improves not only the average performance but also the stability of training, as evidenced by the reduced standard deviations. 
This finding suggests that both proposed components contribute meaningfully to stabilizing the learning process. 
This trend is consistent with our discussion in the manuscript.
}

\rev{
When ResGS is used alone without any constraint on view selection ($^\ddagger$ResGS), the improvements become noticeably less consistent across scenes and runs. 
This suggests that the benefits of ResGS cannot be fully realized without an appropriate view selection mechanism. 
In contrast, combining it with view selection generally yields stronger results than using either component in isolation.
These results support our discussion in the main manuscript that the proposed components are complementary and thus exhibit a synergistic effect in most scenes.
}

\clearpage

\section{Supplementary Video Overview}
\label{Supp:supple_video_overview}

To complement the static visualizations provided in this document, we include a supplementary video that offers dynamic and comprehensive renderings of our results. 
This video is intended to provide a deeper visual understanding of the improvements achieved by our proposed SA-ResGS framework across various evaluation scenarios. 
The video includes:
 \begin{itemize}
    \item \textbf{Novel View Rendering.} We present extended 360-degree novel-view trajectories captured along spiral and circular camera paths, going beyond the discrete test views shown in the main paper and supplementary figures. These renderings highlight the effectiveness of our physically grounded view selection and residual supervision in preserving structural consistency and photometric quality across challenging viewpoints.
    \item \textbf{Ablation Study Comparisons.} To illustrate the impact of each component, we show side-by-side comparisons of different model variants under continuous camera movement. These scenes demonstrate robustness and fidelity improvements from residual supervision and surface-aware physically grounded view selection, especially in sparse-view or occluded regions.
\end{itemize}

\end{document}